\documentclass[conference]{IEEEtran}
\usepackage{cite}
\usepackage{amsmath,amssymb,amsfonts}
\usepackage{graphicx}
\usepackage{textcomp}
\usepackage{nicefrac}
\usepackage{pifont}
\usepackage{soul}
\usepackage[font=footnotesize,labelfont=bf]{caption}
\usepackage{enumitem}
\usepackage[caption=false, font=footnotesize]{subfig}
\usepackage[table,xcdraw]{xcolor}
\usepackage{titlesec}
\usepackage{anyfontsize}
\usepackage[normalem]{ulem}
\usepackage{wrapfig}
\usepackage{multirow}
\usepackage[export]{adjustbox}
\usepackage{stackengine}    
\usepackage{float}    
\usepackage[edges]{forest}
\usepackage{array}
\usepackage{stfloats}
\usepackage{indentfirst}
\usepackage{booktabs}
\usepackage{pifont}
\usepackage{placeins}
\usepackage{makecell}
\usepackage{natbib}
\usepackage{xspace}
\usepackage{hyperref}
\usepackage[hyperpageref]{backref}
\hypersetup{
 colorlinks,
 linkcolor=blue,
 urlbordercolor=black,
 pdfborderstyle={/S/U/W 1},
 breaklinks
 }

\usepackage{ragged2e}
 
\usepackage[capitalize]{cleveref}
\crefname{section}{Sec.}{Secs.}
\Crefname{section}{Section}{Sections}
\Crefname{table}{Table}{Tables}
\crefname{table}{Tab.}{Tabs.}

\definecolor{folderbg}{RGB}{124,166,198}
\definecolor{folderborder}{RGB}{110,144,169}
\definecolor{IGNGREEN}{RGB}{153, 211, 142}
\definecolor{TITLES}{RGB}{153, 211, 142}
\definecolor{TITLES_PRE}{RGB}{247, 212, 188}

\newcommand*\bestb[1]{\cellcolor[HTML]{ebfce9}\textbf{#1}}

\newlength\Size
\titleformat{name=\section}[block]
  {\centering\rmfamily\small}
  {}
  {0pt}
  {\colorsection}
\titlespacing*{\section}{0pt}{\baselineskip}{\baselineskip}

\makeatletter
\newcommand{\colorsection}[1]{%
  \colorbox{TITLES}{\parbox{\dimexpr\linewidth-2\fboxsep}{%
    \ifnum\value{section}>0
      \textbf{\thesection.\ #1}  
    \else
      \textbf{#1} 
    \fi
  }}}
\makeatother

\newcommand{\resetSectionNumbering}{%
  \setcounter{section}{0}
  \setcounter{subsection}{0} 
}
\def\BibTeX{{\rm B\kern-.05em{\sc i\kern-.025em b}\kern-.08em
    T\kern-.1667em\lower.7ex\hbox{E}\kern-.125emX}}
    
\tikzset{%
  folder/.pic={%
    \filldraw [draw=folderborder, top color=folderbg!50, bottom color=folderbg] (-1.05*\Size,0.2\Size+5pt) rectangle ++(.75*\Size,-0.2\Size-5pt);
    \filldraw [draw=folderborder, top color=folderbg!50, bottom color=folderbg] (-1.15*\Size,-\Size) rectangle (1.15*\Size,\Size);},
  file/.pic={%
    \filldraw [draw=folderborder, top color=folderbg!5, bottom color=folderbg!10] (-\Size,.4*\Size+5pt) coordinate (a) |- (\Size,-1.2*\Size) coordinate (b) -- ++(0,1.6*\Size) coordinate (c) -- ++(-5pt,5pt) coordinate (d) -- cycle (d) |- (c) ;},
}

\forestset{%
  declare autowrapped toks={pic me}{},
  pic dir tree/.style={%
    for tree={folder,font=\ttfamily,grow'=0,},
    before typesetting nodes={%
      for tree={edge label+/.option={pic me},},},
  },
  pic me set/.code n args=2{%
    \forestset{%
      #1/.style={inner xsep=2\Size,pic me={pic {#2}},}
    }
  },
  pic me set={directory}{folder},
  pic me set={file}{file},
}
    
\newenvironment{Tabular}[2][1]
  {\def\arraystretch{#1}\tabular{#2}}
  {\endtabular}
  
\newcommand{\FLAIRHUB}{FLAIR-HUB\xspace}% 
\newcommand{\false}[0]{\textcolor{red}{\ding{55}}}
\newcommand{\true}[0]{\textcolor{green}{\ding{52}}}

\begin{document}

\twocolumn[{
  \begin{@twocolumnfalse}
    \centering\underline{\makebox[16cm]{\large{FLAIR: French Land cover from Aerospace ImageRy.}}} \\
    \vspace{-0.38cm}
    \begin{figure}[H]
    \centering
    \includegraphics[width=2.052\linewidth]{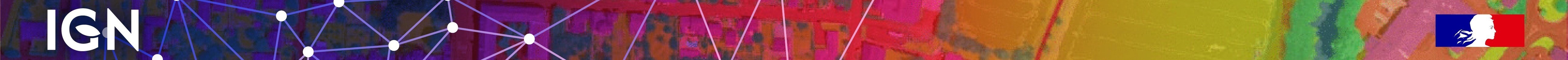}
    \end{figure}
    \centering
    \vspace{-0.3cm}
    \Large{FLAIR-HUB: Large-scale Multimodal Dataset for Land Cover and Crop Mapping} \\
    \vspace{+0.5cm}
    \normalsize{Anatol Garioud, Sébastien Giordano, Nicolas David, Nicolas Gonthier}    \\
    \vspace{+0.1cm}
    \normalsize{Institut national de l’information géographique et forestière (IGN), France} \\
    \vspace{+0.1cm}
    \centering\normalsize{\textit{flair@ign.fr}} \\                 
    \vspace{0.6cm}

    \begin{center}
      \includegraphics[width=0.87\textwidth]{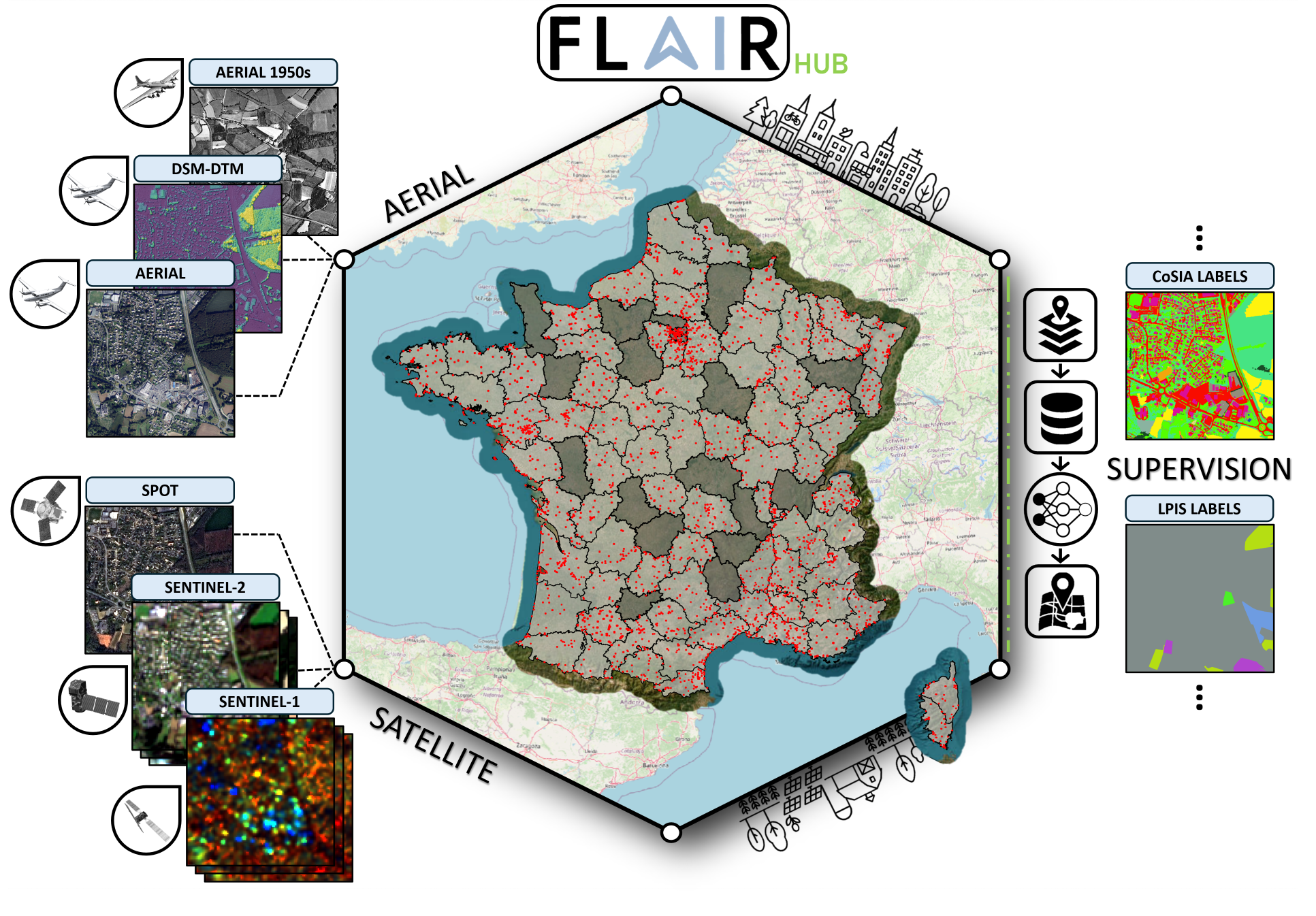}
    \end{center}

    \vspace{1.2cm}

\section*{Abstract}
\begin{justify}

    The growing availability of high-quality Earth Observation (EO) data enables accurate global land cover and crop type monitoring. However, the volume and heterogeneity of these datasets pose major processing and annotation challenges. To address this, the French National Institute of Geographical and Forest Information (IGN) is actively exploring innovative strategies to exploit diverse EO data, which require large annotated datasets. IGN introduces FLAIR-HUB, the largest multi-sensor land cover dataset with very-high-resolution (20\:cm) annotations, covering 2528\:km² of France. It combines six aligned modalities: aerial imagery, Sentinel-1/2 time series, SPOT imagery, topographic data, and historical aerial images. Extensive benchmarks evaluate multimodal fusion and deep learning models (CNNs, transformers) for land cover or crop mapping and also explore multi-task learning. Results underscore the complexity of multimodal fusion and fine-grained classification, with best land cover performance (78.2\% accuracy, 65.8\% mIoU) achieved using nearly all modalities. FLAIR-HUB supports supervised and multimodal pretraining, with data and code available at \href{https://ignf.github.io/FLAIR/FLAIR-HUB/flairhub}{https://ignf.github.io/FLAIR/FLAIR-HUB/flairhub}.

\end{justify}
  \end{@twocolumnfalse}
}]

\clearpage

\section*{Dataset overview}

\vspace{-0.5cm}
\begin{table}[h!]
\centering
\renewcommand{\arraystretch}{1.5}
\begin{Tabular}[1.5]{|p{8.4cm}|}
\hline \rowcolor[HTML]{C0C0C0} \textbf{Figures}\\  \hline
\end{Tabular}
\par\vskip1.2pt
\begin{Tabular}[0.8]{|p{8.4cm}|}
\hline
\rowcolor[HTML]{dbdbdb} \color{black}\ding{212} \color{black} 63\,202\,918\,400 pixels annotated at 0.20\:m spatial resolution \\
\rowcolor[HTML]{dbdbdb} \color{black}\ding{212} \color{black} 241\,100 patches (512$\times$512) \\
\rowcolor[HTML]{dbdbdb} \color{black}\ding{212} \color{black} 74 spatio-temporal domains and 2\,822 areas covering 2\,528\:km²\\
\rowcolor[HTML]{dbdbdb} \color{black}\ding{212} \color{black} 15 land cover semantic classes (+ 4 optional ones) \\[1em]
\rowcolor[HTML]{dbdbdb} \color{black}\ding{212} \color{black} 23/31/46 crop types in 3-level class hierarchy \\[1em] 
\rowcolor[HTML]{dbdbdb} \color{black}\ding{212} \color{black} 256\,221 Sentinel-2 acquisitions \\[1em] 
\rowcolor[HTML]{dbdbdb} \color{black}\ding{212} \color{black} 532\,696 Sentinel-1 acquisitions \\[1em]
\rowcolor[HTML]{dbdbdb} \color{black}\ding{212} \color{black} 1.6\:m SPOT images aligned \\[1em] 
\rowcolor[HTML]{dbdbdb} \color{black}\ding{212} \color{black} Aligned historical aerial images \\[1em]
\rowcolor[HTML]{dbdbdb} \color{black}\ding{212} \color{black} 20\:cm resolution DSM and DTM aligned \\[1em] \hline
\end{Tabular}
\end{table}

\vspace{-0.5cm}
\begin{table}[h!]
\centering
\renewcommand{\arraystretch}{1.5}
\begin{Tabular}[1.5]{|p{8.4cm}|}
\hline \rowcolor[HTML]{C0C0C0} \textbf{Structure}\\  \hline
\end{Tabular}
\par\vskip1.2pt
\hspace{0.02cm}
\begin{Tabular}[0.8]{|p{8.4cm}|}
\hline 
\rowcolor[HTML]{dbdbdb}  
{\footnotesize
\begin{forest}
  pic dir tree, where level=0{}{directory,},
  for tree={ s sep=0.1cm, l sep=1cm, font=\rmfamily }
  [\textbf{FLAIR-HUB dataset}
    [DOMAIN\_SENSOR\_DATATYPE [ROI [PATCH.tif, file]]]
    [GLOBAL\_ALL\_MTD [GLOBAL\_MODALITY\_*.gpkg, file]]
  ]
\end{forest}}\\ \hline
\end{Tabular}
\vspace{-4mm}
\end{table}

\section{Context}

In recent years, remote sensing and Earth Observation (EO) had a growing impact on many scientific fields and economic sectors. Extracting information about the Earth's surface from the sky or space is a key research area. This topic is involved in 11 of the 17 United Nations Sustainable Development Goals \cite{cfodds_transforming_2015,vinuesa_role_2020}. In particular, the automatic analysis of EO images plays an important role in mapping human activities and their impact on the environment. For example, it is useful for applying the European regulation on products derived from deforestation \cite{european_parliament_regulation_2023}, for achieving France's no net land take target \cite{LoiClimat,soilDegradationEU}, or to monitor soils degradation \cite{FAO}.                               

An increasing number of regional and national mapping agencies deployed image recognition models to monitor urbanisation, agricultural and forestry areas, risk prevention, and public policy \cite{garcia2022catlc,dlr_data_2021}. In this context the French National Institute of Geographical and Forest Information (IGN) \cite{IGN}, in response to the growing availability of high-quality EO data, is actively exploring innovative strategies to integrate these data with heterogeneous characteristics, especially to monitor land cover and crop type across the territory of France and provide reliable and up-to-date geographical reference datasets.

The \mbox{FLAIR \#1} dataset \cite{FLAIR-1}, which focused on aerial imagery for semantic segmentation, was released to facilitate research in the field. Building upon this dataset, the \mbox{FLAIR \#2} dataset \cite{FLAIR-2,garioud_flair_2023a} extends the capabilities by incorporating a new input modality, namely Sentinel-2 satellite image time series, and introduces a new test dataset. Both \mbox{FLAIR \#1} and \#2 datasets are part of the currently used by IGN to produce the French national land cover map reference \textit{Occupation du sol à grande échelle} (OCS-GE) \cite{ocsge}.  

In this paper, we introduce \textbf{\mbox{\FLAIRHUB}}, the increased version of the "French Land cover from Aerospace ImageRy" dataset, the largest multi-sensor land-cover dataset with very-high-resolution annotations. \FLAIRHUB combines very-high-resolution (VHR, $20$\:cm) images, photogrammetry-derived surface models, and optical Sentinel-2 and SAR Sentinel-1 multi-spectral satellite time series, high-resolution SPOT satellite images and historical analog aerial images from the 1950's.

These acquisitions' diverse spatial, spectral, and temporal resolutions offer valuable complementary perspectives for land cover and crop analysis. Over $63$ billion pixels have been hand-annotated by geospatial experts, using a nomenclature of $19$ land-cover classes and 23 crop type classes. The data spans 2\:528\:km$^2$ across French sub-regions featuring diverse bioclimatic attributes at various times of the year, thus displaying complex and challenging domain shifts. 

In addition to this new dataset, we provided an extensive evaluation of the gain made with this dataset on different experiments of multimodal fusion with a recent computer vision backbone \cite{liu2021swin}.

\mbox{FLAIR-HUB} combines heterogeneous and diverse data aiming to foster the development of new large-scale semantic segmentation methods. Given its scale and the complexity of the task it exhibits, it presents an exciting challenge for the machine learning communities. It is also an excellent dataset for multimodal self-supervised methods \cite{astruc2025omnisat,astruc2024anysat,xiong2024neural,stewart2024ssl4eo}  or data fusion methods \cite{garioud_flair_2023a,garnot2022multi,audebert2016semantic} thanks to the spatial alignment between modalities. It is also a dataset that will evolve with the addition of new aligned modalities (\textit{e.g.}, hyperspectral, LIDAR) or new annotations (\textit{e.g.}, hedgerows, land use classes).

\section{Related Work}

\begin{table*}[t]
\caption{\textbf{Land Cover Datasets.} Publicly available datasets for semantic segmentation of land cover using optical remote sensing imagery. Topo refers to topographic information such as DSM, DEM, or slope. $\ddagger$ : The FLAIR dataset is included in the new \FLAIRHUB dataset.\\}
\label{tab:SOA_landcover}
\centering
\scriptsize
\renewcommand{\arraystretch}{1.5}
\setlength{\tabcolsep}{6.5pt}
\begin{tabular}{lllllclll}
\toprule
\multirow{2}{*}{Dataset} & \multicolumn{4}{c}{Land Cover Annotation} & & \multicolumn{3}{c}{Acquisition} \\
\cline{2-5}\cline{7-9}
\addlinespace[2pt] 
& Pixels $\times 10^6$ & Resolution & Classes & Source & & Resolution & Extent (km$^2$) & Source  \\
\midrule
\midrule
Vaihingen \citep{vaihingen} & 82 & 8\:cm & 6 & visual interpretation && 8\:cm & 1 & aerial\\
\rowcolor{black!5} EuroSAT \citep{helber2019eurosat} & 110 & 50\:m & 10 & EU Urban Atlas \citep{UrbanAtlas} && 10\:m & 11\,059 & Sentinel-2\\
MultiSenGE \citep{MULTISENGE} & 534 & 10\:m & 14 & visual interpretation && 10\:m & 57\,433 & Sentinel-1\&-2  \\
\rowcolor{black!5} Landcovernet \citep{alemohammad2020landcovernet} & 589 & 10\:m & 7 & semi-automatic (MODIS \citep{modis}) && 10\:m & 58\,982 & Sentinel-2\\  
MiniFrance \citep{minifrance} & 1\,510 & 50\:m & 14 & EU Urban Atlas \citep{UrbanAtlas}  && 50\:cm & 53\,000 & aerial \\ 
\rowcolor{black!5} DynamicEarthNet \citep{toker2022dynamicearthnet} & 1\,889 & 3\:m & 7 & visual interpretation && 3\:m & 16\,986 & Sentinel-1\&-2, PlanetFusion\\
LoopNet \citep{Boonpook_ijgi12010014} & 3\,133 & 30 m & 6 & visual interpretation && 30\:m & 2\,820k & Landsat 8 \\
\rowcolor{black!5} OpenEarthMap \citep{xia2023openearthmap} & 4\,931 & 25–50\:cm & 8 & visual interpretation && 25–50\:cm & 799 & aerial, UAV,, satellite \\
Five-Billion-Pixels \citep{tong2023enabling} & 5\,000 & 4\:m & 24 & visual interpretation && 4\:m & 50\,000 & Gaofen-2\\
\rowcolor{black!5} LoveDA \citep{wang2021loveda} &  6\,000 & 30\:cm & 7 & visual interpretation && 30\:cm & 536 & aerial \\ 
Dynamic World  \citep{brown2022dynamic} & 6\,348 & 50m & 9 & semi-automatic / iteratif && 10\:m & 634k & Sentinel-2  \\
\rowcolor{black!5} DeepGlobe \citep{demir2018deepglobe} & 6\,867 & 50\:cm & 7 & visual interpretation && 50\:cm & 1\,717 & Wordlview-2/3, GeoEye-1 \\
BigEarthNet \citep{sumbul2021bigearthnet} & 8\,500 & 100\:m & 19 & semi-automatic (CLC \citep{cover2000corine}) && 10\:m & 850\:k & Sentinel-1\&-2 \\
\rowcolor{black!5} FLAIR$^{\ddagger}$ \cite{garioud_flair_2023a}  & 20\,385 & 20\:cm & 19 & visual interpretation && 20\:cm\,/\,10\:m & 817 & aerial, Topo, Sentinel-2\\
CatLC \citep{garcia2022catlc} & 25\,600 & 1 m & 41 & visual interpretation && 1\:m &  25\,600 & aerial, Sentinel-1\&-2, Topo\\
\rowcolor{black!5} SeasoNet \cite{kossmann2022seasonet} & 63\,353 & 10 m & 33 & semi-automatic (CLC \citep{cover2000corine}) && 10\:m & 748k & Sentinel-2 \\  % 
\midrule
\textbf{\FLAIRHUB (Ours)} & \textbf{63\,203} & \textbf{20\:cm} & \textbf{19/23} & \textbf{visual interpretation} && \textbf{\makecell[l]{20\:cm\\1.6\,/\,10\:m}} & \textbf{2\,528} & \textbf{ \makecell[l]{aerial, Topo, SPOT6-7\\ Sentinel-1\&-2}}\\
\bottomrule
\end{tabular}
\end{table*}

\subsection{Semantic Segmentation for remote sensing imagery}

Land cover and crop type mapping can be translated into multi-classes pixel classification, also known as semantic segmentation.
For almost a decade, deep learning is the de facto solution for semantic segmentation task \cite{mo2022review}, also in the case of remote sensing imagery \cite{yuan2021review}. The communities have created more efficient models based on artificial intelligence from FCN \cite{long2015fully} to UNet \cite{U-Net}, DeepLab \cite{chen2018encoder}, Vision Transformer \cite{dosovitskiy2021an} and Swin Transformer \cite{liu2021swin}. The objective was to find a way to train bigger and better models by adding more long-distance dependency in the inner features and more parameters. This has been done by adding attention mechanics and better optimization schemes.

\subsection{Land cover datasets}

Numerous land-cover datasets have been introduced to train semantic segmentation methods, see Table~\ref{tab:SOA_landcover}. Existing datasets usually present a trade-off: they either offer high-resolution annotations but cover a small extent (like Vaihingen \citep{vaihingen}), or provide large-extent coverage but with low-resolution annotations (such as BigEarthNet \citep{sumbul2021bigearthnet}, SeasoNet \cite{kossmann2022seasonet} or SEN12MS \citep{SEN12MS}), some of them are even the output of fully automatic process \cite{zhang2022urbanwatch,SEN12MS}. In contrast, FLAIR \cite{garioud_flair_2023a} and \FLAIRHUB~ offer very high-resolution annotations (20\:cm) and covers a large portion of the French territory.
\FLAIRHUB is equivalent to SEN12MS \citep{SEN12MS} in number of pixels but it provides high-quality very high resolution annotation compared to automatic annotation at 10m ground sampling distance.

\FLAIRHUB comprises over $63$ billion manually annotated pixels, which is more than $3$ times more than FLAIR \cite{garioud_flair_2023a} or CatLC \cite{garcia2022catlc},  the closest counterparts to our dataset. 

The spatial resolution of the annotation is crucial in land-cover analysis. Insufficient resolution prevents the precise measurements of surfaces and boundaries. Furthermore, small-scale features, such as individual houses, lone trees or roads, may not be captured accurately, limiting the potential applications of the derived segmentation. This new dataset tends to answer some of the current challenges about operational very high resolution land cover \cite{mallet_current_2020} that are spatial and semantic accuracies, upscaling and data fusion. 

\begin{table*}[b]
\caption{\textbf{Agricultural Land Datasets.} Publicly available datasets for monitoring agricultural land. \textit{Multi-year | Multi-temporal} : A multi-year dataset includes data from different years over different areas, whereas a multi-temporal dataset provides data from multiple years or times for each area. \textit{TS | SITS} : SITS (Satellite Image Time Series) indicates instance or segmentation datasets that include one image per timestamp in the time series. TS (Time Series) refers to classification datasets using tabular time-series data, where each time series represents values of spectral indices (\textit{e.g.}, NDVI) aggregated at the object (\textit{e.g.}, parcel) level. }
\label{tab:SOA_LPIS}
\centering
\scriptsize
\renewcommand{\arraystretch}{1.2}
\setlength{\tabcolsep}{3.6pt}
\begin{tabular}{llllp{1mm}lllp{1mm}lllp{1mm}l}
\toprule
\multirow{2}{*}{Type / Dataset} & \multicolumn{3}{c}{Data} & & \multicolumn{3}{c}{RoI} & & \multicolumn{3}{c}{Annotation} & & \multirow{2}{*}{Size} \\
\cline{2-4}\cline{6-8}\cline{10-12}
\addlinespace[2pt] 
& Source & Patch Size & \#Samples & & Extend & Areas  & temporality & & Parcels & Classes & Source \\
\midrule
\midrule
SITS / Crop type & & & & & & & & & & & & &\\
\midrule
MunichCrops \cite{russwurm2018multi} & Sentinel-2 & 48$\times$48 & - & & \makecell[l]{4\,284\:km$^{2}$ \\ Munich} & 1 &  \makecell[l]{multi-year\\ 2016-2017}& & 137\:k & 17 & LPIS & & 42\:Go \\
\rowcolor{black!5} \Gape[0pt][2pt]{\makecell[l]{Crop Type \\Mapping Ghana \cite{m2019semantic}}} & \Gape[0pt][2pt]{\makecell[l]{Sentinel-1\\Sentinel-2\\Planets}}& 32$\times$32 & - & & \Gape[0pt][2pt]{\makecell[l]{ - \\ Ghana}} & - & 2016 &  &  8\,937  & 4-24 & Survey & & 310\;Go\\
CV4A Kenya \cite{kerner2020field} & Sentinel-2 & 2016$\times$3035 & 4 & &  \makecell[l]{ $\sim$2\,450\:km$^{2}$\\Western Kenya} & 4 & 2019 & & $>$3\,000 & 7 & Survey & & 3.5\:Go \\
\rowcolor{black!5} ZueriCrops \cite{turkoglu2021crop} & Sentinel-2 & 24$\times$24 & 28\:k &  & \Gape[0pt][2pt]{\makecell[l]{2400\:km$^{2}$\\Zurich - Swiss}} & 1 & 2019 & & 116\:k & 5-14-48 & \Gape[0pt][2pt]{\makecell[l]{LPIS\\FOAG}}  & & - \\
Pastis-R \cite{garnot2021panoptic} &  \makecell[l]{Sentinel-1\\Sentinel-2}& 128$\times$128  & 2433 & & \makecell[l]{ $\sim$4\,000\:km$^{2}$\\France} & 4 & 2019 & & 124\:k & 18 & \makecell[l]{LPIS\\FR} &  & 54\:Go \\
\rowcolor{black!5} Sen4AgriNet \cite{sykas2021sen4agrinet} & Sentinel-2 &  366$\times$366 & 225\:k & & \Gape[0pt][2pt]{\makecell[l]{All France\\Catalonia}} & 1 & \Gape[0pt][2pt]{\makecell[l]{multi-year\\2016-2020}} & & 42\:M & 9-158 &\Gape[0pt][2pt]{\makecell[l]{LPIS\\FR-Ca}}  & & 10\:To \\
DENETHOR \cite{kondmann2021denethor} & \makecell[l]{Sentinel-1\\Sentinel-2\\Planets} & -  & - & & \makecell[l]{ 1\,152\:km$^{2}$\\Germany} & 2 & \makecell[l]{multi-temporal\\2018-2019} & & 4\,500 &  9 & LPIS & & 254\:Go \\
\rowcolor{black!5} \Gape[0pt][2pt]{\makecell[l]{AgriSen-COG \cite{selea2023agrisen}\\TS-SITS*}} & Sentinel-2 & 366$\times$366 & 41\,000 &  &  \Gape[0pt][2pt]{\makecell[l]{ - \\5 Country}}  & 5 & \makecell[l]{multi-temporal\\2019-2020}  & & $\sim$7\:M & 11-*  & LPIS & & \Gape[0pt][2pt]{\makecell[l]{28\:Go\\(Label)}}\\
\midrule
TS / Crop type & & & & & & & & & & & & & \\
\midrule
BreizhCrops \cite{russwurm2019breizhcrops} & \makecell[l]{Sentinel-2\\L2A - L1C} & - & 610k & & \makecell[l]{27200\:km$^{2}$ \\Britanny France} & 1 & 2017 & & 610\:k & 9 & \makecell[l]{LPIS\\FR} & & \makecell[l]{3.2\:Go\\ 8.5\:Go} \\
\rowcolor{black!5} TimeSen2Crop \cite{weikmann2021timesen2crop} & Sentinel-2 & - & 1\,100\:k& & \Gape[0pt][2pt]{\makecell[l]{84k\:km$^{2}$\\Austria}} & 1 & \Gape[0pt][2pt]{\makecell[l]{multi-temporal\\2018-2019}} &  & 1\,100\:k & 16& LPIS & & 1.2\:Go \\
\makecell[l]{CropDeepTrans \cite{barriere2024boosting}\\(early crop)} & \Gape[0pt][2pt]{\makecell[l]{Sentinel-2\\Crop Rot}}  & - & $\sim$7.6\:M & & \Gape[0pt][2pt]{\makecell[l]{270k\:km2 \\FR-NL}} & 2 &  \Gape[0pt][2pt]{\makecell[l]{multi-temporal\\2016-2020}} & & 7.65\:M & 24|32 &  \Gape[0pt][2pt]{\makecell[l]{LPIS\\FR-NL}} & & - \\
\rowcolor{black!5} Sen4Map \cite{sharma2024sen4map} & \Gape[0pt][2pt]{\makecell[l]{Sentinel-2}}  & 64$\times$64 & 335\:k & & Europe & 335\:k &  2018 & & \Gape[0pt][2pt]{\makecell[l]{20 LC\\$\sim$35  Crops}} & Stat Survey  & LUCA & & 1.2\:To \\
\makecell[l]{EuroCropsML\cite{reuss2024eurocropsml}\\(few shot)}& Sentinel-2  & - & 707\:k  &  & \makecell[l]{3 EU Countries\\EE-LV-PT}  & 2 & 2021 & & 707\:k & 176  & \makecell[l]{LPIS\\EuroCrops} & & 4.7\:Go\\
\midrule
Parcel delineation & & & & & & & & & & & & & \\
\midrule
\rowcolor{black!5} AI4SmallFarms \cite{persello2023ai4smallfarms}  & Sentinel-2 & 1\,000$\times$1\,000 & 62 & &\Gape[0pt][2pt]{\makecell[l]{Vietnam\\Cambodia}}   & 62 &t & & 439\:k &- & \Gape[0pt][2pt]{\makecell[l]{image\\labelling}}  & & 1.4\:Go\\
AI4Boundaries \cite{d2023ai4boundaries} & \makecell[l]{Aerial Ortho(1\:m)\\Sentinel-2} & \makecell[l]{512$\times$512\\256$\times$256} & 7\,831 & & \makecell[l]{Europe\\7 countries} & 7\,831  &  \makecell[l]{2019\\S2-composite} & &14.8\:M  & - & LPIS &  & 38\:Go\\
\rowcolor{black!5} Fields of the World \cite{kerner2024fields}  & \Gape[0pt][2pt]{\makecell[l]{Sentinel-2\\multi-date}} & 256$\times$256 & 70.5\:k & & \Gape[0pt][2pt]{\makecell[l]{166\:k\:km$^{2}$\\4 Continents\\24 countries}} & $\sim$80 & \Gape[0pt][2pt]{\makecell[l]{multi-Year\\S2 - 1 date}}& & 1.63\:M & - & \Gape[0pt][2pt]{\makecell[l]{FIBOA\\EuroCrops}} & & - \\

\bottomrule
\end{tabular}
\end{table*}

\begin{table*}[b]
\caption{\textbf{Multi modal Datasets.} Publicly available datasets featuring aligned multi-modal data for Earth observation, with annotations and more than three modalities.
Pixel counts are based on the highest-resolution modality.
SITS (Satellite Image Time Series). \; S1\/2: Sentinel-1\/2.\;  Topo: topographic information such as DSM, DEM, or slope.\; HS: hyperspectral imagery.\; VHR: aerial very high-resolution imagery.\; Historical: legacy VHR imagery.\; LU: land use.\; LC: land cover. $\ddagger$ : The FLAIR dataset is included in the new \FLAIRHUB dataset.}
\label{tab:SOA_MultiModal}
\centering
\scriptsize
\renewcommand{\arraystretch}{1.3}
\setlength{\tabcolsep}{3.5pt}
\scriptsize{
    \begin{tabular}{lcccccccccccclr}
    \toprule
        \multirow{2}{*}{Dataset} & \multirow{2}{*}{\begin{tabular}{c} Number of \\ Modalities \end{tabular}} & \multicolumn{11}{c}{Modality} & \multirow{2}{*}{\begin{tabular}{c}Task\end{tabular}} & \multirow{2}{*}{\begin{tabular}{r}  Pixels \\ $\times 10^9$ \end{tabular}} \\
        \cline{3-13}
        & & VHR & S1 SITS & S2 SITS & SPOT & Landsat t.s. & Topo & ALOS & MODIS SITS & HS & LIDAR &  Historical &  &  \\
        \hline
        DFC 2018 \cite{xu2019advanced} & 3 & \checkmark &  &  &  &  &  &  &  & \checkmark & \checkmark  & & \makecell[l]{LULC \\ Segmentation} & 2.0 \\ %%% Il manque une 
   \rowcolor{black!5} TSAI-TS \cite{astruc2025omnisat,ahlswede2022treesatai} & 3 & &  \checkmark &  \checkmark &  \checkmark &    &  &  &  & & & & \Gape[0pt][2pt]{\makecell[l]{Tree Species \\ Classification}}  & 4.7  \\
    FLAIR$^{\ddagger}$ \cite{garioud_flair_2023a} & 3 & \checkmark  &  & \checkmark &  &  & \checkmark &  &  & & &  & \makecell[l]{LC \\ Segmentation}  &  20  \\
      \rowcolor{black!5}   TalloS \cite{bountos2025fomo} & 3 &  & \checkmark  & \checkmark  &  &  & \checkmark &  &  & & &  & \Gape[0pt][2pt]{\makecell[l]{Tree Species \\ Classification}}  &  0.16  \\
        PASTIS-HD \cite{astruc2025omnisat,garnot2021panoptic}  & 3 &  & \checkmark  & \checkmark & \checkmark &  &  &  &  & & & & \makecell[l]{Crop Type \\ Segmentation} &  7.5  \\
       \rowcolor{black!5}   Neon Trees \cite{weinstein2020benchmark}  & 3 &  & \checkmark  & & &  &  &  &  &  \checkmark & \checkmark  & &  \Gape[0pt][2pt]{\makecell[l]{Tree \\ Detection}} &  363  \\
         CatLC \citep{garcia2022catlc} & 4 & \checkmark & \checkmark & \checkmark &  &  & \checkmark &  &  & & & & \makecell[l]{LC \\ Segmentation}  &  25.6  \\
        \rowcolor{black!5}  S2NAIP \cite{s2naip,satlassuperres} & 4 &  \checkmark  &  \checkmark &  \checkmark &  &  \checkmark &  &  & & & &  & \Gape[0pt][2pt]{\makecell[l]{"World Cover" \\Segmentation}} & 136    \\
        SatlasPretrain \cite{bastani2023satlaspretrain} & 4 &  \checkmark  &  \checkmark &  \checkmark &  &  \checkmark &  &  & & & & & \makecell[l]{"World Cover" \\Segmentation} & 3\,087    \\
       \rowcolor{black!5}  MADS \cite{hu2023mdas}  & 5  &  & \checkmark & \checkmark &  &  & \checkmark & &  & \checkmark \checkmark &  &  &  \Gape[0pt][2pt]{\makecell[l]{LC \\ Segmentation}} & 1.9 \\ 
        Planted \cite{pazos2024planted}  & 5 & \checkmark & \checkmark  &  & & \checkmark &  & \checkmark & \checkmark &  & & & \makecell[l]{Tree \\ Classification} &  3.0  \\
        \midrule
       \bf \FLAIRHUB (Ours) & 6 & \checkmark & \checkmark & \checkmark   & \checkmark &  &  \checkmark  & &  & & & \checkmark  & \makecell[l]{LC \& Crop Type \\ Segmentation} &  63  \\
        \bottomrule
    \end{tabular}
}
\end{table*}

\subsection{Crop Type datasets}

Monitoring agricultural land cover using Earth Observation involves several key tasks and targets, including parcel delineation, crop type classification or segmentation. 

Parcel delineation is similar to an image segmentation task, typically involving two or three classes (\textit{e.g.}, boundary, interior, exterior). This task is especially critical in countries lacking a Land Parcel Information System (LPIS) \cite{kerner2020field}. Datasets for parcel delineation rely on two main sources of labels: existing open-data parcel databases such as LPIS \cite{d2023ai4boundaries}, and manually annotated parcel boundaries \cite{persello2023ai4smallfarms}. Most parcel delineation datasets are based on Sentinel-2 imagery at 10\:m spatial resolution. Some also incorporate PlanetScope data at 3\:m spatial resolution, while AI4Boundaries \cite{d2023ai4boundaries} additionally includes aerial imagery at 1\:m spatial resolution, though this higher-resolution data is not available for all years. Since parcel delineation is particularly valuable in areas where labelled data is sparse, recent datasets aim to combine both manual and declarative sources to leverage their respective advantages. However, due to the limited availability of manually labelled data compared to LPIS, careful sampling is required to ensure geographic balance in the dataset \cite{kerner2024fields}.

Crop type classification is typically framed as a time series classification task and was one of the earliest approaches to crop mapping. The input time series can be constructed either at the pixel level or at the object level (i.e., parcels), when parcel boundaries are available. The development of crop type classification datasets has been significantly accelerated by the open availability of Satellite Image Time Series (SITS), such as Sentinel-1 and Sentinel-2, as well as parcel vector data like the LPIS in the European Union \cite{russwurm2019breizhcrops, weikmann2021timesen2crop, reuss2024eurocropsml}. The large sample sizes enabled by these data sources have made it possible to apply deep learning architectures effectively to the crop type classification task. Since time series classification does not require dense pixel-wise annotations, labels can also be derived from statistical surveys, such as the LUCAS Survey in the EU \cite{d2020harmonised}.

Crop type segmentation extends pixel-level time series classification by incorporating spatio-temporal elements into the model. Although Sentinel-2 imagery has a spatial resolution of 10\:m, frequent acquisitions and rich spectral bands produce datasets that are often one to two orders of magnitude larger than those used for standard time series classification. The availability of extensive label data and satellite imagery has enabled the creation of very large datasets. For instance, the dataset in \cite{sykas2021sen4agrinet} exceeds 10\:TB in size, as it includes all available data with minimal curation \cite{weikmann2021timesen2crop, russwurm2019breizhcrops}. However, as suggested by Roscher et al. \cite{roscher_better_2024}, using a curated subset of the data can improve usability, achieve better class balance, and maintain comparable classification performance.

Across all dataset types shown in \Cref{tab:SOA_LPIS}, a key challenge in their creation lies in homogenizing existing parcel and crop-type databases. Crop classification schemes are often country- or region-specific, making direct integration into a unified dataset difficult. To address this, classifications must be standardized before datasets from different sources can be used together effectively. For smaller datasets, this harmonization is typically done manually by researchers. The resulting crop taxonomy is often a simplified version—either a subset of the original crop types or a grouping based on agronomic similarities. This issue has prompted specific efforts focused on data harmonization. For example, the FIBOA project (Field Boundaries for Agriculture) provides standardized parcel delineation and is used by the Fields of the World dataset \cite{kerner2024fields}. Similarly, the EuroCrops project \cite{schneider2023eurocrops} has harmonized LPIS data across the European Union and has served as a source of standardized crop labels in recent datasets \cite{reuss2024eurocropsml, barriere2024boosting}. EuroCrops introduced the HCAT nomenclature, developed in connection with the EAGLE matrix \cite{arnold2018eagle} by the European Environment Agency. Other datasets, such as \cite{sykas2021sen4agrinet} and \cite{selea2023agrisen}, use crop classifications derived from the Indicative Crop Classification (ICC) developed by the FAO \cite{ICC}. In addition to LPIS, some transnational harmonized sources like the LUCAS statistical survey \cite{d2020harmonised} have been employed, notably in the Sen4Map dataset \cite{sharma2024sen4map}. An alternative to manual or rule-based harmonization is the development of foundation models with few-shot learning capabilities, which could adapt to different national classification systems without requiring extensive retraining \cite{reuss2024eurocropsml}.

Another critical challenge stems from the temporal variability of crop types, which introduces significant transfer learning issues. A model trained on a specific year and region may perform well within its domain but generalize poorly across different years—even within the same area. This limitation has motivated the creation of adapted crop type segmentation datasets. Two strategies have emerged to address this issue. Multiyear datasets contain data from only one year per Region of Interest (ROI), but the year varies across ROIs \cite{sykas2021sen4agrinet}. In contrast, multi-temporal datasets include data from multiple years for each ROI, supporting more robust temporal generalization \cite{barriere2024boosting, selea2023agrisen}. Finally, a particularly relevant variant of the temporal generalization problem is early crop classification, where the objective is to identify crop types as early as possible in the growing season, rather than retrospectively at the end of the year. In this context, historical crop type information, \textit{i.e.}, the crop grown on the parcel in the previous season—can provide valuable prior knowledge, as crop rotations are rarely random and often follow agronomic patterns.

\subsection{Multimodal remote sensing imagery datasets}

Multimodal datasets are valuable resources for developing models that effectively integrate diverse remote sensing modalities, each contributing complementary information. Designing architectures that can fully exploit the specific strengths of different sensors remains an open research challenge \cite{wang2024lmfnet, li2022deep, robinson2021global, xu2019advanced}.

Beyond model design, multimodal datasets play a crucial role in self-supervised pretraining of deep learning models \cite{astruc2024anysat, jakubik2023foundation, astruc2024omnisat, fuller2023croma, tseng2023lightweight}, as well as in thematic applications such as forest monitoring \cite{bountos2025fomo} and super-resolution \cite{satlassuperres}. An increasing number of datasets have become available for pretraining purposes, typically without annotations and featuring uniform global coverage. These are often based on observations from single-sensor satellite constellations like Landsat and Sentinel \cite{stewart2024ssl4eo, bastani2023satlaspretrain}. However, truly multimodal datasets—involving multiple sensors—are still relatively rare \cite{francis2024major, marchisio2021rapidai4eo, mall2024remote, velazquez2025earthview, wang2023ssl4eo, li2024seamo, guo2024skysense}, and only a few of them include ground-truth annotations \cite{mendieta2023towards, xia2025openearthmap}.

Some multimodal datasets provide data from different sensors over non-overlapping regions, resulting in no spatial alignment across modalities \cite{mendieta2023towards}, while others rely on automatically generated labels \cite{marchisio2021rapidai4eo}. Despite these limitations, multimodal datasets can also be leveraged in cross-modal supervision setups, where one modality is used to predict another. For example, Sentinel-2 and PALSAR-2 data can be used to estimate biomass, supervised by GEDI measurements \cite{sialelli2024agbd}, or to perform cloud removal tasks \cite{UnCRtainTS}.

As argued by Roscher et al. \cite{roscher_better_2024}, Earth observation models benefit more from high-quality, diverse, and well-curated datasets than from massive but uniform data acquisitions. The design of \FLAIRHUB is aligned with these findings, offering a dataset that combines extensive modal diversity and large-scale coverage with curated, high-quality annotations.

\Cref{tab:SOA_MultiModal} presents a comparison of multimodal datasets that include at least three spatially aligned modalities and are accompanied by some form of ground truth annotation. Our dataset includes the highest number of aligned modalities, along with a large number of annotated pixels. This is especially notable given that several other datasets rely on automatic or uncurated label sources \cite{bastani2023satlaspretrain, s2naip}. Some datasets, such as FLAIR \cite{garioud_flair_2023a}, PASTIS-HD \cite{astruc2025omnisat}, or Neon Trees \cite{weinstein2020benchmark}, include only three modalities. Others, such as MADS \cite{hu2023mdas} and Planted \cite{pazos2024planted}, do offer five aligned modalities but include fewer annotated pixels than \FLAIRHUB.

To our knowledge, \FLAIRHUB is the only dataset that combines historical aerial very high-resolution (VHR) imagery, SAR, and multispectral time series in a fully aligned setup. Though it does not include LiDAR and hyperspectral data, these modalities remain far less available in open-access, large-scale, well-annotated formats.

\section{Dataset global description and naming conventions}

The first level of the \FLAIRHUB dataset  folder structure is defined by the \textit{modalities per domain} information. This organization was chosen to facilitate the later addition of new domains, modalities or supervisions. More specifically the fist level folder name is as follows : \textit{DOMAIN\_SENSOR\_DATATYPE}.\\

\noindent\textbullet\enspace\textbf{DOMAIN}: The \FLAIRHUB dataset is composed of 74 spatio-temporal domains. We define a spatio-temporal domain as the conjunction of a geographical-extent and an acquisition date. The geographical extent corresponds to the French departments ($\approx$\:100 departments), which are both an administrative subdivision of the territory and the management unit for very high spatial resolution aerial images acquisitions. When aerial surveys and the resulting orthoimages are managed together for two neighbouring departments, we assemble the two department identifiers in the domain's name (\textit{e.g.}, \textit{D059062-2021}, corresponds to departments 59 and 62).\\

\noindent\textbullet\enspace\textbf{SENSOR\_DATATYPE}: At the first level of the directory structure, the modality information is organized into two types: (i) image sources (7 folders) and (ii) supervisions (3 folders). Each domain therefore has on release 10 folders. For the image sources the \textit{SENSOR} part relates either to the name of the sensors that acquired the images  (SENTINEL1-ASC, SENTINEL1-DESC, SENTINEL2, SPOT) or products derived from aerial surveys (AERIAL-RLT, AERIAL, DEM). The \textit{DATATYPE} gives a hint about the data format (TS: Times Series; PAN: Panchromatic; RGBI: Red, Green, Blue, Infrared channels; ELEV: Elevation). For supervisions, the value of the \textit{SENSOR} field is specified to differentiate between supervisions that have a strong link with a specific sensor (for example where annotation was performed on the sensor's images) and those that can possibly be applied to all sensor images (in which case, the value \textit{ALL} is used). The \textit{DATATYPE} is used to exhibit the name of the supervision. In this name we distinguish between \textit{LABEL} if it corresponds to an external annotation (LABEL-COSIA, LABEL-LPIS) and MSK (masks) if it is a layer of information derived from a sensor data such as the SENTINEL2 snow and cloud mask (SENTINEL2\_MSK-SC). A directory containing the metadata of all patches from all domains \textit{GLOBAL\_ALL\_MTD} is also available. The format of each metadata file is a GeoPackage, easily readable with GeoPandas Python package containing both the geometric and attribute information of the patches. Metadata about acquisition dates, geometry or radiometry statistics are provided in the \textit{GLOBAL\_ALL\_MTD} folder. 

The second level of the directory structure represents the Regions Of Interest (ROI). A ROI is a set of contiguous patches. In \FLAIRHUB, all the patches are spatially aligned, meaning that regardless the modality and its spatial resolution, they represent the same coverage on the ground : 102.4\:m$\times$102.4\:m. This choice has been made to have patches of 512$\times$512 pixels at 0.2\:m spatial resolution. The concept of ROIs was introduced to facilitate the annotation process by pooling the labelling cost. \FLAIRHUB dataset is composed of 2\,822 ROIs. While patches have fixed sizes on the ground, ROIs have variable sizes. To ensure a perfect nesting of the patches in the ROIs, their width and length are multiples of 512\:m. On average, a ROI covers 0.90\:km\textsuperscript{2}. Table~\ref{tab:ROI_sizes} provides information on the distribution of ROI sizes. The ROI names (name of the second level folder) is not unique. The ROI identifier corresponds to internal management information that is not important for the dataset, except for the first two characters, which describe the primary and secondary land use (A=Agricultural, F=Forest, N=Natural, U=Urban).

\begin{table}[!htpb]
\caption{{\bf Region of Interest (ROI) Size Distribution.} The area are $512\times512\:m$ multiples.}
\label{tab:ROI_sizes}
\centering
\scriptsize
\setlength{\tabcolsep}{3.5pt}
\begin{tabular}{l|cccccccccccc}
\toprule
\textbf{\makecell[l]{ROI Size\\($512\times512\:m$) $\times$}}  & 1  & 2  & 3  & 4  & 5  & 6  & 8  & 9  & 10  & 12  & 15  & 16  \\
\midrule
\rowcolor{black!5} \textbf{Occurence} & 499 & 496 & 260 & 1\,279 & 6 & 180 & 25 & 18 & 10 & 19 & 23 & 7 \\
\bottomrule
\end{tabular}
\end{table}

The patches are finally available in each ROI folder at third level. Their file naming system was designed to be unique and to contain the maximum amount of information about their type. The file naming convention is structured as follows : \textit{DOMAIN\_SENSOR\_DATATYPE\_ROI\_POSITION}. The \textit{POSITION} information give the relative position of the patch in the ROI (\textit{row-column}). This name is referred as the \textit{patch\_id} in the metadata files. The \textit{GLOBAL\_ALL\_MTD\_GEOM} metadata file provides : the spatial footprint of each patch (\textit{patch\_xmin}, \textit{patch\_ymin}, \textit{patch\_xmax}, \textit{patch\_ymax}) in the reference cartographic system of France \mbox{RF93-Lambert93} (EPSG:2154), the identifier and spatial extent of the ROI to which the patch belongs (\textit{ROI\_id}, \textit{ROI\_xmin}, \textit{ROI\_ymin}, \textit{ROI\_xmax}, \textit{ROI\_ymax}), the position of the patch in the ROI (\textit{patch\_row}, \textit{patch\_col} and the size of the ROI in terms of number of patches (\textit{ROI\_nbpatch}, \textit{ROI\_nbpatchx}, \textit{ROI\_nbpatchy}). The \textit{GLOBAL\_ALL\_MTD\_MODALITIES} is a useful metadata for the patches. It provides, for each patch, the relative path to the patch file of each modality.\\

\begin{table*}[b]
\caption{\bf{ Overview of the different data modalities available across the dataset.} We provide the details about spatial, temporal, spectral, and radiometric resolution for each modality.}
\label{tab:MOD_overview}
\centering
\scriptsize
\renewcommand{\arraystretch}{1.5}
\setlength{\tabcolsep}{3pt}
\begin{tabular}{lrrrrccrcr}
\toprule
Modality &\multicolumn{2}{c}{\textbf{Spatial resolution}} & \makecell[c]{\multirow{2}{*}{\textbf{\makecell[c]{Temporal\\ resolution}}}}& \multicolumn{4}{c}{\textbf{Spectral Resolution}} &   \multirow{2}{*}{\textbf{\makecell[c]{Radiometric\\ resolution}}} & \textbf{Volume}\\
\cline{2-3} \cline{5-8}
 &  Patch size & Pixel size  &   & type & channel per date & \multicolumn{2}{c}{Calibration} &  & \\
\midrule
AERIAL-RLT\_PAN      & 256$\times$256 & 0.4\:m & Mono-temporal  & Panchromatic & 1 & Relative & Domain equalization (DN) & UInt8 & 11.15 Go\\
AERIAL\_RGBI         & 512$\times$512 & 0.2\:m & Mono-temporal & Multi-spectral & 4 & Relative & Domain equalization (DN) & UInt8 & 232.76 Go\\
DEM\_ELEV           & 512$\times$512 & 0.2\:m & Mono-temporal & Multi-channel & 2 & Absolute & Altitude (m)  & Float32  & 365.18 Go\\
SENTINEL1-ASC\_TS     & 10$\times$10 & 10.24\:m & Time Series & Multi-channel & 2 & Absolute & $\sigma_{0}$ Backscatter (no unit)  & Float32  & 16.68 Go\\
SENTINEL1-DESC\_TS     & 10$\times$10 & 10.24\:m & Time Series & Multi-channel & 2 & Absolute & $\sigma_{0}$ Backscatter (no unit)  & Float32  & 17.96 Go\\
SENTINEL2\_TS       & 10$\times$10 & 10.24\:m & Time Series & Multi-spectral & 10 & Absolute & BOA Reflectance (\%)  & UInt16  & 41.61 Go\\
SPOT\_RGBI          & 64$\times$64 & 1.60\:m & Mono-temporal & Multi-spectral & 4 & Absolute & BOA Reflectance (\%)  & UInt16 &  6.24 Go\\
\midrule
AERIAL\_LABEL-COSIA & 512$\times$512 & 0.2\:m & Mono-temporal & Mono-channel & 1 & Absolute & Label & UInt8 & 2.20 Go\\
ALL\_LABEL-LPIS    & 512$\times$512 & 0.2\:m & Mono-temporal & Multi-channel & 3 & Absolute & Label & UInt8 & 6.35 Go\\
SENTINEL2\_MSK-SC  & 10$\times$10 & 10.24\:m & Time Series & Multi-channel & 2 & Absolute & Probability (\%)  & UInt16 &  8.39 Go\\
\bottomrule
\end{tabular}
\end{table*}

\section{Dataset modalities}

\subsection{Mono-temporal and multi-temporal modalities}

In the following, we will detail the characteristics of each source image and supervision. We will focus on describing the spatial, spectral, temporal, and radiometric dimensions of the patches, as well as their origin. The metadata associated with each modality are explained. Table~\ref{tab:MOD_overview} provides a summary of these characteristics. The \mbox{FLAIR \#1} datapaper \cite{FLAIR-1}, had already described some of these modalities (AERIAL\_RGBI, DEM\_ELEV, AERIAL\_LABEL-COSIA). Some of the following paragraphs is taken from this paper.\\

\begin{figure*}[!ht]
    \centering
    \includegraphics[width=0.9\textwidth]{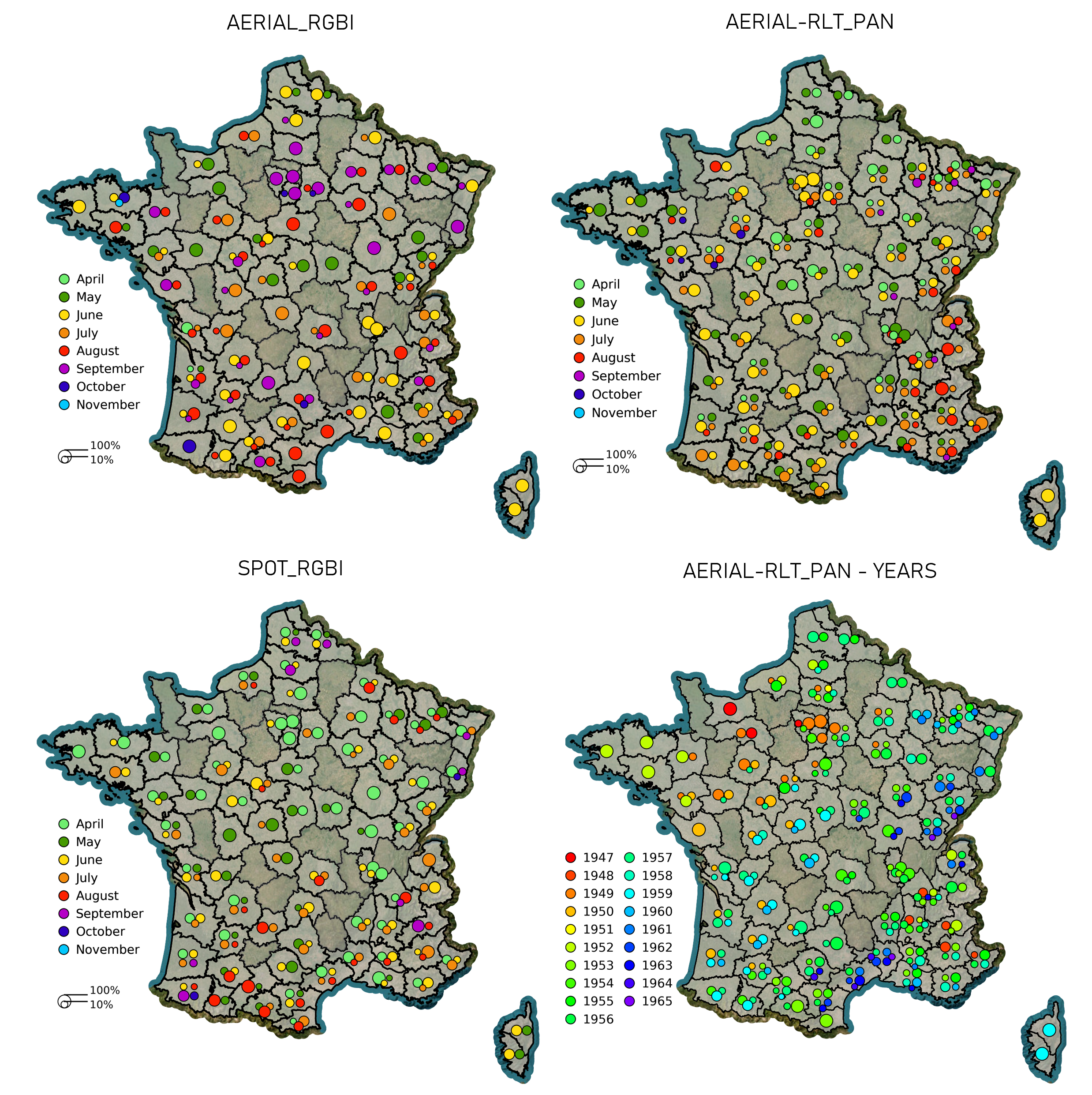}
    \caption{{\bf Temporal Distribution of Mono-temporal Modality Acquisitions.} From top left to bottom right, we provide the information about the month of acquisition for the Aerial VHR images, historical images, SPOT satellite images, and finally, the year of acquisition for the historical data. For each domain, the size of the circle is proportional to the amount of data for that month or year.}
    \label{fig:tempo-mono}
\end{figure*}

\noindent\textbullet\enspace\textbf{AERIAL\_RGBI}: The AERIAL\_RGBI modality was produced using the ORTHO HR\textsuperscript{\textregistered} product; a mosaic of all individual images taken during an aerial survey and mapped onto a cartographic coordinate reference system. Different cameras are used for the aerial image acquisitions. This implies different sensors and consequently, different image characteristics. The final ORTHO HR\textsuperscript{\textregistered} product has a spatial resolution of 0.20\:m (R, G, B and NIR channels). By design, there are no clouds (and therefore no missing data) in the aerial images. Additionally, some radiometric processing methods are applied to obtain the final product. First, radiometric equalization methods are applied to the individual images. Then, a global radiometric correction is carried out on the merged images covering an entire spatial domain to provide a more satisfying colour balance between channels. We consider this radiometric equalization to be relative: within a domain, the radiometric properties are shared (even if the acquisition dates may differ), but there are shifts in radiometry observable between domains (due to both the date of acquisition and the specific radiometric corrections applied). Therefore, the radiometry of R, G, B and NIR images of the ORTHO HR\textsuperscript{\textregistered}  product cannot be considered as a physical measurement of channel reflectance. This radiometric information is encoded as an unsigned 8-bit integer. Each AERIAL\_RGBI patch is structured with a shape of C×H×W where C\:=\:4 channels ans HxW\:=\:512$\times$512 pixels. To take this modality into account more precisely one can use the AERIAL\_MTD\_RADIO-STATS and the AERIAL\_MTD\_DATES metadata. AERIAL\_MTD\_RADIO-STATS provides the mean and standard deviation per patch for the Red, Green, Blue, and Infrared channels. AERIAL\_MTD\_DATES gives for each patch the date and time of acquisition, the name of the original individual image and the name of the camera used. The temporal distribution of the aerial images can be seen in \Cref{fig:tempo-mono}.\\

\noindent\textbullet\enspace\textbf{AERIAL-RLT\_PAN}: Such as the AERIAL\_RGBI modality, the AERIAL-RLT\_PAN is an orthoimage produced with individual aerial surveys images. The image mosaic is stated to be from the 1950's, but in reality, the images intersecting the \mbox{FLAIR-HUB} patches date from 1947 to 1965 (see \Cref{fig:tempo-mono}). To keep things simple, we included the date 195X in the domain name of these modalities. 4 domains in the dataset correspond to the same department but with different acquisition dates. These double domains share the same associated AERIAL-RLT\_PAN domain. As a result, AERIAL-RLT\_PAN is the only modality with 70 domains instead of 74. During these years, aerial photography was not conducted at the department level but rather on a much more local scale, without any specified spatial resolution. Consequently, even though the data has been resampled to a pixel size of 0.4\:m in this dataset, the actual spatial resolution can varies significantly from one area to another. Thanks to other unpublished metadata, we estimate that the actual spatial resolution can vary by a factor of 3, ranging from 0.4\:m to 1.2\:m. The AERIAL-RLT\_MTD\_DATES metadata provides the acquisition date of each patch and the name of the original historical aerial image. All of the images have been acquired in the Panchromatic interval, being sensitive from blue to red wavelengths. Then, intra-domain radiometric equalization is very challenging due to the nature of old film-based images (vignetting). Locally, there can be significant radiometric differences within a single domain. Additionally, statistical equalization techniques are applied, which result in a significant reduction in global contrast. Therefore, handling the radiometry of this modality is particularly challenging.  As with the AERIAL\_RGBI modality, we consider the calibration to be relative and not corresponding to a physical measurement. The value is encoded as an 8-bit unsigned integer. The AERIAL-RLT\_MTD\_RADIO-STATS metadata provides mean and standard deviation information of the panchromatic channel of each patch. Each patch is structured with a shape of H$\times$W\:=\:256$\times$256 pixels.\\

\noindent\textbullet\enspace\textbf{DEM\_ELEV}: This modality has two channels : the Digital Surface Model (DSM) and the Digital Terrain Model (DTM). The DSM gives the altitude, in meters (absolute calibration), for each pixel. Thanks to dense matching techniques, the DSM is derived from the same aerial survey that is used to produce the AERIAL\_RGBI modality. This gives the DSM the same spatial resolution as the AERIAL\_RGBI  but more importantly prevents temporal shifts and ground cover changes between the two products making them temporally coherent. However, there still are small geometric differences between the two products, because orthoimages are projected on the DTM. Moreover, dense matching techniques are applied automatically, potentially introducing noise and artifacts. In particular, radiometrically homogeneous areas (\textit{e.g.}, parts of the aerial image with little texture) tend to lead to locally false and varying 3D information. Apart from these artifacts, the vertical accuracy of the DSM can be considered to be twice the spatial resolution (0.4\:m). The DTM information comes from the RGE ALTI Digital Terrain Model. This product is a national DTM available at a spatial resolution of 1\:m. It is constructed from different sources such as dense matching of aerial images, airborne Lidar or, for mountainous areas, from airborne Synthetic Aperture Radar acquisitions. Depending on the source, the vertical accuracy of the RGE ALTI DTM can vary between 0.3\:m and 7\:m. DTM provides altitude from ground surface, removing buildings, trees, and other objects. The difference between the DSM and the DTM information is then related to the height of buildings or trees. Each patch is therefore sized C$\times$H$\times$W\:=\:2$\times$512$\times$512. Each channel represents an altitude and is therefore encoded as a single-precision floating-point (Float32).\\

\noindent\textbullet\enspace\textbf{SPOT\_RGBI}: This modality has been produced using images from satellite SPOT 6-7 that are acquired each year to produce the French SPOT annual mosaic. For each \mbox{FLAIR-HUB} patch, we chose to use images from the annual SPOT 6-7 mosaic of the same year as the AERIAL\_RGBI images, ensuring temporal proximity between the two modalities (see \Cref{fig:tempo-mono} for the temporal distribution of both modality). However, there can be a gap of several months between the two observations, leading to differences in object appearances in the images (\textit{e.g.}, agriculture, forest) or actual land cover changes. The acquisition date, time, and the name of the original SPOT 6-7 image are provided in the SPOT\_MTD\_DATES metadata table. Initially the images are distributed at 1.50\:m resolution but were resampled in the \FLAIRHUB dataset to 1.60\:m to be a multiple of 0.2\:m. The images include four spectral channels: Red, Green, Blue, and Infrared. Such as Sentinel-2 images the radiometric information is calibrated to Level-2A bottom-of-the-atmosphere reflectance (absolute calibration, no unit). Mean and standard deviation of the 4 channels are given in the SPOT\_MTD\_RADIO-STATS metadata file. The reflectance percentage is encoded as a UInt 16 : 0\:=\:0\% reflectance and to 10 000\:=\:100\% reflectance. The shape of each patch is then C$\times$H$\times$W\:=\:4$\times$64$\times$64.\\

\begin{figure*}[!ht]
    \centering
    \includegraphics[width=0.9\textwidth]{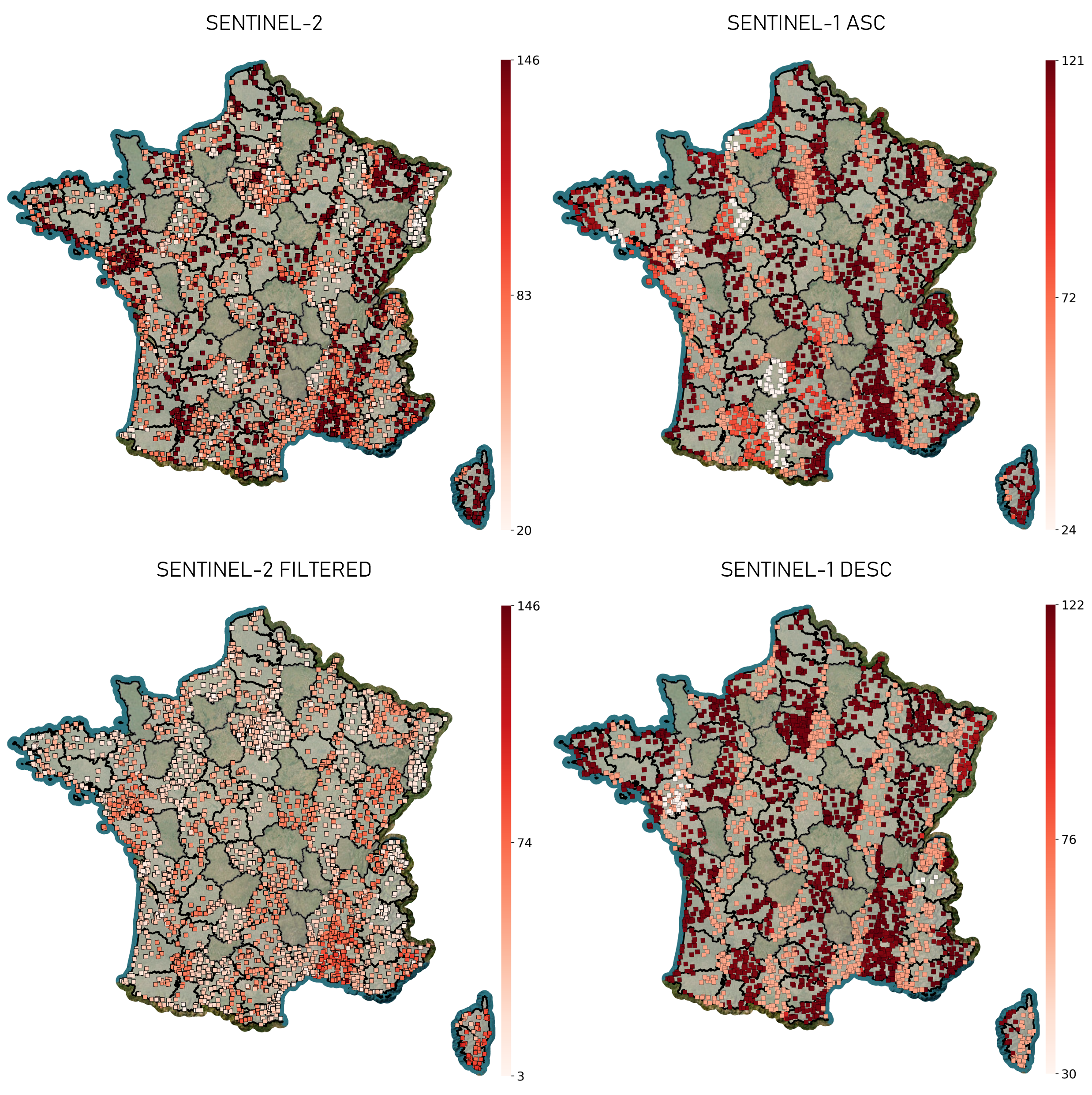}
    \caption{{\bf Spatio-temporal distribution of multi-temporal modality acquisitions.} We plot the number of acquisitions per area for the different STIS; areas are buffered by 5\:km for visualization purposes. The acquisition orbits can be distinguished.}
        \label{fig:tempo-multi}
\end{figure*}

\noindent\textbullet\enspace\textbf{SENTINEL-2\_TS \& \_MSK-SC}: Yearly acquisitions from the Copernicus Sentinel-2A and Sentinel-2B satellites are provided for each area. The time-series (\textit{\_TS)} data correspond to Level-2A bottom-of-the-atmosphere reflectance. The dataset includes 10 spectral channels (B02, B03, B04, B05, B06, B07, B08, B8A, B11, B12), excluding the atmospheric bands with a 60\:m spatial resolution. To ensure consistency with other modalities and fit the spatial extent of patches to a multiple of 0.2\:m, the spatial resolution of Sentinel-2 images was resampled to 10.24\:m. While the nominal revisit time at the equator is 5 days, the actual length of the time series varies significantly across different areas due to orbits, ground segment data gaps, or acquisition failures. Consequently, the number of acquisitions in the dataset ranges from 20 to 146 as it can be seen on \Cref{fig:tempo-multi}. Level-2A data include geophysical masks for snow and cloud cover (\textit{\_MSK-SC}), which are associated with the time-series data to filter out unfavourable acquisition conditions. To reduce the number of files, each dataset patch is structured with a shape of (T$\times$C)$\times$H$\times$W, where T represents the acquisition dates stacked in the first dimension. C\:=\:10 for \textit{\_TS} data and C\:=\:2 for \textit{\_MSK-SC}. The SENTINEL2\_MTD\_DATES metadata is available for analysing time series. For each patch, it provides a dictionary containing the length of the time series and the acquisition date corresponding to each position in the series. This metadata is complemented by SENTINEL2\_MTD\_RADIO-STATS, which describes, in dictionary form, the means and standard deviations of the 10 bands per date. Pixels with a probability of being cloudy strictly greater than 0 are excluded from the mean and standard deviation calculation. When the number of samples for statistical computation is zero, the value \textit{nan} is returned. On the contrary to FLAIR dataset \cite{garioud_flair_2023a,FLAIR-2}, we only consider aligned pixels and do not provide context information from Sentinel-2 time series. That means a reduction of about $93$\% of pixels from Sentinel-2 per VHR patches.\\

\noindent\textbullet\enspace\textbf{SENTINEL-1ASC \& -1DESC\_TS}: Yearly time series from the C-Band Copernicus Sentinel-1A and Sentinel-1B satellites are provided. Ground Range Detected (GRD) products are used in dual-polarization mode (VV and VH). Both ascending (\textit{ASC\_TS)} and descending orbits (\textit{DESC\_TS)} are included separately, as their incidence angles differ significantly. The Sigma nought ($\sigma_{0}$) backscattering coefficient, which represents the normalized radar backscatter intensity of the surface, is calculated for both polarization channels. It provides essential information on surface properties such as roughness, moisture content, and land cover type. No speckle filtering was applied and data averaging results from the GRD product's equivalent number of looks, which is approximately 4. Similar to Sentinel-2 time-series data, each Sentinel-1 patch is structured with a shape of (T$\times$C)$\times$H$\times$W, where T represents the acquisition dates stacked in the first dimension, and C\:=\:2 corresponding to the two polarization channels (VV and VH). \Cref{fig:tempo-multi} illustrates the number of acquisition available for these modalities. To align with the spatial extent requirements and ensure consistency across modalities, the spatial resolution of Sentinel-1 images was resampled to 10.24\:m. Just like the metadata describing the Sentinel-2 series, the files SENTINEL1-ASC\_MTD\_DATES, SENTINEL1-DESC\_MTD\_DATES, SENTINEL1-ASC\_MTD\_RADIO-STATS, and SENTINEL1-DESC\_MTD\_RADIO-STATS describe the lengths, dates, and statistics of the Sentinel-1 series.\\

\begin{table*}[!t]
\small
\centering
\footnotesize
\setlength{\tabcolsep}{3.5pt}
\renewcommand{\arraystretch}{1.25}

\caption{{\bf Semantic classes and their frequency in the LABEL-COSIA nomenclature of the \mbox{FLAIR-HUB} dataset.}}

\begin{tabular}{p{0.6cm}lccrrrrrr}
 & \textbf{Class}              & \textbf{LABEL-COSIA}  &  \textbf{Baseline}  & \textbf{Pixels}   & \textbf{\% total} & \textbf{\% train} & \textbf{\% valid} & \textbf{\% test} \\ \hline
\cellcolor[HTML]{db0e9a} & Building                     & 0 &  \true        & 3\,483\,982\,647   &  5.51    & 5.33   & 5.42 & 6.13     \\
\cellcolor[HTML]{9999ff} & Greenhouse                   & 1 &  \true        & 134\,298\,230     &   0.21   & 0.23   & 0.15  & 0.20    \\
\cellcolor[HTML]{3de6eb} & Swimming pool                & 2 &  \true        & 17\,885\,590      &  0.03    & 0.03   & 0.02  & 0.04    \\
\cellcolor[HTML]{f80c00} & Impervious surface           & 3 &  \true        & 5\,796\,512\,286   &  9.17    & 8.84   & 8.81  & 10.44   \\
\cellcolor[HTML]{938e7b} & Pervious surface             & 4 &  \true        & 3\,530\,039\,654   &  5.59    & 5.60   & 6.04  & 5.21    \\
\cellcolor[HTML]{a97101} & Bare soil                    & 5 &  \true        & 2\,539\,309\,904   &  4.02    & 4.21   & 3.96  & 3.49    \\
\cellcolor[HTML]{1553ae} & Water                        & 6 &  \true        & 3\,308\,863\,698   &  5.24    & 5.27   & 5.60  & 4.86    \\
\cellcolor[HTML]{ffffff} & Snow                         & 7 &  \true        & 443\,134\,338     &  0.70    & 0.72   & 0.40  & 0.88    \\
\cellcolor[HTML]{55ff00} & Herbaceous vegetation        & 8 &  \true        & 10\,998\,905\,498  &  17.40   & 16.79  & 16.42 & 19.99   \\
\cellcolor[HTML]{fff30d} & Agricultural land            & 9 &  \true        & 8\,665\,649\,328   &  13.71   & 14.17  & 14.69 & 11.59   \\
\cellcolor[HTML]{e4df7c} & Plowed land                  & 10 & \true        & 1\,733\,051\,984   &  2.74    & 3.04   & 3.03  & 1.63    \\
\cellcolor[HTML]{660082} & Vineyard                     & 11 & \true        & 1\,647\,328\,848   &  2.61    & 2.67   & 2.69  & 2.35    \\
\cellcolor[HTML]{46e483} & Deciduous                    & 12 & \true        & 12\,731\,200\,586  &  20.14   & 20.23  & 19.95 & 20.04   \\
\cellcolor[HTML]{194a26} & Coniferous                   & 13 & \true        & 4\,227\,196\,348   &  6.69    & 6.46   &  6.35 & 7.62    \\
\cellcolor[HTML]{f3a60d} & Brushwood                    & 14 & \true        & 3\,451\,432\,094   &  5.46    & 5.63   & 5.66  & 4.80    \\
\cellcolor[HTML]{8ab3a0} & Clear cut                    & 15 & \false       & 378\,090\,812     &  0.60    & 0.61   & 0.70  & 0.49    \\
\cellcolor[HTML]{c5dc42} & Ligneous                     & 16 & \false       & 2\,809\,408       &  0.00    & 0.00   & 0.00  & 0.00    \\
\cellcolor[HTML]{6b714f} & Mixed                        & 17 & \false       & 36\,603\,366      &  0.06    & 0.05   & 0.01 & 0.12     \\
\cellcolor[HTML]{000000} & Undefined                    & 18 & \false       & 74\,348\,348      &  0.12    & 0.12   & 0.09 & 0.13     \\ \hline
\end{tabular}

\label{tab:LABEL_COSIA}
\end{table*}

\noindent\textbullet\enspace\textbf{AERIAL\_LABEL-COSIA}:
The  AERIAL\_LABEL-COSIA supervision consists in determining the land cover at the pixel-level. It is based on photo-interpretation of the AERIAL\_RGBI images and has been manually produced by experts. An initial spatial multi-level image segmentation approach \cite{PYRAM} was applied, simplifying the labelling at the cluster level. We note that this segmentation was not necessarily final, but was modified interactively when deemed appropriate. It was specified that movable objects (\textit{e.g.}, cars, boats) are not to be annotated as such, but to be classified as the underlying cover. For example, a car on an asphalt road is labelled as an impervious surface \cite{FLAIR-1}. We explicitly named this supervision AERIAL\_LABEL-COSIA because it is both temporally and geometrically fully consistent with the AERIAL\_RGBI images. The AERIAL\_LABEL-COSIA patches are then shaped as C$\times$H$\times$W\:=\:512$\times$512.

The land cover classification consists of 19 classes, ranging from 0 to 18. Table~\ref{tab:LABEL_COSIA} provides the list of classes and their respective label number and their frequency in terms of pixel count and percentage (for total, train, valid and test partitions). Please note that the class order has changed compared to previous versions of the dataset (\cite{FLAIR-1, FLAIR-2}) to ensure that labels start at 0, classes are thematically organized, and classes corresponding to weak labels appear at the end of the nomenclature. Only the first 15 classes are used in the \FLAIRHUB experiments. Classes 0 to 4 correspond to different types of anthropized areas. These are the categories of interest to be used, for example, to monitor soil land take. Classes 5 to 8 represent natural surfaces without agricultural intensive usage. Classes 9 to 11 relate to agricultural areas, while classes 12 to 17 correspond to forested areas. Class 16 (ligneous) and 17 (mixed) can be considered as weak labels. Due to the polygon-based annotation process , the photo-interpreter may sometimes be unable to distinguish between the \textit{deciduous} and \textit{coniferous} classes. In such cases, they use the \textit{ligneous} label. Similarly, if both \textit{deciduous} and \textit{coniferous} are present within the same polygon, the \textit{mixed} code is used. The geometric segmentation was fine enough to ensure that this weak labels remain very rare ($\approx$0.06\% of the annotated pixels). Finally, class 19 indicates cases where the photo-interpreter could not provide an annotation (\textit{e.g.}, shadows, uncertainty between certain classes). For a more in-depth analysis of the AERIAL\_LABEL-COSIA labels, the AERIAL\_MTD\_LABEL-COSIA metadata file is available. This file provides, for each patch, the histogram of AERIAL\_LABEL-COSIA labels (unit: pixels).\\

Even though annotations are made with photo-interpretation, some errors are unavoidable, especially for classes that are visually hard to distinguish, such as bare soil and {pervious surfaces}. Around $~$37\:k randomly chosen polygons were manually annotated, remaining hidden from the annotating teams. This accounted for an area of 18.7 km$^2$ equivalent to approximately 468 million pixels. Annotation batches not achieving 95\% accuracy were rejected and sent back for re-annotation. This iterative process fostered productive exchanges between the annotators and independent geography experts, ensuring a high-quality dataset.\\

\begin{figure}[!htpb]
    \centering
    \includegraphics[height=\textheight, keepaspectratio]{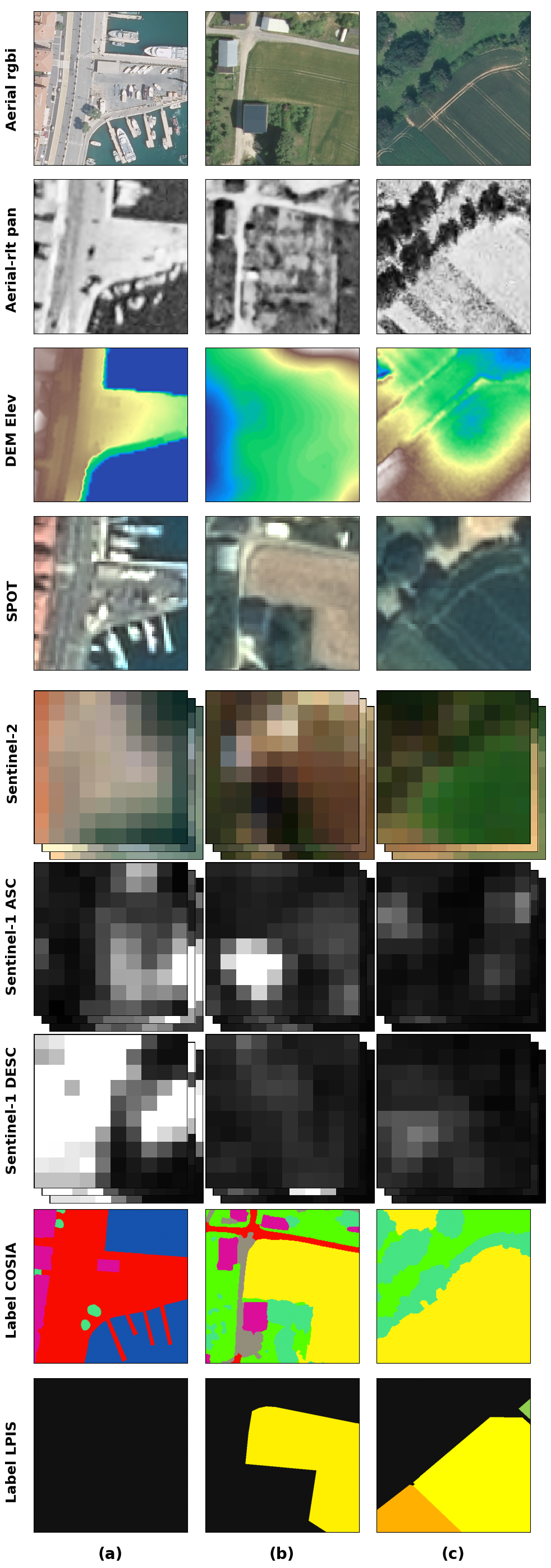}
    \caption{{\bf Example patches from the dataset}, illustrating all available modalities. We only plot one image per satellite time series. ASC stands for ascendant and DESC for descendant.}
        \label{fig:test2}
\end{figure}

\begin{table*}[htpb]
\small
\centering
\footnotesize
\setlength{\tabcolsep}{2.2pt}
\renewcommand{\arraystretch}{1.25}

\caption{{\bf Semantic classes and their frequency in the LABEL-LPIS multilevel nomenclature of the \mbox{FLAIR-HUB} dataset.}}

\begin{tabular}{p{0.6cm}lccccccc|lc|lc|}
\hline
 & \textbf{Class LV.1} & \textbf{LV.1} & \textbf{Baseline} & \textbf{Pixels} & \textbf{\%total} & \textbf{\%train} & \textbf{\%valid} & \textbf{\%test} & \textbf{Class LV.2} & \textbf{LV.2} & \textbf{Class LV.3} & \textbf{LV.3} \\ \hline
\cellcolor[HTML]{92d050} &  &  &  &  &  &  &  &  &  &  & Grasses monoculture & 0 \\ \cline{12-13} 
\multirow{-2}{*}{\cellcolor[HTML]{92d050}} & \multirow{-2}{*}{Grasses} & \multirow{-2}{*}{0} & \multirow{-2}{*}{\true} & \multirow{-2}{*}{6\,318\,583\,131} & \multirow{-2}{*}{10.0} & \multirow{-2}{*}{9.36} & \multirow{-2}{*}{8.66} & \multirow{-2}{*}{12.91} & \multirow{-2}{*}{Grasses} & \multirow{-2}{*}{0} & Grasses mixture & 1 \\ \cline{1-13}
\cellcolor[HTML]{d7e600} &  &  &  &  &  &  &  &  &  &  & Winter wheat & 2 \\ \cline{12-13} 
\multirow{-2}{*}{\cellcolor[HTML]{d7e600}} & \multirow{-2}{*}{Wheat} & \multirow{-2}{*}{1} & \multirow{-2}{*}{\true} & \multirow{-2}{*}{1\,639\,060\,670} & \multirow{-2}{*}{2.59} & \multirow{-2}{*}{2.84} & \multirow{-2}{*}{3.18} & \multirow{-2}{*}{1.41} & \multirow{-2}{*}{Wheat} & \multirow{-2}{*}{1} & Spring wheat & 3 \\ \cline{1-13}
\cellcolor[HTML]{e0e000} &  &  &  &  &  &  &  &  &  &  & Winter barley & 4 \\ \cline{12-13} 
\multirow{-2}{*}{\cellcolor[HTML]{e0e000}} & \multirow{-2}{*}{Barley} & \multirow{-2}{*}{2} & \multirow{-2}{*}{\true} & \multirow{-2}{*}{658\,266\,930} & \multirow{-2}{*}{1.04} & \multirow{-2}{*}{1.09} & \multirow{-2}{*}{0.94} & \multirow{-2}{*}{0.96} & \multirow{-2}{*}{Barley} & \multirow{-2}{*}{2} & Spring barley & 5 \\ \cline{1-13}
\cellcolor[HTML]{fff100} & Maize & 3 & \true & 1\,288\,052\,739 & 2.04 & 2.09 & 2.37 & 1.63 & Maize & 3 & Maize & 6 \\ \cline{1-13}
\cellcolor[HTML]{FFFF00} &  &  &  &  &  &  &  &  &  &  & Sorghum & 7 \\ \cline{12-13} 
\cellcolor[HTML]{FFFF00} &  &  &  &  &  &  &  &  & \multirow{-2}{*}{Sorghum/Millet} & \multirow{-2}{*}{4} & Millet / Foxtail millet & 8 \\ \cline{10-13} 
\cellcolor[HTML]{FFFF00} &  &  &  &  &  &  &  &  &  &  & Winter durum wheat & 9 \\ \cline{12-13} 
\cellcolor[HTML]{FFFF00} &  &  &  &  &  &  &  &  &  &  & Winter triticale & 10 \\ \cline{12-13} 
\cellcolor[HTML]{FFFF00} &  &  &  &  &  &  &  &  &  &  & Winter oat & 11 \\ \cline{12-13} 
\cellcolor[HTML]{FFFF00} &  &  &  &  &  &  &  &  & \multirow{-4}{*}{Other winter cereals} & \multirow{-4}{*}{5} & Winter rye & 12 \\ \cline{10-13} 
\cellcolor[HTML]{FFFF00} &  &  &  &  &  &  &  &  &  &  & Spring oat & 13 \\ \cline{12-13} 
\cellcolor[HTML]{FFFF00} &  &  &  &  &  &  &  &  & \multirow{-2}{*}{Other spring cereals} & \multirow{-2}{*}{6} & Other spring cereals & 14 \\ \cline{10-13} 
\multirow{-9}{*}{\cellcolor[HTML]{FFFF00}} & \multirow{-9}{*}{Other cereals} & \multirow{-9}{*}{4} & \multirow{-9}{*}{\true} & \multirow{-9}{*}{383\,132\,669} & \multirow{-9}{*}{0.61} & \multirow{-9}{*}{0.62} & \multirow{-9}{*}{0.63} & \multirow{-9}{*}{0.55} & Other cereals & 7 & Other cereals & 15 \\ \cline{1-13}
\cellcolor[HTML]{E8E8E8} & Rice & 5 & \true & 42\,118\,356 & 0.07 & 0.09 & 0.07 & 0.00 & Rice & 8 & Rice & 16 \\ \cline{1-13}
\cellcolor[HTML]{DCEAF7} &  &  &  &  &  &  &  &  & Hemp/Tobacco & 9 & Hemp/Tobacco & 17 \\ \cline{10-13} 
\cellcolor[HTML]{DCEAF7} &  &  &  &  &  &  &  &  &  &  & Fiber flax & 18 \\ \cline{12-13} 
\multirow{-3}{*}{\cellcolor[HTML]{DCEAF7}} & \multirow{-3}{*}{Hemp/Flax/Tobacco} & \multirow{-3}{*}{6} & \multirow{-3}{*}{\true} & \multirow{-3}{*}{47\,,265\,258} & \multirow{-3}{*}{0.07} & \multirow{-3}{*}{0.04} & \multirow{-3}{*}{0.09} & \multirow{-3}{*}{0.17} & \multirow{-2}{*}{Flax} & \multirow{-2}{*}{10} & Other flax & 19 \\ \cline{1-13}
\cellcolor[HTML]{D29EAD} & Sunflower & 7 & \true & 473\,323\,562 & 0.75 & 0.90 & 0.98 & 0.12 & Sunflower & 11 & Sunflower & 20 \\ \cline{1-13}
\cellcolor[HTML]{d29ed0} & Rapeseed & 8 & \true & 366\,394\,829 & 0.58 & 0.57 & 0.91 & 0.35 & Rapeseed & 12 & Rapeseed & 21 \\ \cline{1-13}
\cellcolor[HTML]{ffbe99} &  &  &  &  &  &  &  &  &  &  & Mustard & 22 \\ \cline{12-13} 
\multirow{-2}{*}{\cellcolor[HTML]{ffbe99}} & \multirow{-2}{*}{Other oilseed crops} & \multirow{-2}{*}{9} & \multirow{-2}{*}{\true} & \multirow{-2}{*}{7\,823\,820} & \multirow{-2}{*}{0.01} & \multirow{-2}{*}{0.02} & \multirow{-2}{*}{0.00} & \multirow{-2}{*}{0.00} & \multirow{-2}{*}{Other oilseed crops} & \multirow{-2}{*}{13} & Other oilseed crops & 23 \\ \cline{1-13}
\cellcolor[HTML]{FFC000} & Soy & 10 & \true & 122\,451\,006 & 0.19 & 0.27 & 0.10 & 0.04 & Soy & 14 & Soy & 24 \\ \cline{1-13}
\cellcolor[HTML]{ff9000} &  &  &  &  &  &  &  &  &  &  & Spring peas & 25 \\ \cline{12-13} 
\cellcolor[HTML]{ff9000} &  &  &  &  &  &  &  &  &  &  & Winter protein crops & 26 \\ \cline{12-13} 
\multirow{-3}{*}{\cellcolor[HTML]{ff9000}} & \multirow{-3}{*}{Other protein crops} & \multirow{-3}{*}{11} & \multirow{-3}{*}{\true} & \multirow{-3}{*}{147\,499\,283} & \multirow{-3}{*}{0.23} & \multirow{-3}{*}{0.25} & \multirow{-3}{*}{0.39} & \multirow{-3}{*}{0.06} & \multirow{-3}{*}{Other protein crops} & \multirow{-3}{*}{15} & {\color[HTML]{242424} Other protein crops} & 27 \\ \cline{1-13}
\cellcolor[HTML]{009999} &  &  &  &  &  &  &  &  & Alfalfa & 16 & Alfalfa & 28 \\ \cline{10-13} 
\cellcolor[HTML]{009999} &  &  &  &  &  &  &  &  &  &  & Clover & 29 \\ \cline{12-13} 
\multirow{-3}{*}{\cellcolor[HTML]{009999}} & \multirow{-3}{*}{Fodder legumes} & \multirow{-3}{*}{12} & \multirow{-3}{*}{\true} & \multirow{-3}{*}{385\,503\,065} & \multirow{-3}{*}{0.61} & \multirow{-3}{*}{0.67} & \multirow{-3}{*}{0.60} & \multirow{-3}{*}{0.45} & \multirow{-2}{*}{Other fodder legumes} & \multirow{-2}{*}{17} & Other fodder legumes & 30 \\ \cline{1-13}
\cellcolor[HTML]{808000} & Beetroots & 13 & \true & 117\,492\,248 & 0.19 & 0.20 & 0.26 & 0.09 & Beetroots & 18 & Beetroots & 31 \\ \cline{1-13}
\cellcolor[HTML]{a7a700} & Potatoes & 14 & \true & 67\,363\,659 & 0.11 & 0.12 & 0.13 & 0.06 & Potatoes & 19 & Potatoes & 32 \\ \cline{1-13}
\cellcolor[HTML]{89896d} &  &  &  &  &  &  &  &  & Fruits and vegetables & 20 & Fruits and vegetables & 33 \\ \cline{10-13} 
\cellcolor[HTML]{89896d} &  &  &  &  &  &  &  &  & \makecell[l]{Aromatic/Medicinal\\plants} & 21 & \makecell[l]{Aromatic/Medicinal\\plants} & 34 \\ \cline{10-13} 
\cellcolor[HTML]{89896d} &  &  &  &  &  &  &  &  &  &  & Buckwheat & 35 \\ \cline{12-13} 
\multirow{-4}{*}{\cellcolor[HTML]{89896d}} & \multirow{-4}{*}{Other arable crops} & \multirow{-4}{*}{15} & \multirow{-4}{*}{\true} & \multirow{-4}{*}{307\,449\,041} & \multirow{-4}{*}{0.49} & \multirow{-4}{*}{0.59} & \multirow{-4}{*}{0.39} & \multirow{-4}{*}{0.26} & \multirow{-2}{*}{Other arable crops} & \multirow{-2}{*}{22} & {\color[HTML]{242424} Other arable crops} & 36 \\ \cline{1-13}
\cellcolor[HTML]{F2CFEE} & {\color[HTML]{242424} Vineyard} & 16 & \true & 1\,141\,947\,919 & 1.81 & 1.92 & 1.65 & 1.57 & {\color[HTML]{242424} Vineyard} & 23 & {\color[HTML]{242424} Vineyard} & 37 \\ \cline{1-13}
\cellcolor[HTML]{6f6633} & Olive groves & 17 & \true & 51\,694\,187 & 0.08 & 0.09 & 0.04 & 0.09 & Olive groves & 24 & Olive groves & 38 \\ \cline{1-13}
\cellcolor[HTML]{ac8141} & Fruit orchards & 18 & \true & 504\,280\,801 & 0.80 & 0.87 & 1.00 & 0.42 & Fruit orchards & 25 & Fruit orchards & 39 \\ \cline{1-13}
\cellcolor[HTML]{996633} & Nut orchards & 19 & \true & 80\,408\,145 & 0.13 & 0.16 & 0.03 & 0.11 & Nut orchards & 26 & Nut orchards & 40 \\ \cline{1-13}
\cellcolor[HTML]{80c1d7} &  &  &  &  &  &  &  &  & Lavandin & 27 & Lavandin & 41 \\ \cline{10-13} 
\cellcolor[HTML]{80c1d7} &  &  &  &  &  &  &  &  &  &  & Berries & 42 \\ \cline{12-13} 
\multirow{-3}{*}{\cellcolor[HTML]{80c1d7}} & \multirow{-3}{*}{Other permanent crops} & \multirow{-3}{*}{20} & \multirow{-3}{*}{\true} & \multirow{-3}{*}{97\,546\,687} & \multirow{-3}{*}{0.15} & \multirow{-3}{*}{0.03} & \multirow{-3}{*}{0.50} & \multirow{-3}{*}{0.27} & \multirow{-2}{*}{Other permanent crops} & \multirow{-2}{*}{28} & Other permanent crops & 43 \\ \cline{1-13}
\cellcolor[HTML]{000000} & Mixed crops & 21 & \true & 230\,723\,268 & 0.37 & 0.37 & 0.39 & 0.32 & Mixed crops & 29 & Mixed crops & 44 \\ \cline{1-13}
\cellcolor[HTML]{000000} & Background & 22 & \true & 48\,724\,537\,127 & 77.09 & 76.84 & 76.68 & 78.15 & Background & 30 & Background & 45 \\ \hline
\end{tabular}
\label{tab:LABEL_LPIS}
\end{table*}

\noindent\textbullet\enspace\textbf{ALL\_LABEL-LPIS}:
The AERIAL\_LABEL-LPIS modality provides information on agricultural surfaces, structured into a semantic crop classification with three hierarchical levels. This annotation is derived from the French LPIS (Land Parcel Identification System), which consists of declarative parcel data digitally submitted by farmers. These declarations are made within the framework of the European Common Agricultural Policy (CAP), which provides subsidies and support based on land use. However, farmers are not obligated to declare all their parcels but only those relevant to their subsidy claims. As a result, not all agricultural parcels present within a given ROI are included in the AERIAL\_LABEL-LPIS modality, since only parcels declared under the CAP are represented. For instance, it is known that vineyard surfaces are frequently absent from the labels due to non-declaration.\\

LPIS data is digitized by farmers using an online platform, based on the most recent available aerial orthoimagery which is identical to that used in the AERIAL\_RGBI modality. To ensure temporal consistency, LPIS data for each Region of Interest (ROI) was selected from the same year as the corresponding AERIAL\_RGBI imagery. Consequently, the AERIAL\_LABEL-LPIS modality offers temporally aligned supervision across multiple years.\\

The original LPIS classification includes over 230 crop types. To enhance consistency across years and simplify downstream tasks, these classes were harmonized and organized into a three-level hierarchical taxonomy. While this taxonomy diverges from HCAT \cite{schneider2023eurocrops} at the second and third levels, it remains broadly similar at the first level. The second and third level classes were determined based on crop availability within the ROIs. Crop types with limited spatial extent (\textit{e.g.}, present in $<10$ ROIs or domains) were merged with similar crops based on phenological traits and growing seasons. Certain LPIS classes—those dominated by ligneous vegetation (\textit{e.g.}, pastured woodlands or oak/chestnut groves), aquatic areas (\textit{e.g.}, salt marshes or reed beds), or artificial surfaces (\textit{e.g.}, greenhouses) were grouped under the background class. In contrast, mixed-crop classes were deliberately left unmerged with homogeneous crop classes. These mixed classes are better interpreted as unknown and are therefore excluded from training loss computation. The final taxonomy contains 23, 31, and 46 classes at levels 1, 2, and 3 respectively. These have been rasterized as separate channels within a single TIFF file, with a spatial resolution of 20 cm—matching that of the AERIAL\_RGBI modality\\

Spatial discrepancies between AERIAL\_LABEL-LPIS and AERIAL\_LABEL-COSIA are expected due to the declarative nature of LPIS data integrating land use alongside land cover. Since parcel boundaries in LPIS are manually digitized by farmers and may not align precisely with the actual land cover, pixel-level correspondence between the two labels is not guaranteed. As a result, pixels within a single LPIS parcel can belong to different AERIAL\_LABEL-COSIA classes, particularly along the edges or in areas with internal heterogeneity.\\

As previously mentioned, LPIS data are declarative in nature, provided by farmers, and therefore lack an associated recall threshold at the object (parcel) level. Nonetheless, certain quality criteria must be met for declared parcels. First, the overall error rate of the declared crop type should be approximately 2\%, and remain below 5\%. Second, with regard to the geometric accuracy of parcel boundaries, the precision for parcel block boundaries separating a parcel from the background is expected to be around 1\:m. In contrast, the precision for internal boundaries between adjacent parcels may be lower, with discrepancies of up to 5\:m. This reduced accuracy stems from the fact that digitization is performed prior to crop sowing, using ortho-images from previous years, as current-year imagery is not yet available. While external parcel block boundaries tend to be temporally stable and can therefore be digitized more precisely, internal parcel boundaries may vary annually and may not correspond to visible features in historical imagery. Finally, as shown in Table~\ref{tab:LABEL_LPIS}, the selected ROIs in the current dataset were not chosen to ensure a minimum area for all crop types. Consequently, this may affect the performance of models trained using LPIS labels.\\

\Cref{fig:test2} provides a visual comparison of the different data modalities available in the dataset, shown over three patches sampled from geographically distinct regions. This illustration highlights the diversity of sensor inputs and annotations in terms of spatial, temporal, and radiometric resolutions.

\begin{figure*}[!t]
    \centering
    \includegraphics[width=1\textwidth]{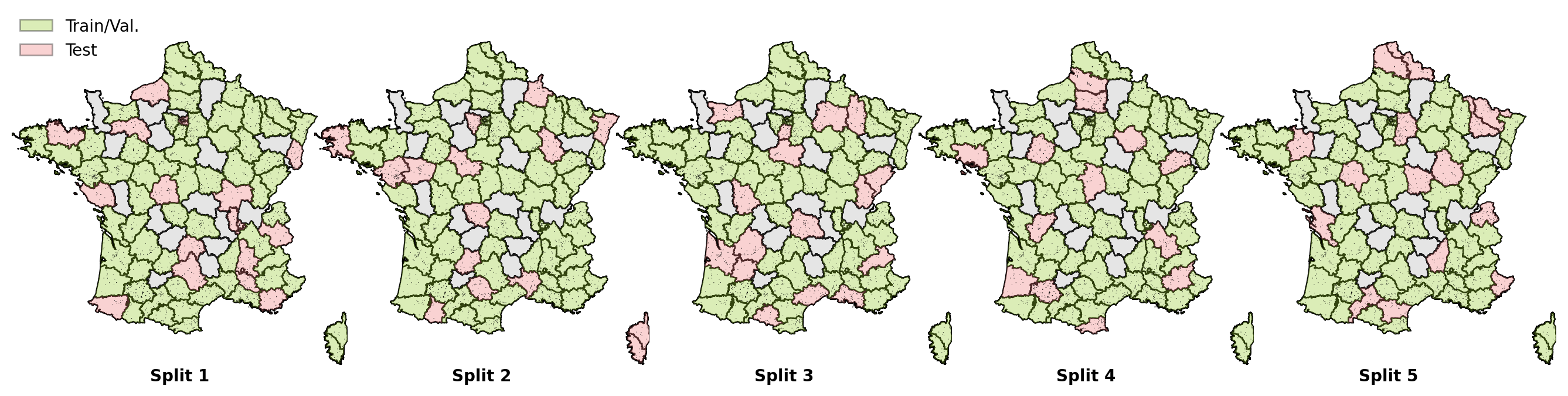}
    \caption{{\bf Spatial distribution of splits in a k-fold configuration.} Split 1 corresponds to the official \FLAIRHUB split (also named \textit{split\_flairhub}).}
        \label{fig:splits}
\end{figure*}

\subsection{Official dataset partitions}
\label{sec:dataset_partitions}

Information regarding the usage of patches for training, validation, or testing is provided in the metadata file \textit{GLOBAL\_ALL\_MTD\_SPLIT}. This file establishes the correspondence between each \textit{patch\_id} and its associated data split. Seven predefined splits are available: \textit{split\_flairhub}, \textit{split\_1}, \textit{split\_2}, \textit{split\_3}, \textit{split\_4}, \textit{split\_5}, and \textit{split\_flairchallenge}.\\

\noindent\textbullet\enspace\textbf{split\_1, \dots, split\_5}:
The general approach for creating these splits was to expertly define five clusters of domains. Each cluster is designed to represent all types of landscapes and climates. A separation between train+validation and test is then performed domain-wise. Split\_1 used cluster 1 for testing and clusters 2 to 5 for training and validation. Similarly, for splits 2 to 5, the test partition was rotated accordingly. Next, the separation between training and validation was performed ROI-wise. This choice was made to ensure that both the training and validation sets contain a sufficient number of domains while maintaining spatial independence between patches. Indeed, contiguous patches belonging to the same ROI all share the same usage, either training or validation.  The train/validation split was performed to obtain a 80\%\:/\:20\% distribution between train and validation, respectively. Splits 1 to 5 are therefore the splits we will use when a model needs to be evaluated through cross-validation. \Cref{fig:splits} illustrates the five splits and the rotation of the test set.\\

\noindent\textbullet\enspace\textbf{split\_flairhub}:
The split named \textit{split\_flairhub} is the one primarily used for the experiments presented in this paper. It is identical to \textit{split\_1} from the cross-validation setup. For convenience, this split has been duplicated under a separate name. In this split, the TRAIN set comprises 152\,225 patches, the VALIDATION set contains 38\,175 patches, and the TEST set includes 50\,700 patches. \Cref{tab:LABEL_COSIA} shows the relative proportions of AERIAL\_LABEL-COSIA across the training, validation, and test sets, while \Cref{tab:LABEL_LPIS} reports the corresponding distribution for ALL\_LABEL-LPIS.\\

\noindent\textbullet\enspace\textbf{split\_flairchallenge}:
The entirety of the FLAIR~\#1 \cite{FLAIR-1} and FLAIR~\#2 \cite{FLAIR-2,garioud_flair_2023a} datasets is included within the \mbox{FLAIR-HUB} dataset. FLAIR \#1 and FLAIR \#2 datasets differ only in their test partitions. The \textit{split\_flairchallenge} metadata enables the reproduction of experiments from both publications by indicating, for each \mbox{FLAIR-HUB} patch, whether it was not used in either dataset (\textit{none}), or whether it was used for training (\textit{train}), validation (\textit{valid}), the FLAIR~\#1 test set (\textit{test-1}), or the FLAIR~\#2 test set (\textit{test-2}). The FLAIR~\#1 and FLAIR~\#2  test sets are included in the test set of the official split : \textit{split\_flairhub}.

\begin{figure*}[!ht]
    \centering
    \includegraphics[width=1\textwidth]{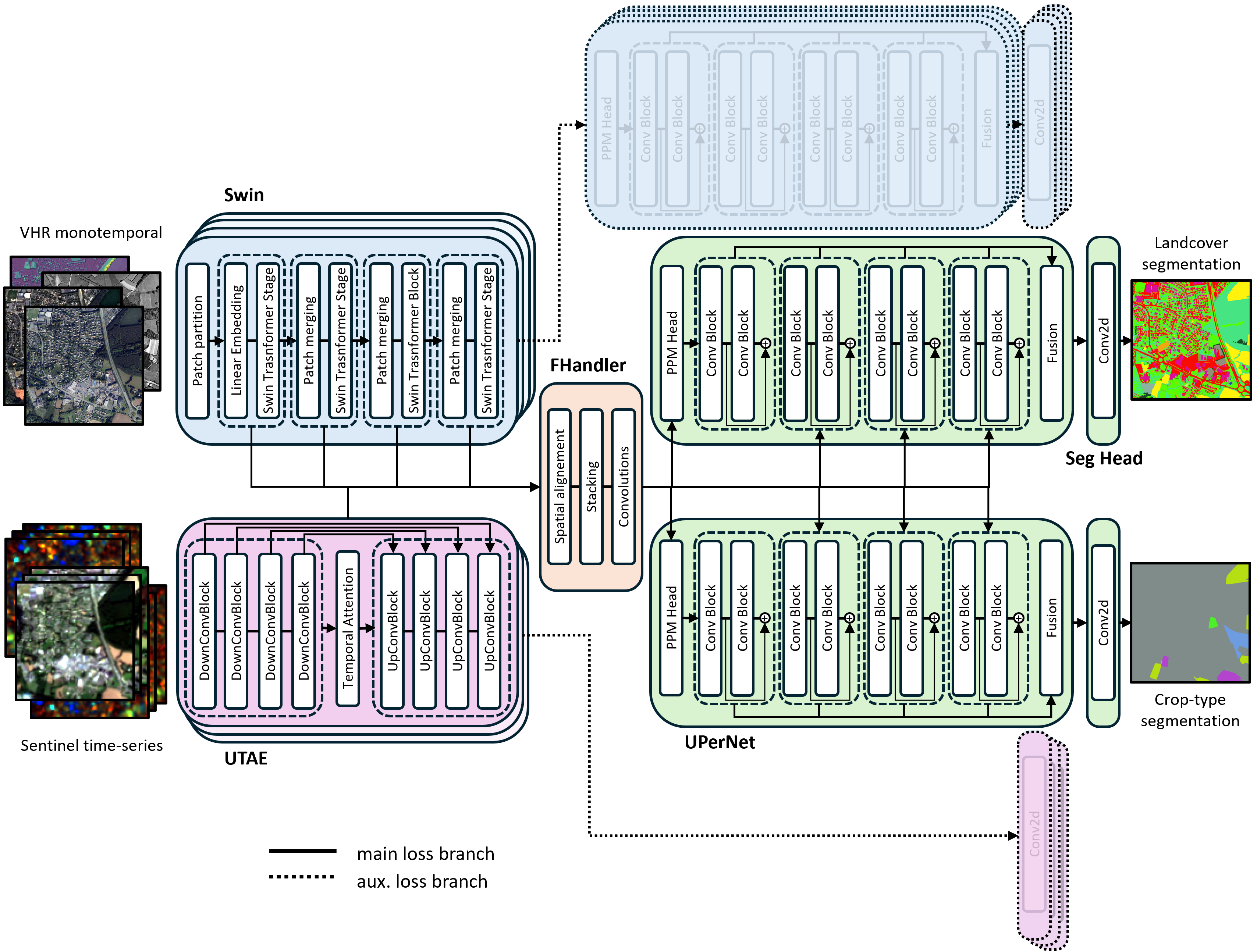}
    \caption{{\bf Architecture of the baseline \mbox{UPerFuse} model} designed for multimodal fusion and multi-task semantic segmentation. The transparent modules correspond to auxiliary loss branches. }
        \label{fig:archi}
\end{figure*}

\section{Baseline architecture}
\label{sec:baseline_archi}

The baseline architecture, namely \mbox{\textbf{UPerFuse}}, integrates multiple feature extraction and a fusion strategy which are shown in Figure~\ref{fig:archi}. It is built upon four primary components: a Swin Transformer feature extractor, a UTAE spatio-temporal encoder, a fusion mechanism, and a UPerNet decoder for segmentation.\\

\noindent\textbf{Swin Transformer}: the Swin Transformer \cite{liu2021swin} module processes mono-temporal data and is designed for hierarchical spatial feature extraction. The input imagery is first partitioned into patches, which undergo linear embedding to transform them into feature representations. These representations are processed through multiple Swin Transformer stages, interspersed with patch merging operations. The feature representations from the patches are processed through a series of Swin Transformer stages, each containing a specific number of Swin Transformer blocks (\textit{i.e.}, 2, 2, 6 and 2). Each stage alternates between layers of regular and shifted window-based multi-head self-attention. These stages include patch merging operations that reduce spatial resolution while increasing the channel dimension. The Swin Transformer blocks within these stages leverage layer normalization, multi-layer perceptrons, and attention mechanisms to capture long-range dependencies and hierarchical feature representations. Skip connections are used to facilitate gradient flow and preserve information across layers, providing a comprehensive understanding of spatial structures.\\

\noindent\textbf{UTAE}: the UTAE (U-Net with Temporal Attention Encoder, \cite{UTAE}) module processes multi-temporal data. It consists of a series of Down-Convolution blocks that progressively reduce the spatial resolution while enhancing feature representation. A temporal attention mechanism is applied to capture dependencies across different time steps in multi-temporal imagery. The extracted features are then upsampled through multiple Up-Convolution blocks to restore spatial resolution while preserving temporal context.\\

\noindent\textbf{Fusion Mechanism}: the outputs from the Swin Transformer and UTAE modules are combined through a dedicated FusionHandler module. This fusion mechanism performs spatial alignment via interpolation, feature stacking, and convolutional refinement.\\

\noindent\textbf{UPerNet Decoder}: the fused features are processed using the UPerNet decoder \cite{upernet}, which consists of a Pyramid Pooling Module (PPM) and multiple convolutional blocks. The PPM aggregates multi-scale contextual information, which is subsequently refined through a series of convolutional blocks. Skip connections and hierarchical fusion ensure the preservation of fine-grained spatial details.\\

\noindent\textbf{Segmentation Head}: Final segmentation predictions are obtained through a series of convolutional layers applied to either the UPerNet or UTAE output, depending on the input modality. If only multi-temporal data is used, the UTAE logits outputs are directly passed to the segmentation heads. The workflow dynamically adapts based on the input modalities (mono-temporal or multi-temporal). If any mono-temporal modality is active, all feature maps are passed through the FusionHandler, enabling per-stage alignment and merging of feature maps. The merged feature maps are then processed for each task (\textit{e.g.}, land cover or crop-type mapping) using a common UPerNet decoder followed by a segmentation head. When only multi-temporal modalities are utilized, the UTAE logits outputs are directly employed for segmentation without requiring additional fusion.\\

\noindent\textbf{Auxiliary branches} are integrated for each modality to improve gradient flow. These branches process encoder feature maps independently through separate decoders, bypassing the fusion step, and directly feeding into segmentation heads. This design ensures robust feature extraction while maintaining flexibility in handling diverse remote sensing data inputs.\\

\noindent\textbf{Network supervision}: The total loss function of the proposed model is designed to handle multiple tasks, multiple modalities, and auxiliary losses, with weighted contributions for each component.

\noindent\textit{Main Loss}: for each task \( t \), let \( \mathcal{L}_{t} \) be the primary loss computed using a task-specific criterion. Given the ground truth labels \( Y_t \) and the predicted logits \( Z_t \), the main loss is defined as:

\begin{equation}
    \mathcal{L}_{t}^{\text{main}} = \mathcal{L}_{\text{task}}(Z_t, Y_t)
\end{equation}

where \( \mathcal{L}_{\text{task}} \) represents the task-specific loss function (\textit{e.g.}, cross-entropy loss with class weighting).

\noindent\textit{Auxiliary Loss}: Auxiliary losses are introduced to enhance feature learning by deep supervision \cite{szegedy2015going} and to prevent the multimodal model from focusing on one specific input data. Let \( \mathcal{M} \) be the set of auxiliary modalities. For each task \( t \), the auxiliary loss is computed as:

\begin{equation}
    \mathcal{L}_{t}^{\text{aux}} = \frac{1}{|\mathcal{M}|} \sum_{m \in \mathcal{M}} \mathcal{L}_{\text{aux}}(Z_t^m, Y_t)
\end{equation}

where \( Z_t^m \) represents the auxiliary logits derived from modality \( m \), and \( \mathcal{L}_{\text{aux}} \) follows the same loss formulation as the main loss.

\noindent\textit{Total Loss Function}: the final loss function balances the main and auxiliary losses with task-specific weights \( w_t \) and a global auxiliary loss weight \( \beta \):

\begin{equation}
    \mathcal{L}_{\text{total}} = \sum_{t \in \mathcal{T}} w_t \left( \mathcal{L}_{t}^{\text{main}} + \beta \mathcal{L}_{t}^{\text{aux}} \right)
\end{equation}

where \( \mathcal{T} \) denotes the set of all tasks, \( w_t \) is the weight associated with task \( t \), and \( \beta \) is a global coefficient that balances auxiliary losses against the main task losses.

\begin{table*}[b]
\caption{{\bf Radiometric statistics of the mono-temporal modalities.} Mean and standard deviation of the mono-temporal modalities across different dataset partitions of the \textit{split\_flairhub}.}
\label{tab:MOD_stats}
\centering
\scriptsize
\renewcommand{\arraystretch}{1.5}
\setlength{\tabcolsep}{2pt}
\begin{tabular}{cccccccccccc}
\toprule
\midrule
& & \multicolumn{2}{c}{TRAIN} &  \multicolumn{2}{c}{VAL} &  \multicolumn{2}{c}{TRAIN$+$VAL} &  \multicolumn{2}{c}{TEST} & \multicolumn{2}{c}{ALL}\\

&           & mean & std   & mean & std &  mean & std &  mean & std & mean & std\\
\midrule
AERIAL\_RGBI & R       & 105.66 & 52.40   & 105.68 & 51.55 &  105.66 & 52.23 &  104.20 & 52.80 & 105.35 & 52.36\\
AERIAL\_RGBI  & G       & 111.36 & 45.81   & 111.30 & 44.87 &  111.35 & 45.62 &  111.16 & 45.65 & 111.31 & 45.63\\
AERIAL\_RGBI  & B       & 102.19 & 44.61   & 102.12 & 43.00 &  102.18 & 44.30 &  101.75 & 44.74 & 102.09 & 44.39\\
AERIAL\_RGBI & I       & 106.83 & 39.71   & 105.64 & 40.06 &  106.59 & 39.78 &  104.36 & 40.82 & 106.12 & 40.01\\
 \midrule
DEM\_ELEV   & DSM    & 321.15 & 549.12   & 270.82 & 486.63 &  311.06 & 537.55 &  384.14 & 523.28 & 326.43 & 535.41\\
DEM\_ELEV  & DTM    & 317.16 & 549.41   & 266.78 & 486.82 &  307.06 & 537.82 &  380.10 & 523.90 & 322.42 & 535.75\\
\midrule
SPOT\_RGBI   & R      & 433.63 & 314.09   &  431.79 & 307.38 &  433.26 & 312.76 &  427.38 & 366.04 & 432.03 & 324.70\\
SPOT\_RGBI   & G      & 509.12 & 286.21   &  507.29 & 278.16 &  508.75 & 284.61 &  502.06 & 352.90 & 507.34 & 300.28\\
SPOT\_RGBI   & B      & 468.3 & 228.49   &   465.67 & 215.86 &  467.77 & 226.02 &  462.90 & 283.88 & 466.75 & 239.36\\
SPOT\_RGBI   & I      & 1\,134.58 & 530.77   & 1\,146.81 & 589.68 &  1\,137.03 & 543.11 &  1\,085.83 & 502.72 & 1\,126.26 & 535.28\\
\midrule
AERIAL-RLT\_PAN  & PAN    & 125.84 & 38.68   & 126.24 & 37.5 &  125.92 & 38.45 &  125.95 & 39.57 & 125.92 & 38.69\\
\bottomrule
\end{tabular}
\end{table*}

\section{Benchmark framework and metric}

\textbf{Framework and settings}: our implementation is based on PyTorch Lightning \cite{PNING}, leveraging the segmentation-models-pytorch (SMP) \cite{smp} library to access pretrained Timm encoders \cite{Timm}. Additionally, we incorporate the U-TAE network from its official repository \cite{UTAE}, using the default architecture but with increased encoder and decoder widths.\\

The model is optimized using AdamW \cite{adamOptimizer}, which decouples weight decay from the optimization step to improve stability. We set the weight decay to 0.01 and use $\beta$ parameters of (0.9, 0.999), which control the moving averages of gradient moments. Learning rate scheduling is managed by OneCycleLR, which dynamically adjusts the learning rate throughout training to improve convergence, incorporating an initial warm-up phase of 20\%. The training procedure spans 150 epochs with a batch size of 5 and an initial learning rate of 0.00005. Experiments are conducted on a high-performance computing (HPC) cluster using 4 to 6 nodes, each equipped with 4 NVIDIA Tesla V100 (32GB memory), A100 or H100 (80GB memory) GPUs. The Distributed Data Parallel strategy in PyTorch Lightning is used to ensure efficient distributed training. Data augmentation techniques include cloud removal and temporal averaging.\\

For the AERIAL\_RGBI and SPOT\_RGBI imagery, only three channels were used in the experiments: Infrared, Red, and Green. Although an initial setup included the four channels, only these three were retained based on performance evaluations. For both modalities and the DEM\_ELEV channels, as well as AERIAL\_RLT-PAN, input data were normalized using the statistics (mean and standard deviation) reported in \Cref{tab:MOD_stats}, which were computed over the combined TRAIN and VAL partitions of the \textit{split\_flairhub}.\\

\textbf{Metric}: The performance of the semantic segmentation models is evaluated using the mean Intersection over Union (mIoU) and Overall Accuracy (O.A.) metrics. For the land-cover task, we exclude the ill-defined classes (see Table~\ref{tab:LABEL_COSIA}) and thus evaluate the results over the remaining 15 classes. For the LPIS crop-type task, two classes, rice and other oilseed crops, are absent from the test set, so mIoU is computed over the remaining classes.

\section{Benchmark results}

\subsection{CNN-based versus Transformer-based Model}

\begin{table*}[!b]
\caption{{\bf Per-Class Evaluation for Land Cover Segmentation – Comparison of Architectures.} 
Class-wise IoU scores for different encoders and decoders baselines using aerial imagery only.}
\label{tab:Baselines_archi}
\scriptsize
\centering
\setlength{\tabcolsep}{4.7pt}
\renewcommand{\arraystretch}{1.3}
\begin{tabular}{ll|cc|ccccccccccccccc|cc}
\toprule
\rotatebox{0}{ENC.}&
\rotatebox{0}{DEC.}&
\rotatebox{80}{mIoU}&
\rotatebox{80}{O.A.}&
\rotatebox{80}{building}&
\rotatebox{80}{greenhouse}&
\rotatebox{80}{pool}&
\rotatebox{80}{imperv.  }&
\rotatebox{80}{pervious}&
\rotatebox{80}{bare soil}&
\rotatebox{80}{water}&
\rotatebox{80}{snow}&
\rotatebox{80}{herbaceous}&
\rotatebox{80}{agriculture}&
\rotatebox{80}{plowed}&
\rotatebox{80}{vineyards} &
\rotatebox{80}{deciduous} &
\rotatebox{80}{coniferous} &
\rotatebox{80}{brushwood} &
\rotatebox{80}{PARA.} &
\rotatebox{80}{EP.}
\\\midrule

\rowcolor{black!5} ResNet50 & UNet & 51.5 & 72.6 & 80.3 & 32.5 & 56.2 & 70.8 & 44.3 & 46.6 & 82.8 & 0.6 & 48.1 & 52.4 & 30.2 & 72.7 & 69.2 & 58.7 & 26.7 & 32.5 & 48 \\ 
ResNet50 & SegFormer & 51.6 & 72.5 & 77.3 & 56.1 & 52.6 & 69.2 & 40.8 & 46.7 & 81.0 & 0.9 & 48.7 & 54.0 & 25.3 & 71.5 & 69.4 & 57.5 & 23.6 & 24.8 & 19 \\ 
\rowcolor{black!5} ResNet50 & DeepLabV3 & 61.7 & 75.5 & 81.1 & 70.1 & 58.0 & 73.1 & 54.0 & 58.6 & 86.8 & 72.1 & 50.8 & 54.9 & 32.6 & 75.1 & 70.4 & 60.6 & 27.0 & 39.6 & 76 \\ 
\rowcolor{black!5} ResNet50 & UPerNet & 58.9 & 74.2 & 80.0 & 71.0 & 53.6 & 72.0 & 52.4 & 57.8 & 85.7 & 47.8 & 48.0 & 53.4 & 33.2 & 74.4 & 69.2 & 58.4 & 27.1 & 30.0 & 49 \\ \hline
\rowcolor{black!5} ConvNeXtV2 & UNet & 64.2 & 77.2 & 84.2 & 76.2 & 60.0 & 75.2 & 56.2 & 63.0 & 89.0 & \bestb{72.5} & 54.2 & 57.1 & 36.3 & 77.5 & 71.3 & 60.4 & 29.3 & 92.8 & 46 \\ 
ConvNeXtV2 & SegFormer & 63.1 & 76.5 & 83.6 & 74.0 & 59.8 & 74.5 & 56.5 & 62.6 & 88.8 & 67.1 & 51.9 & 55.2 & 32.9 & 77.1 & 71.3 & 61.3 & 29.9 & 88.5 & 39 \\ 
\rowcolor{black!5} ConvNeXtV2 & DeepLabV3 & 63.25 & 76.8 & 83.7 & 75.4 & 59.2 & 75.2 & 56.3 & 60.6 & 89.2 & 65.4 & 52.8 & 56.9 & 34.6 & 77.5 & 71.2 & 60.5 & \bestb{30.3} & 96.2 & 35 \\
ConvNeXtV2 & UPerNet & 63.8 & 77.0 & 83.5 & 76.5 & 59.4 & 74.8 & 56.5 & 63.0 & 89.5 & 67.8 & 53.8 & 57.3 & 34.7 & 78.5 & 70.8 & 61.2 & 29.2 & 90.2 & 57 \\ \hline
\rowcolor{black!5} Swin - Base & UNet & \bestb{64.8} & \bestb{77.9} & \bestb{84.7} & \bestb{79.0} & \bestb{62.2} & \bestb{76.2} & 57.5 & \bestb{64.2} & \bestb{90.6} & 63.8 & \bestb{54.9} & \bestb{58.3} & 37.6 & 78.3 & \bestb{72.0} & 62.5 & 30.1 & 92.0 & 100 \\
Swin - Base & SegFormer & 64.4 & 77.4 & 83.5 & 77.2 & 61.1 & 75.4 & 57.4 & 63.4 & 89.2 & 67.3 & 53.5 & 58.0 & \bestb{38.4} & 78.4 & 71.4 & 62.8 & 29.7 & 87.6 & 49 \\
\rowcolor{black!5} Swin - Base & DeepLabV3 & 61.3 & 76.0 & 80.2 & 73.6 & 44.7 & 72.0 & 55.2 & 60.3 & 89.4 & 64.0 & 51.7 & 57.4 & 34.8 & 77.8 & 69.6 & 61.7 & 27.8 & 95.4 & 106 \\ 
\rowcolor{black!5} Swin - Base & UPerNet & 64.1 & 77.5 & 83.9 & 78.4 & 61.6 & 75.7 & 57.2 & 62.9 & 90.3 & 63.4 & 54.3 & 57.1 & 34.8 & 77.7 & 71.7 & 62.6 & 30.2 & 89.4 & 79 \\

\midrule \midrule

\rowcolor{black!5} ResNet50 & UPerNet & 58.9 & 74.2 & 80.0 & 71.0 & 53.6 & 72.0 & 52.4 & 57.8 & 85.7 & 47.8 & 48.0 & 53.4 & 33.2 & 74.4 & 69.2 & 58.4 & 27.1 & 30.0 & 49 \\
ResNext50 & UPerNet & 58.5 & 74.5 & 81.0 & 73.0 & 54.6 & 72.2 & 53.1 & 57.8 & 86.2 & 32.5 & 49.1 & 55.5 & 31.0 & 77.0 & 68.5 & 57.7 & 27.7 & 29.4 & 87 \\
\rowcolor{black!5} HRNet32 & UPerNet & 61.2 & 75.1 & 81.6 & 69.9 & 58.1 & 73.1 & 53.6 & 60.8 & 86.6 & 66.6 & 50.4 & 55.2 & 33.0 & 74.8 & 68.9 & 56.0 & 28.9 & 33.4 & 90 \\
ConvNextV2(t) & UPerNet & 62.7 & 76.4 & 82.6 & 75.3 & 59.1 & 73.8 & 55.1 & 60.2 & 88.6 & 64.8 & 53.2 & 55.8 & 35.4 & 76.1 & 70.9 & 60.6 & 29.5 & 29.8 & 43 \\ 
MiT-B2 & UPerNet & 62.7 & 76.2 & 83.2 & 77.9 & 57.8 & 74.8 & 56.3 & 62.2 & 88.4 & 57.3 & 51.3 & 56.5 & 37.4 & \bestb{79.1} & 69.7 & 58.4 & 30.1 & 25.6 & 81 \\ \hline
\rowcolor{black!5} Swin - Tiny & UPerNet & 62.2 & 76.2 & 82.4 & 72.1 & 58.7 & 74.3 & 55.7 & 60.9 & 88.5 & 64.4 & 52.6 & 55.4 & 30.8 & 76.4 & 70.9 & 60.4 & 28.9 & 29.4 & 111 \\
Swin - Small & UPerNet & 63.2 & 76.9 & 83.5 & 77.0 & 60.8 & 75.0 & 56.4 & 61.4 & 89.4 & 56.8 & 53.5 & 57.1 & 37.7 & 77.6 & 70.9 & 61.9 & 28.7 & 50.7 & 114 \\
\rowcolor{black!5} Swin - Base & UPerNet & 64.1 & 77.5 & 83.9 & 78.4 & 61.6 & 75.7 & 57.2 & 62.9 & 90.3 & 63.4 & 54.3 & 57.1 & 34.8 & 77.7 & 71.7 & 62.6 & 30.2 & 89.4 & 79 \\
Swin - Large & UPerNet & \bestb{64.8} & 77.7 & 84.1 & 77.4 & 61.5 & 75.9 & \bestb{57.6} & 64.1 & 90.4 & 68.5 & 54.4 & 58.2 & 36.1 & 79.0 & 71.7 & \bestb{63.0} & 30.2 & 199.4 & 106 \\ 

\bottomrule 
\end{tabular}
\end{table*}

In \Cref{tab:Baselines_archi}, we compare different architectures for the encoder and decoder part using aerial imagery only (3 bands: infrared, red, and green) as input. For this configuration, we evaluate several seminal encoder backbones, including convolutional neural networks (CNNs) such as ResNet \cite{ResNet}, ResNeXt \cite{resnext}, ConvNeXtV2 \cite{woo2023convnext} and HRNet \cite{hrnet}, as well as transformer-based models such as the Swin Transformer \cite{liu2021swin} and the Mix Transformer (MiT-BX) encoders \cite{xie2021segformer}. To ensure a fair comparison, we selected versions of these architectures with a similar number of parameters. For the Swin Transformer, we evaluated different model sizes, from "Tiny" to "Large", to explore the trade-off between model complexity and performance. The Tiny variant is optimized for lightweight and fast inference, while the Large variant prioritizes accuracy at the cost of increased computational load. On the decoder side, we compare four well-established architectures: UNet \cite{U-Net}, SegFormer \cite{xie2021segformer}, DeepLabV3 \cite{chen2017rethinking} and UPerNet \cite{upernet}.

\noindent\textbf{Decoder performance across fixed encoders}: In the first three blocks of results in \Cref{tab:Baselines_archi}, we alternately fixed the encoder (ResNet50, ConvNeXtV2, and Swin-Base) while varying the decoder (UNet, SegFormer, DeepLabV3, UPerNet). The choice of decoder is particularly important when using the ResNet-50 encoder. Indeed, there is a 10.2\% gap in mIoU and a 2.9\% difference in OA between the best and worst-performing decoders. Spatial context is known to be very important in the case of semantic segmentation for land cover classification. As a consequence, decoders that incorporate mechanisms to enhance spatial context, such as DeepLabV3 and UPerNet, outperform U-Net. These results had already been observed in \cite{garioud_flair_2023a} for a ResNet-34 encoder and FPN, DeepLabV3 decoders. Although the SegFormer decoder is designed to capture spatial context, its relatively lower parameter count appears to limit its ability to produce strong results. The impact of decoder choice on performance is considerably reduced when using Swin-Base or ConvNeXtV2 encoders. With the ConvNeXtV2 encoder, the performance difference between decoders is only 1.1\% in mIoU and 0.7\% in OA; with the Swin-Base encoder, the difference is 3.5\% in mIoU and 1.9\% in OA. We hypothesize that when the encoder effectively captures spatial context and learns spatial relationships between objects, the specific decoding strategy becomes less critical. Notably, with these encoders, the U-Net decoder consistently achieves the best results.

\noindent\textbf{Encoder performance with fixed decoder}: In the fourth block of results in \Cref{tab:Baselines_archi}, the decoder is fixed to UPerNet, while several encoders are tested to evaluate their impact on performance. Among the encoders, the ConvNextV2 yields the best performance with approximately 30\:M parameters, comparable in size to the Swin Tiny. As expected, transformer-based models, such as Swin-Tiny and MiT-B2,  outperform traditional CNN-based architectures,such as ResNet50, ResNeXt50, and HRNet32, confirming recent trends in semantic segmentation literature. However the ConvNextV2 demonstrates competitive performance with Swin Transformers, despite being a convolution-based architecture. This highlights the significance of modern optimization paradigms compared to new module (such as attention), which appear to play a crucial role in achieving state-of-the-art results.

Despite this, we ultimately selected the Swin Transformer as our default encoder due to its favourable balance between accuracy and computational efficiency across different configurations, and its greater maturity in modular integration with existing frameworks. We opted for UPerNet as the decoder in our final pipeline due to its flexibility and compatibility with multi-scale feature fusion. Moreover, the Swin-UPerNet architecture won the FLAIR \#1 challenge \cite{FLAIR-1} and, we observed that the hierarchical nature of Swin encoders helps to mitigate stitching issues between overlapping patches during inference on very large areas and demonstrates greater robustness within our production lines. 

\noindent\textbf{Scaling Swin encoders: performance vs. complexity}: Finally, in the fifth part of \Cref{tab:Baselines_archi}, we evaluate the impact of encoder size by testing variants of the Swin encoder architecture. The best performance is achieved with the Swin-Large encoder. This can be attributed to the large volume of annotations available in the \FLAIRHUB dataset, which supports training more complex models. However, the performance gain compared to the Swin-Base encoder is modest, with only +0.7\% in mIoU and +0.2\% in OA, despite Swin Large having more than twice the number of parameters. This trade-off between model complexity and performance led us to adopt the Swin-Base encoder as the baseline configuration.

\begin{table*}[b]
\caption{\textbf{Quantitative Evaluation for Land Cover Segmentation.}
Performance of the UPerFuse architecture with different input modalities during training and testing. Auxiliary losses are used in configurations with more than one modality. PARA.: number of model parameters (in millions).\; EP.: epoch with best validation score.\; SITS: Satellite Image Time Series.\; S1/2: Sentinel-1/2.
}
\label{tab:Baselines_LC}
\centering
\scriptsize
\renewcommand{\arraystretch}{1.5}
\setlength{\tabcolsep}{3.5pt}
\scriptsize{
    \begin{tabular}{l|cccccc|cc|cc}
    \toprule
        Model ID & Aerial VHR & Elevation & SPOT & S2 SITS & S1 SITS & Historical & PARA. & EP. & O.A. & mIoU \\ 
        \midrule 
         LC-A & \checkmark & & & & & & 89.4  &  79 & 77.5 & 64.1 \\
         \rowcolor{black!5}   LC-B & \checkmark & \checkmark & & &  & & 181.4 & 124 & 78.1 & 65.1 \\
         LC-C & \checkmark & \checkmark & \checkmark  & & & & 270.6 & 131 & 78.2 & 65.2 \\
         \rowcolor{black!5} LC-D & \checkmark &  &   & \checkmark & & & 93.9 & 85 & 77.6 & 64.7 \\
         LC-E & \checkmark & &  & & \checkmark & & 95.8 & 98 & 77.7 & 64.5 \\
         \rowcolor{black!5} LC-F & \checkmark & &  &  \checkmark & \checkmark & &  97.7 & 64 & 77.7 & 64.9 \\
         LC-G &  & &  &  \checkmark &  & &  0.9 & 89 & 57.8 & 34.2 \\
         \rowcolor{black!5} LC-H &  & &  &  & \checkmark & &  1.8 & 106 & 54.5 & 28.2 \\
         LC-I &  & &  \checkmark &  &  & &  89.2 & 94 & 64.1 & 43.5 \\
         \rowcolor{black!5} LC-J &  & \checkmark &  & &  & & 89.4 & 97 & 67.4 & 51.2 \\
         LC-K & \checkmark & &  &  &  & \checkmark &  181.4 & 45 & 77.6 & 64.3 \\
         \rowcolor{black!5} LC-L & \checkmark & \checkmark & \checkmark & \checkmark  & \checkmark & &  276.4 & 121 & \bestb{78.2} &  \bestb{65.8}\\
         LC-ALL & \checkmark & \checkmark & \checkmark  & \checkmark & \checkmark & \checkmark &  365.8 & 129 & \bestb{78.2} & 65.6 \\
        \bottomrule
    \end{tabular}
}
\end{table*}

\begin{table*}[!b]
\caption{\textbf{Per-Class Evaluation for Land Cover Segmentation.} 
Per-class performance of the UPerFuse architecture with different input modalities. Auxiliary losses are used in configurations with more than one modality. 
%PARA.: number of model parameters (in millions).\; EP.: epoch with best validation score.\; t.s.: time series.\; S1/2: Sentinel-1/2.
}

\label{tab:Baselines_LC_classes}
\centering
\scriptsize
\renewcommand{\arraystretch}{1.5}
\setlength{\tabcolsep}{3.5pt}
\scriptsize{
    \begin{tabular}{l|c|cccccccccccccccc}
    \toprule
        Model ID & 
        mIoU &
        \rotatebox{80}{building}&
        \rotatebox{80}{greenhouse}&
        \rotatebox{80}{pool}&
        \rotatebox{80}{imperv.}&
        \rotatebox{80}{pervious}&
        \rotatebox{80}{bare soil}&
        \rotatebox{80}{water}&
        \rotatebox{80}{snow}&
        \rotatebox{80}{herbaceous}&
        \rotatebox{80}{agriculture}&
        \rotatebox{80}{plowed} &
        \rotatebox{80}{vineyards} &
        \rotatebox{80}{deciduous} &
        \rotatebox{80}{coniferous} &
        \rotatebox{80}{brushwood} \\
        \midrule 
        
         LC-A & 64.1 & 83.9 & 78.4 & 61.6 & 75.7 & 57.2 & 62.9 & 90.3 & 63.4 & 54.3 & 57.1 & 34.8 & 77.7 & 71.7 & 62.6 & 30.2 \\
         \rowcolor{black!5}  LC-B & 65.1 & 85.1 & \bestb{80.4} & 62.5 & 76.2 & 58.1 & 64.1 & 90.8 & 62.6 & \bestb{55.2} & 57.8 & 37.8 & \bestb{79.6} & 72.3 & 63.1 & 31.3 \\
         LC-C & 65.2 & 85.2 & 79.1 & 62.1 & 76.4 & 58.3 & 64.8 & \bestb{90.9} & 64.4 & 55.1 & 58.4 & 37.4 & 78.6 & 72.3 & 63.4 & \bestb{31.8} \\
         \rowcolor{black!5} LC-D & 64.7 & 84.0 & 78.9 & 61.2 & 75.8 & 57.5 & 63.0 & 90.5 & 68.3 & 54.4 & 57.5 & 36.9 & 78.1 & 71.9 & 62.9 & 29.4 \\
         LC-E & 64.5 & 84.1 & 78.9 & 62.0 & 76.0 & 57.6 & 63.7 & 90.6 & 62.7 & 54.7 & 57.4 & 37.1 & 78.1 & 71.9 & 63.1 & 30.2 \\
         \rowcolor{black!5} LC-F & 64.9 & 84.0 & 79.3 & 61.1 & 75.6 & 57.7 & 63.8 & 90.5 & 68.1 & 54.9 & 56.9 & 37.9 & 78.1 & 71.7 & 63.7 & 29.6 \\
         LC-G & 34.2 & 34.9 & 0.0 & 0.0 & 38.3 & 27.4 & 33.6 & 65.3 & 67.5 & 34.4 & 42.1 & 10.2 & 41.1 & 56.0 & 48.2 & 14.5 \\
         \rowcolor{black!5} LC-H & 28.2 & 42.4 & 1.3 & 0.0 & 35.7 & 23.0 & 36.8 & 57.5 & 10.7 & 29.4 & 42.3 & 5.5 & 25.2 & 53.1 & 46.5 & 13.7 \\
         LC-I & 43.5 & 57.2 & 49.8 & 13.8 & 53.2 & 40.0 & 44.0 & 71.0 & 62.0 & 36.9 & 48.2 & 4.6 & 42.1 & 58.7 & 52.9 & 18.1 \\
         \rowcolor{black!5} LC-J & 51.2 & 76.1 & 70.3 & 27.0 & 58.5 & 38.8 & 49.6 & 82.1 & 81.7 & 37.5 & 50.9 & 11.9 & 50.4 & 63.3 & 46.0 & 23.8 \\
         LC-K & 64.3 & 83.8 & 77.6 & 59.4 & 75.5 & 57.4 & 63.1 & 90.0 & 62.6 & 53.5 & 57.9 & \bestb{38.0} & 78.6 & \bestb{72.5} & \bestb{64.2} & 30.7 \\
         \rowcolor{black!5} LC-L & \bestb{65.8} & \bestb{85.3} & 79.1 & 62.0 & \bestb{76.6} & 58.2 & 64.7 & 90.5 & \bestb{73.4} & 55.1 & \bestb{58.6} & 37.5 & 78.6 & 72.3 & 63.5 & 31.1\\
         LC-ALL & 65.6 & \bestb{85.3} & 80.3 & \bestb{62.8} & 76.5 & \bestb{58.5} & \bestb{65.1} & 90.8 & 67.6 & 55.0 & \bestb{58.6} & 37.9 & 78.4 & 72.3 & 63.2 & 31.4 \\
        \bottomrule
    \end{tabular}
}
\end{table*}

\subsection{Multimodality Fusion}
\vspace{0.5em}

\noindent\textit{B.1 Land Cover Mapping}
\vspace{0.5em}

In \Cref{tab:Baselines_LC}, we report land cover segmentation performance across various combinations of input modalities, using a fixed UPerFuse architecture (see \Cref{sec:baseline_archi}) and identical hyperparameters. The best results are achieved when incorporating nearly all available modalities (denoted LC-L), reaching 78.2\% OA and 65.8\% mIoU. Notably, the inclusion of historical imagery appears to slightly reduce performance (LC-L vs. LC-ALL). This outcome is expected, since historical images were primarily included to support future transfer learning tasks. Their purpose is to enable models trained on recent very high-resolution (VHR) data to generalize to much older imagery. However, the domain gap and temporal shift of these images introduces noise that affects segmentation accuracy.\\

\noindent\textbf{LC-A vs. LC-L}: Adding all available modalities yields only marginal improvements compared to using the Aerial VHR modality alone (OA: +0.9\%, mIoU: +1.7\%). Users applying the models in production environments must carefully consider whether these modest gains justify the additional complexity and preprocessing effort required for multimodal data. However, the current limited improvement is likely explained by the annotation process, which was performed on the Aerial VHR images. As a result, this modality benefits from strong geometric and temporal consistency with the reference data, which likely facilitates model learning and performance.\\

\noindent\textbf{LC-A vs. LC-B}: Adding elevation information to the Aerial VHR images provides a consistent improvement in performance (OA: +0.6\%, mIoU: +1.0\%). As shown in \Cref{tab:Baselines_LC_classes}, this benefit is also observed at the class level. Nearly all classes show increased IoU scores when 3D information is included. The only exception is the snow class, for which IoU decreases. While altitude data is expected to reduce class confusion, its effectiveness appears limited in regions with low visual texture. In such cases, noise and artefacts in the elevation data may introduce additional sources of confusion, offsetting potential gains.\\

\noindent\textbf{LC-B vs. LC-C}: Incorporating SPOT imagery alongside VHR and elevation data is expected to provide additional value for two main reasons. First, the availability of a second temporal observation of ground objects, even at a coarser 1.6-meter resolution, may assist in distinguishing certain classes. Second, and more importantly, SPOT data are absolutely calibrated in ground surface reflectance. This radiometric consistency has the potential to reduce discrepancies between the radiometric properties of various VHR aerial images. Despite these theoretical advantages, the addition of SPOT data does not lead to any notable improvement in segmentation performance (OA: +0.1\%, mIoU: +0.1\%).\\

\noindent\textbf{LC-A vs. LC-D.E.F}: The addition of time series modalities results in only limited performance gains. Specifically, including Sentinel-2 time series (LC-D) yields an increase of 0.1\% in overall accuracy and 0.6\% in mIoU. Adding Sentinel-1 time series (LC-E) leads to a gain of 0.2\% in overall accuracy and 0.4\% in mIoU, while the use of Sentinel-1 and Sentinel-2 (LC-F) does not provide a change in overall accuracy and an increase of 0.4\% in mIoU. In theory, temporal information could help reduce confusion between certain land cover classes, such as deciduous and coniferous forests or herbaceous and agricultural areas. However, the observed improvements are minimal and as shown in \Cref{tab:Baselines_LC_classes}, the limited contributions are consistently small across classes and across the different temporal modalities (S1, S2, and S1+S2). The only notable exception is the plowed class, for which mIoU increases by 2.1\% with S2 images, 2.3\% with S1 images, and 3.1\% when both are used. This finding highlights a form of complementarity between optical and radar time series for this specific class. The results for LC-D can also be compared with those from FLAIR \#2 \cite{FLAIR-2,garioud_flair_2023a}, where the addition of Sentinel-2 images led to larger improvements: an mIoU gain of 1.23\% using fixed hyperparameters, and up to 3.90\% when comparing best runs and optimized hyperparameter settings. A key difference between FLAIR-HUB and FLAIR \#2 is the removal of the super-patch data, which involved the use of larger spatial footprints for Sentinel-2 patches. This suggests that the performance gains observed in FLAIR \#2 were more likely due to an expanded spatial context rather than the model’s ability to learn temporal features.\\

\noindent\textbf{LC-A vs. LC-G.H.I.J}: When using only a single input modality, we can observe that the LC-A (Aerial VHR) configuration performs very well, coming close to the best results achieved with multiple sources of information. These results indicate that the shape and texture of land cover objects, well captured by very high spatial resolution, are key variables for their discrimination. The single-modality temporal configurations Sentinel-2 (LC-G) and Sentinel-1 (LC-H) are less effective, with respective mIoU scores of 34.2\% and 28.2\%. These results are disappointing but can be explained by the fact that the size of objects in several land cover classes from the nomenclature is on the same scale or smaller than the spatial resolution of Sentinel-2 and Sentinel-1 images. The class-wise metrics (\Cref{tab:Baselines_LC_classes}) highlight, for example, the inability to effectively learn the \textit{greenhouse} and \textit{swimming pool} classes. Furthermore, it is worth noting that the chosen baseline for encoding temporal information, UTAE, has very few parameters (on the order of one million), which is significantly smaller compared to the Swin-Base baseline, with around 90 million parameters. The results achieved using only the SPOT source (LC-I) are satisfying with 64.1\% of OA and 43.5\% mIoU. We still observe difficulties in learning classes with small objects, while the IoUs for the other classes remain close to the aerial configuration. Furthermore, the land cover annotation is temporally and spatially consistent with the Aerial VHR source. Therefore, it is not a true reference for computing metrics on the SPOT image. This is reflected in the poor results for the \textit{plowed} class. The configuration using only the Elevation modality (LC-J) achieves good results, with 67.4\% OA and 51.2\% mIoU. The first channel of this modality, the DSM (Digital Surface Model), is derived from the same source images as those used to produce the Aerial VHR modality. However, we observed that combining Aerial VHR with Elevation yields results very similar to using Aerial VHR alone. This suggests that features such as object texture and shape can also be learned from the Elevation modality alone. The addition of color remains crucial for several classes, such as \textit{swimming pools}, differences between types of non-vegetated soils (impervious, pervious, or plowed), and for distinguishing various types of vegetation (\textit{vineyards}, \textit{coniferous}).\\

\begin{table*}[!b]
\caption{{\bf Per-Class Evaluation for Land Cover Segmentation – Ablation Study.} 
Class-wise IoU scores for the Swin Base-UP baseline using aerial imagery and Sentinel-1/2 time series (denoted setting LC-F). Results include mean and standard deviation over a 5-fold training and evaluation procedure.
}
\label{tab:Baselines_LC-F}
\scriptsize
\centering
\setlength{\tabcolsep}{5.2pt}
\renewcommand{\arraystretch}{1.3}
\begin{tabular}{ll|cc|ccccccccccccccc|cc}
\toprule
\rotatebox{0}{Model ID}&
\rotatebox{0}{CONF.} &
\rotatebox{80}{mIoU}&
\rotatebox{80}{O.A.}&
\rotatebox{80}{building}&
\rotatebox{80}{greenhouse}&
\rotatebox{80}{pool}&
\rotatebox{80}{imperv.}&
\rotatebox{80}{pervious}&
\rotatebox{80}{bare soil}&
\rotatebox{80}{water}&
\rotatebox{80}{snow}&
\rotatebox{80}{herbaceous}&
\rotatebox{80}{agriculture}&
\rotatebox{80}{plowed} &
\rotatebox{80}{vineyards} &
\rotatebox{80}{deciduous} &
\rotatebox{80}{coniferous} &
\rotatebox{80}{brushwood} &
\rotatebox{80}{PARA.} &
\rotatebox{80}{EP.}
\\\midrule
\multicolumn{19}{c}{Evaluation on Fold 1} \\
\midrule
LC-F & - & 64.3 & 77.5 & \bestb{84.1} & 78.3 & 61.5 & 75.8 & 57.1 & 63.7 & 90.5 & 61.8 & 53.9 & 56.9 & 38.0 & 78.1 & \bestb{71.8} & 63.1 & 29.4 & 95.2 & 72 \\
\rowcolor{black!5}  & + dropout & 64.3 & 77.3 & 83.1 & 77.4 & 57.5 & 75.0 & 57.5 & 63.6 & 89.3 & 71.2 & 54.7 & 56.6 & 36.4 & \bestb{78.8} & 71.6 & 62.2 & \bestb{29.8} & 95.2 & 61 \\
 & + auxloss & \bestb{64.9} & 77.7 & 84.0 & \bestb{79.3} & 61.1 & 75.6 & \bestb{57.7} & 63.8 & 90.5 & 68.1 & \bestb{54.9} & 56.9 & 37.9 & 78.1 & 71.7 & \bestb{63.7} & 29.6 & 97.7 & 64 \\
\rowcolor{black!5}  &  + sentemp & 64.7 & \bestb{77.8} & \bestb{84.1} & 78.3 & \bestb{62.0} & \bestb{75.9} & 57.6 & 63.9 & \bestb{91.0} & 64.9 & 54.7 & \bestb{58.4} & \bestb{38.2} & 78.3 & 71.7 & 63.0 & 29.3 & 95.2 & 86 \\ 
 & + aux \& drop & \bestb{64.9} & 77.3 & 83.4 & 78.3 & 58.4 & 75.1 & 57.1 & \bestb{65.5} & 90.6 & \bestb{76.3} & 53.3 & 57.2 & 36.5 & 78.3 & \bestb{71.8} & 62.3 & 29.2 & 97.7 & 79 \\
\midrule
\multicolumn{19}{c}{5-fold Evaluation} \\
\midrule
\rowcolor{black!5} LC-F  &  split\_1 & 64.3 & 77.5 & 84.1 & 78.3 & 61.5 & 75.8 & 57.1 & 63.7 & 90.5 & 61.8 & 53.9 & 56.9 & 38.0 & 78.1 & 71.8 & 63.1 & 29.4 & 95.2 & 72 \\
LC-F  &  split\_2 & 67.4 & 79.7 & 82.7 & 58.2 & 57.4 & 75.1 & 59.5 & 67.2 & 90.5 & 92.8 & 54.4 & 65.2 & 47.0 & 81.1 & 76.2 & 62.4 & 41.0 & 95.2 & 43 \\
\rowcolor{black!5} LC-F  &  split\_3 & 66.9 & 78.6 &  84.6 & 75.9 & 60.0 & 75.4 & 59.7 & 69.2 & 86.4 & 78.0 & 54.1 & 62.3 & 42.5 & 82.4 & 73.2 & 59.9 & 39.9 & 95.2 & 66\\
LC-F  &  split\_4 & 69.1 & 80.0 & 86.1 & 78.0 & 60.0 & 75.0 & 52.4 & 73.4 & 91.6 & 81.2 & 55.3 & 67.5 & 53.9 & 89.0 & 75.2 & 63.8 & 33.5 & 95.2 & 49 \\
\rowcolor{black!5} LC-F  &  split\_5 & 66.2 & 79.2 & 84.7 & 73.5 & 60.6 & 76.6 & 58.8 & 59.8 & 88.7 & 62.4 & 55.0 & 63.4 & 53.7 & 82.4 & 75.1 & 63.1 & 34.9 & 95.2 & 78 \\ \cline{3-21}
 & Average & 66.8 & 79.0 & 84.4 & 72.8 & 60.0 & 75.6 & 57.5 & 66.7 & 89.5 & 75.2 & 54.5 & 63.1 & 47.0 & 82.6 & 74.3 & 62.5 & 35.7 & - & 62 \\
\rowcolor{black!5}  & $\pm$ & 1.6 & 0.9 & 1.1 & 7.5 & 1.4 & 0.6 & 2.7 & 4.6 & 1.8 & 11.8 & 0.5 & 3.5 & 6.2 & 3.6 & 1.6 & 1.4 & 4.3 & - & 13 \\
\bottomrule 
\end{tabular}
\end{table*}

\begin{table*}[t]
\caption{\textbf{Quantitative Evaluation for Land Cover Segmentation.}  
Performance of the best configuration (UPerNet architecture) on the FLAIR \#1 \cite{FLAIR-1} and FLAIR \#2 (also FLAIR) \cite{FLAIR-2,garioud_flair_2023a} test sets. The metrics are computed only on 12 classes (and not 15). PARA.: number of model parameters (in millions).
}
\label{tab:Baselines_Legacy}

\centering
\scriptsize
\renewcommand{\arraystretch}{1.5}
\setlength{\tabcolsep}{5pt}
\scriptsize{
    \begin{tabular}{lcrrcccccc}
    \toprule
    \multirow{ 2}{*}{Model} & \multirow{ 2}{*}{Input Data} & \multirow{ 2}{*}{\makecell[l]{Supervision\\ Pixel $\times 10^9$}} & \multirow{ 2}{*}{PARA.} & \multicolumn{2}{c}{FLAIR \#1} & &  \multicolumn{2}{c}{FLAIR \#2} \\ \cline{5-6} \cline{8-9}
     &  &  & & \multicolumn{2}{c}{mIoU} & &  \multicolumn{2}{c}{mIoU}  \\ \midrule 
\rowcolor{black!5}    U-Net \cite{FLAIR-1} & aerial, Topo & 20.3 & 24.4 & \multicolumn{2}{c}{55.7} & & \multicolumn{2}{c}{X}  \\
    U-T\&T \cite{garioud_flair_2023a} & aerial, Topo, Sentinel-2 & 20.3 & 27.3 & \multicolumn{2}{c}{X} & & \multicolumn{2}{c}{58.6} \\
\rowcolor{black!5}    UPerFuse (LC-L) & aerial, Topo, SPOT, Sentinel-1\&-2 & 63.2 & 276.3 & \multicolumn{2}{c}{64.1} & & \multicolumn{2}{c}{65.0} \\ 
    \bottomrule
    \end{tabular}
}
\end{table*}

In \Cref{tab:Baselines_LC-F}, we report the performance of various network architectures under the LC-F configuration, with and without enhancement strategies. These strategies include modality dropout (dropout), auxiliary losses (auxloss), and monthly temporal averaging of time series inputs (sentemp). To evaluate the variability of the training process, we also include the results of a 5-fold cross-validation, which are presented in the final rows of \Cref{tab:Baselines_LC-F}.

Enhancement strategies yield modest and inconsistent improvements across classes. The auxiliary loss configuration achieves the highest overall mIoU (64.9\%) and shows slight gains for classes such as greenhouse (+1.0\%), snow (+6.3\%), herbaceous (+1.0\%), and coniferous (+0.6\%). These results suggest that auxiliary losses may support better training dynamics by improving gradient flow and promoting more effective use of multimodal inputs. Temporal averaging yields the highest overall accuracy (77.8\%) and small gains for classes like pool, impervious surfaces, agriculture, and plowed fields. While the latter improves slightly (from 38.0\% to 38.2\%), the limited magnitude of these changes makes it difficult to draw strong conclusions about the specific benefits of temporal averaging. The modality dropout configuration does not lead to significant overall improvement. Some class-level variations are observed, such as higher IoU for snow (+9.4\%), but these may reflect training variance rather than a consistent effect. When combined with auxiliary loss, performance again reaches 64.9\% mIoU, though with a different distribution of class-level gains. Overall, the enhancements show only limited impact, and their contributions appear to be context-dependent.

The 5-fold evaluation highlights the impact of dataset partitioning on model performance (see \Cref{sec:dataset_partitions} for details). While the average mIoU across folds is 66.8\% with a standard deviation of $\pm$\:1.6, split\_1 stands out with notably lower scores (mIoU: 64.3\%, OA: 77.5\%) compared to the other folds, which exceed 66\% mIoU and reach up to 69.1\%. This drop is likely due to the larger validation set in split\_1, which combines FLAIR \#1 and FLAIR \#2 test sets domains. Substantial variability is noticeable at the class level. For instance, snow IoU ranges from 61.8\% in split\_1 to 92.8\% in split\_2, largely due to differences in the presence of snow-covered areas in the training, validation and test sets. This uneven representation directly impacts the learning and evaluation and contributes to the observed fluctuation. Similarly, plowed class increases from 38.0\% to 53.9\%. Other classes such as greenhouse ($\pm$\:7.5) and bare soil ($\pm$\:4.6) also exhibit marked differences, indicating that spatial and seasonal heterogeneity in both training and test splits significantly influences class-level generalization.

\Cref{tab:Baselines_Legacy} reports the performance of our best configuration, UPerFuse (LC-L), on the FLAIR \#1 \cite{FLAIR-1} and FLAIR \#2 \cite{FLAIR-2, garioud_flair_2023a} test sets. The model achieves 64.1\% mIoU on FLAIR \#1, compared to 55.7\% with U-Net, and 65.0\% on FLAIR \#2, compared to 58.6\% with U-T\&T. These improvements are notable but must be interpreted with care: UPerFuse was trained on a much larger dataset, with nearly three times more annotated pixels, and supervision was applied over 15 classes, although the metrics here are computed on 12. The input configuration also includes additional modalities such as SPOT and Sentinel-1, which were not used in the previous baselines. The gains observed are therefore the result of several combined factors, including model capacity, input diversity, and training data volume.\\

\noindent\textit{B.2 Crop Type Mapping}
\vspace{0.5em}

\begin{table*}[b]
\caption{\textbf{Quantitative Evaluation for crop mapping.}  
Performance of the UPerFuse architecture with different input modalities. PARA.: number of model parameters (in millions).\; EP.: epoch with best validation score.\; SITS: Satellite Image Time Series.\; S1/2: Sentinel-1/2. Classes with zero pixels in the test set are excluded from mIoU computation.
}

\label{tab:Baselines_LPIS}
\centering
\scriptsize
\renewcommand{\arraystretch}{1.5}
\setlength{\tabcolsep}{3.5pt}
\scriptsize{
    \begin{tabular}{l|cccc|rc|cc}
    \toprule
        Model ID & Aerial VHR & SPOT & S2 SITS & S1 SITS & PARA. & EP. & O.A. & mIoU \\ 
        \midrule 
        \multicolumn{8}{c}{LV.1 - 23 classes (2 classes removed)} \\
        \midrule 
        LPIS-A & \checkmark & & & & 89.4 & 91 & 86.6 & 24.4 \\
        \rowcolor{black!5} LPIS-B & \checkmark & \checkmark & & & 181.2 & 99 & 87.1 & 26.1 \\
        LPIS-C & \checkmark & & \checkmark & & 93.9 & 100 & 87.5 & 29.8 \\
        \rowcolor{black!5} LPIS-D & \checkmark & & \checkmark & \checkmark & 97.7 & 80 & \bestb{88.0} & 36.1 \\
        LPIS-E & \checkmark & \checkmark & \checkmark & & 183.1 & 46 & 87.6 & 30.3 \\
        \rowcolor{black!5} LPIS-F & & & \checkmark & &  0.9 & 62 & 85.3 & 23.8 \\
        LPIS-G & & & & \checkmark &  1.8 & 77 & 84.5 & 18.1 \\
        \rowcolor{black!5} LPIS-H & & & \checkmark & \checkmark & 2.8 & 61 & 84.9 & 23.8 \\
        LPIS-I & & \checkmark & \checkmark & \checkmark & 97.5 & 49 & 87.2 & \bestb{39.2} \\
        \rowcolor{black!5} LPIS-J & \checkmark & \checkmark & \checkmark & \checkmark & 186.9 & 53 & \bestb{88.0} & 35.4 \\
        LPIS-K & & \checkmark & & & 89.2 & 14 & 84.5 & 15.1 \\
        \midrule
        \multicolumn{8}{c}{LV.2 - 31 classes (3 classes removed)} \\
        \midrule
        LPIS-I & & \checkmark & \checkmark & \checkmark& 97.5 & 74 & 87.5 & 29.6 \\
        \midrule
        \multicolumn{8}{c}{LV.3 - 46 classes (8 classes removed)} \\
        \midrule
        LPIS-I & & \checkmark & \checkmark & \checkmark & 97.5 & 111 & 87.3 & 21.4 \\
        \bottomrule
    \end{tabular}
}
\end{table*}
\begin{table*}[t]
\caption{{\bf Per-Class Evaluation for 1st Level Crop Mapping.}  
Class-wise IoU scores for the Base-UP baseline with different input modalities. Auxiliary losses are used in configurations with more than one modality. The \textit{rice} and other \textit{oilseed crops} classes are excluded from mIoU computation due to having zero pixels in the test set.
}
\label{tab:LPIS_per_class_perf}
\centering
\scriptsize
\renewcommand{\arraystretch}{1.5}
\setlength{\tabcolsep}{3.5pt}
\scriptsize{
    \begin{tabular}{l|c|ccccccccccccccccccccc}
    \toprule
        
        \makecell[l]{Model ID}& 
        \rotatebox{80}{mIoU} &
        \rotatebox{80}{grasses}  & 
        \rotatebox{80}{wheat}  & 
        \rotatebox{80}{barley}    &
        \rotatebox{80}{maize} & 
        \rotatebox{80}{o. cereals}  & 
        \rotatebox{80}{\makecell[l]{flax/hemp\\tobacco}}  & % flax/hemp/tobacco 
        \rotatebox{80}{sunflower}  & 
        \rotatebox{80}{rapeseed}   &  
       % \rotatebox{80}{other oilseed crops} & 
        \rotatebox{80}{soy}   & 
        \rotatebox{80}{o. protein c.}  & % other protein crops
        \rotatebox{80}{\makecell[l]{fodder\\legumes}}     & 
        \rotatebox{80}{beetroots} & 
        \rotatebox{80}{potatoes} & 
        \rotatebox{80}{o. arable c.}     & 
        \rotatebox{80}{vineyard}     & 
        \rotatebox{80}{olive groves}  & 
        \rotatebox{80}{\makecell[l]{fruits\\orchards}}& 
        \rotatebox{80}{nut orchards}     & 
        \rotatebox{80}{o. permanent c.} & 
        \rotatebox{80}{mixed c.}    &  
        \rotatebox{80}{background}   \\ 
        \midrule 
        \rowcolor{black!5}LPIS-A & 24.4 & 49.4 & 34.2 & 13.1 & 60.5 & 3.5 & 2.7 & 12.6 & 38.0 & 0.0 & 3.1 & 13.3 & 53.9 & 7.5 & 19.7 & 43.4 & 13.5 & 36.8 & 2.9 & \bestb{14.8} & 1.5 & 88.6 \\
        LPIS-B & 26.1 & 50.1 & 41.8 & 16.3 & 62.0 & 3.1 & 2.3 & 16.6 & 45.1 & 0.1 & 3.0 & 22.6 & 57.9 & 11.6 & 14.7 & 42.2 & 14.4 & \bestb{37.5} & 1.9 & 13.8 & 3.2 & \bestb{88.8} \\
        \rowcolor{black!5}LPIS-C & 29.8 & 49.9 & 50.1 & 27.8 & 75.1 & 5.5 & 2.7 & 22.0 & 58.7 & 11.9 & 4.8 & 26.2 & 68.1 & 9.6 & 17.2 & 41.2 & \bestb{19.5} & 36.1 & 1.5 & 7.1 & 1.3 & 88.7 \\
        LPIS-D & 36.1 & 51.3 & 59.9 & 40.9 & 77.5 & 7.0 & 9.3 & \bestb{50.2} & 77.8 & 28.7 & 10.1 & 24.3 & \bestb{79.6} & 7.4 & 20.2 & 42.3 & 13.0 & 36.7 & 15.8 & 12.8 & 4.0 & 88.7 \\
        \rowcolor{black!5}LPIS-E & 30.3 & 50.4 & 48.7 & 20.1 & 76.2 & 3.7 & 0.6 & 26.6 & 63.4 & 7.9 & 8.5 & 24.9 & 75.9 & 11.4 & \bestb{25.8} & 41.1 & 9.4 & 35.6 & 1.9 & 13.5 & 2.6 & \bestb{88.8} \\
        LPIS-F & 23.8 & 43.1 & 59.6 & \bestb{48.8} & 68.8 & 2.6 & 0.0 & 28.0 & 70.9 & 12.4 & \bestb{20.9} & 22.8 & 1.5 & 0.0 & 10.1 & 24.5 & 0.0 & 0.0 & 0.0 & 0.0 & 0.0 & 86.3 \\
        \rowcolor{black!5}LPIS-G & 18.1 & 37.0 & 53.9 & 10.6 & 45.2 & 10.2 & 0.0 & 20.5 & 68.1 & 2.7 & 2.0 & 12.5 & 0.0 & 0.0 & 3.9 & 20.2 & 0.0 & 8.2 & 0.0 & 0.0 & 0.0 & 86.0 \\
        LPIS-H & 23.8 & 44.6 & 57.5 & 48.7 & 61.7 & 6.7 & 0.0 & 42.7 & 69.6 & 5.1 & 15.4 & 16.7 & 0.0 & 0.0 & 7.9 & 21.4 & 0.0 & 16.6 & 0.0 & 0.0 & 0.0 & 86.2 \\ 
        \rowcolor{black!5}LPIS-I & \bestb{39.2} & 47.6 & \bestb{65.7} & 46.0 & 74.5 & \bestb{14.0} & \bestb{57.0} & 44.1 & \bestb{81.6} & \bestb{51.8} & 8.7 & \bestb{28.2} & 75.2 & 7.2 & 22.8 & 33.0 & 14.2 & 27.8 & \bestb{29.8} & 0.3 & \bestb{5.5} & 87.6 \\
        LPIS-J & 35.4 & \bestb{52.0} & 57.4 & 31.0 & \bestb{78.3} & 8.2 & 10.6 & 45.8 & 71.9 & 33.7 & 8.9 & 27.2 & 75.3 & \bestb{14.4} & 22.1 & \bestb{44.6} & 16.4 & 36.6 & 6.6 & 12.1 & 2.3 & 88.7 \\
        LPIS-K & 15.1 & 42.5 & 34.3 & 11.2 & 28.8 & 0.3 & 0.0 & 2.9 & 20.6 & 0.0 & 0.0 & 1.9 & 1.5 & 0.0 & 17.7 & 27.6 & 0.4 & 26.5 & 12.0 & 0.0 & 0.1 & 87.4 \\        
        \bottomrule
    \end{tabular}
}
\end{table*}

\Cref{tab:Baselines_LPIS} presents the performance of various configurations for the crop mapping task. Some classes are excluded from the mIoU computation due to their absence in the test set. Overall, the results underscore the difficulty of the task, particularly as the number of target classes increases. This is especially evident in the lower mIoU scores observed in the last rows of the table, which correspond to more detailed nomenclatures. The performance degradation is largely attributable to strong class imbalance and the presence of rare or sparsely represented classes. A more granular analysis is provided in \Cref{tab:LPIS_per_class_perf}, which reports per-class IoU scores for the Level-1 crop mapping task across different input modality configurations. These results are based on the Swin-Base UPerNet baseline, with auxiliary losses applied in all multimodal settings. This table highlights the impact of input modality combinations on class-wise performance and reveals the high variability in segmentation accuracy, particularly for rare crop types.\\

\noindent\textbf{Severe Imbalance in Crop Type Distribution}: It is important to note that the ROIs in the dataset were originally selected for the land cover mapping task. As a result, the class distribution is highly skewed for crop classification: the background class alone accounts for approximately 78\% of the test set, grasses represent about 12\%, and all other crop classes individually account for less than 2\% (as it can be seen in \Cref{tab:LABEL_LPIS}). This imbalance significantly impacts learning and evaluation for most classes. Given the strong class imbalance in the dataset, OA is more reflective of the model’s training dynamics and performance on dominant classes, while mIoU offers complementary insight into class-wise behaviour. For the Level-1 nomenclature, the highest OA (88.0\%) is achieved by both LPIS-D and LPIS-J, which include aerial VHR imagery combined with other modalities. The best mIoU (39.2\%) is obtained by LPIS-I, which excludes aerial imagery and relies solely on SPOT and Sentinel-1/2 time series.
\Cref{tab:LPIS_per_class_perf} illustrates the differing behaviour of crop type classes, which is closely correlated with their frequency at Level-1 in the dataset. The background class, which accounts for approximately 78\% of the dataset, consistently achieves the highest IoU scores across all configurations (LPIS-A to LPIS-J), with a large margin exceeding 28\% over the second-best class. Grasses, the second most frequent class (around 12\%), also attain relatively high IoU values across all configurations, ranging from 37\% to 52\%. For the remaining classes, which are less frequent (between 2.5\% and 0.5\%), IoU scores exhibit greater variability across configurations. This variability appears to be somewhat correlated with the presence of specific modalities in the input configuration and therefore varying acquisition dates. However, for very rare classes, it becomes difficult to identify consistent patterns. A likely explanation is that the scarcity of training examples for these low-frequency classes results in high variability between training runs of the same configuration—variability that is comparable to that observed between different configurations thus limiting the ability to draw reliable conclusions.

\noindent\textbf{Comparing Crop Type and Land Cover Mapping Tasks}: While OA increases from LC-A to LPIS-A, the mIoU drops significantly. This decrease is expected, as land cover classes are specifically designed to be distinguishable using mono-temporal aerial imagery, whereas many LPIS classes require multi-temporal information for correct identification. As a result, class confusion is more pronounced in the LPIS labelling task. The increase in OA may seem counterintuitive. A likely explanation is that the LPIS classes dominating the pixel distribution (Background, Grasses, and Vineyards) can still be reliably segmented from aerial imagery alone. These three classes represent over 88\% of the test set pixels, thus inflating OA despite lower class-wise performance. As expected, multi-temporal modalities prove more beneficial for the LPIS task than for land cover segmentation. The addition of Sentinel-1 or Sentinel-2 time series significantly improves mIoU, with an 11.9\% gain from LPIS-A to LPIS-D. Moreover, configurations using only S1 or S2 (\textit{e.g.}, LPIS-F) perform comparably to the aerial-only LPIS baseline (just a 0.6\% drop in mIoU), whereas they perform substantially worse on the land cover task (showing a 29.9\% drop in mIoU from LC-A to LC-G).

\begin{table*}[b]
\caption{{\bf Per-Class Evaluation for Crop Mapping.}  
Class-wise IoU scores for the Base-UP baseline (setting LPIS-I) on the Validation and Test partition. The \textit{rice} and other \textit{oilseed crops} classes are excluded from mIoU computation due to having zero pixels in the test set.
}
\label{tab:Baselines_LPIS_ValTest}
\centering
\scriptsize
\renewcommand{\arraystretch}{1.5}
\setlength{\tabcolsep}{3.5pt}
\scriptsize{
    \begin{tabular}{l|c|cccccccccccccccccccccc}
    \toprule
        
        \makecell[l]{Class $\rightarrow$ \\
        Set $\downarrow$}& 
        \rotatebox{80}{mIoU} &
        \rotatebox{80}{grasses}  & 
        \rotatebox{80}{wheat}  & 
        \rotatebox{80}{barley}    &
        \rotatebox{80}{maize} & 
        \rotatebox{80}{o. cereals}  & 
        \rotatebox{80}{rice}  & 
        \rotatebox{80}{\makecell[l]{flax/hemp\\tobacco}}  & 
        \rotatebox{80}{sunflower}  & 
        \rotatebox{80}{rapeseed}   &  
        \rotatebox{80}{soy}   & 
        \rotatebox{80}{o. protein c.}  &
        \rotatebox{80}{\makecell[l]{fodder\\legumes}}     & 
        \rotatebox{80}{beetroots} & 
        \rotatebox{80}{potatoes} & 
        \rotatebox{80}{o. arable c.}     & 
        \rotatebox{80}{vineyard}     & 
        \rotatebox{80}{olive groves}  & 
        \rotatebox{80}{\makecell[l]{fruits\\orchards}}& 
        \rotatebox{80}{nut orchards}     & 
        \rotatebox{80}{o. permanent c.} & 
        \rotatebox{80}{mixed c.}    &  
        \rotatebox{80}{background}   \\ 
        \midrule 
        Validation & 81.4 & 43.7 & 77.3 & 62.4 & 78.2 & 16.9 & 60.1 & 40.9 & 83.7 & 89.8 & 40.6 & 57.6 & 31.2 & 79.5 & 50.3 & 14.8 & 48.3 & 11.5 & 47.5 & 6.5 & 0.1 & 2.4 & 89.7 \\
        
        Test & 39.2 & 47.6 & 65.7 & 46.0 & 74.5 & 14.0 & - & 57.0 & 44.1 & 81.6 & 51.8 & 8.7 & 28.2 & 75.2 & 7.2 & 22.8 & 33.0 & 14.2 & 27.8 & 29.8 & 0.3 & 5.5 & 87.6
 \\ 
        \bottomrule
    \end{tabular}
}
\end{table*}

\noindent\textbf{Limitations of Modality Contribution}: Configurations that combine multi-temporal inputs (S1, S2, or both) with high-resolution imagery (aerial VHR or SPOT) generally achieve the best performance for the LPIS task. This likely results from the complementarity of modalities: high-resolution imagery provides fine-grained textural features, while time series data capture phenological dynamics critical for distinguishing between crop types. However, performance differences across specific modality combinations are not always intuitive. For instance, the Aerial+S2 configuration (LPIS-C) does not consistently outperform either of the single-modality baselines such as Aerial (LPIS-A) or S2 (LPIS-F). In some cases, the differences are substantial: for example, barley and rapeseed exhibit IoU drops of 21\% and 12.2\%, respectively, in LPIS-C compared to LPIS-F. A similar pattern is observed when comparing LPIS-I and LPIS-J: adding aerial imagery to a configuration using SPOT and Sentinel-1/2 time series (LPIS-I) results in large IoU decreases for Flax/Hemp ($-40.6$\%) and Soy ($-18.1$\%). These findings suggest that simply adding modalities does not guarantee improved performance. The observed inconsistencies may stem from limitations in both the dataset and the model architecture. Potential contributing factors include increased model complexity leading to overfitting, interactions between the main and auxiliary losses when adding modalities, and limitations of the current temporal encoder. These factors indicate a need for more effective fusion strategies and potentially stronger architectures to fully leverage multimodal data.

\noindent\textit{Aerial / SPOT modalities}: As expected, the mono-temporal modalities and particularly the aerial imagery yield higher IoU scores for classes associated with textural patterns visible in very high-resolution (VHR) imagery. Vineyards and orchards are notable examples, with very low IoU when using only S1 and/or S2 modalities, but a significant IoU increase when aerial data is included in the configuration. More surprisingly, a similar trend is observed for beetroot and potato classes, which is more difficult to explain. One hypothesis is that these crops may exhibit unique spatial or structural characteristics during the aerial acquisition period that are detectable at 20\:cm resolution. Some classes, such as wheat and maize, also show relatively high IoU scores using only aerial imagery. For maize, this may be because many aerial acquisitions occur when the crop is nearly fully grown, resulting in distinct textural patterns at VHR scale (see \Cref{fig:tempo-mono}). For wheat, the explanation is less straightforward. However, given the low IoU scores for other cereals (\textit{e.g.}, barley) and the relatively higher frequency of wheat in the dataset, it is plausible that the model learns generalized cereal textures and defaults to classifying all cereals (except maize) as wheat. To better understand why LPIS-I (which excludes aerial imagery) outperforms LPIS-J (which includes all modalities), we introduced an additional configuration, LPIS-K, using only the SPOT modality. Unfortunately, this experiment did not provide clear answers. In LPIS-K, background IoU sits between that of LPIS-A (Aerial) and LPIS-F (S2), which aligns with expectations based on the relative spatial resolutions (20\:cm, 1.5\:m, and 10\:m, respectively). For most other classes, LPIS-K produces lower or comparable IoU values relative to LPIS-A, with larger drops for classes characterized by fine spatial textures (\textit{e.g.}, maize, vineyard) than for those with broader spatial features (\textit{e.g.}, orchards).

\noindent\textit{Temporal modalities}: As anticipated, S2 appears to be one of the most informative sources for the LPIS task. When added to Aerial VHR, it leads to substantial performance gains: LPIS-C outperforms LPIS-A by +5.4\% mIoU, and LPIS-E outperforms LPIS-B by +4.2\%. S2 is also part of the best performing configuration, LPIS-I. Overall, the addition of S2 improves the IoU for most cereal, oleaginous, and proteinous crop classes, with many classes showing gains between +5\% and +15\% compared to the same configuration without S2. In contrast, Sentinel-1 (S1) performs poorly as a stand-alone modality, which aligns with expectations given its sensor characteristics. Its contribution becomes more nuanced when combined with other modalities. For instance, LPIS-D (Aerial VHR + S2 + S1) significantly improves over LPIS-C (Aerial VHR + S2), yielding a +6.5\% mIoU gain. However, adding S1 to S2 in LPIS-H provides no benefit over LPIS-F, and the improvement from LPIS-E to LPIS-J is modest (+2.0\%). These results suggest that S1 contributes most effectively when paired with a high-resolution modality, such as Aerial VHR or SPOT.\\

\begin{table*}[!t]
\caption{{\bf Per-Class Evaluation for Crop Type Mapping – KFold evaluation.} 
Class-wise IoU scores for the Swin Base-UP baseline using SPOT imagery and Sentinel-1/2 time series (denoted setting LPIS-I). mIoU is computed for each fold by excluding classes that have no corresponding pixels in the test set.
}
\label{tab:Baselines_LPIS_KFOLD}
\scriptsize
\centering
\setlength{\tabcolsep}{2.2pt}
\renewcommand{\arraystretch}{1.3}
\begin{tabular}{ll|cc|cccccccccccccccccccccc|cr}
\toprule
\rotatebox{0}{Model ID}&
\rotatebox{0}{CONF.} &
\rotatebox{80}{mIoU}&
\rotatebox{80}{O.A.}&
\rotatebox{80}{grasses}  & 
\rotatebox{80}{wheat}  & 
\rotatebox{80}{barley}    &
\rotatebox{80}{maize} & 
\rotatebox{80}{o. cereals}  & 
\rotatebox{80}{rice}  & 
\rotatebox{80}{\makecell[l]{flax/hemp\\tobacco}}  &
\rotatebox{80}{sunflower}  & 
\rotatebox{80}{rapeseed}   &  
\rotatebox{80}{soy}   & 
\rotatebox{80}{o. protein c.}  & 
\rotatebox{80}{\makecell[l]{fodder\\legumes}}     & 
\rotatebox{80}{beetroots} & 
\rotatebox{80}{potatoes} & 
\rotatebox{80}{o. arable c.}     & 
\rotatebox{80}{vineyard}     & 
\rotatebox{80}{olive groves}  & 
\rotatebox{80}{\makecell[l]{fruits\\orchards}}& 
\rotatebox{80}{nut orchards}     & 
\rotatebox{80}{o. permanent c.} & 
\rotatebox{80}{mixed c.}    &  
\rotatebox{80}{background} &
\rotatebox{80}{PARAM.} &
\rotatebox{80}{EP.}

\\\midrule
\rowcolor{black!5} LPIS-I & split\_1 & 39.2 & 87.2 & 47.6 & 65.7 & 46.0 & 74.5 & 14.0 & - & 57.0 & 44.1 & 81.6 & 51.8 & 8.7 & 28.2 & 75.2 & 7.2 & 22.8 & 33.0 & 14.2 & 27.8 & 29.8 & 0.3 & 5.5 & 87.6 & 97.5 & 49\\
LPIS-I & split\_2 & 34.2 & 88.7 & 44.8 & 69.8 & 56.4 & 74.4 & 14.8 & 2.7 & 1.9 & 61.4 & 85.1 & 32.5 & 17.1 & 31.6 & 69.7 & 16.0 & 24.8 & 45.4 & 6.2 & 36.5 & 7.2 & 1.2 & 3.5 & 89.5 & 97.5 & 108 \\
\rowcolor{black!5} LPIS-I & split\_3 & 39.2 & 88.0 & 42.3 & 75.6 & 68.0 & 67.2 & 10.8 & 30.2 & 22.7 & 60.2 & 91.4 & 34.3 & 39.1 & 27.8 & 82.8 & 4.3 & 9.9 & 46.4 & 14.3 & 37.3 & 6.6 & 0.8 & 1.3 & 89.0 & 97.5 & 119 \\
LPIS-I & split\_4 & 40.7 & 85.2 & 37.9 & 73.3 & 54.9 & 76.8 & 6.5 & - & 17.9 & 74.8 & 81.3 & 48.8 & 47.2 & 20.4 & 74.2 & 38.7 & 13.4 & 37.8 & 20.8 & 34.0 & 0.3 & 2.9 & 5.4 & 86.6 & 97.5 & 57\\
\rowcolor{black!5} LPIS-I & split\_5 & 42.8 & 87.4 & 45.0 & 72.4 & 56.0 & 73.7 & 9.5 & - & 8.3 & 77.3 & 81.6 & 49.5 & 36.8 & 20.8 & 68.7 & 42.5 & 11.1 & 46.5 & 17.3 & 47.4 & - & 2.0 & 2.2 & 88.0 & 97.5 & 43\\

\bottomrule 
\end{tabular}
\end{table*}

\Cref{tab:Baselines_LPIS_ValTest} presents class-wise IoU scores for the best configuration (LPIS-I) on both the validation and test sets, confirming previous observations. A clear drop in performance is observed on the test set, with the overall mIoU decreasing from 81.4\% to 39.2\%. Several classes, such as fodder legumes, beetroots, and mixed crops, show substantial declines, while others like soy and olive groves are no longer detected at all. This discrepancy highlights the greater difficulty of the test set and the model’s limited ability to generalize to unseen regions. This performance gap is primarily due to the severely under-represented classes and data split strategy, where the validation set shares zones with the training data, while the test set covers distinct domains. Furthermore, an additional challenge of the \FLAIRHUB crop mapping task is that the different \textit{domains} span three years, introducing both spatial and temporal generalization issues. Crop type classes exhibit greater variability across years compared to land cover classes, making the task particularly sensitive to temporal shifts.\\

\begin{table*}[b]
\caption{\textbf{Quantitative Evaluation in the Multi-task Setting.}  
Performance of the UPerFuse architecture for both land cover and crop mapping tasks. The evaluation is performed using the best input modality configuration for each task. PARA.: number of model parameters (in millions).\; EP.: epoch with best validation score.\; SITS: Satellite Image Time Series. \; S1/2: Sentinel-1/2.
}
\label{tab:Baselines_Multitask}
\centering
\scriptsize
\renewcommand{\arraystretch}{1.5}
\setlength{\tabcolsep}{3.5pt}
\scriptsize{
    \begin{tabular}{l|l|ccccc|cc|cccc}
    \toprule
        Training & Model ID & Aerial VHR & Elevation & SPOT & S2 SITS & S1 SITS & PARA. & EP. & LC mIoU & LC O.A. & LPIS mIoU & LPIS O.A. \\ 
        \midrule 
        Only Land Cover & LC-L & \checkmark & \checkmark & \checkmark & \checkmark & \checkmark &  276.4 & 121 & 65.8 & 78.4 & X & X \\
        \rowcolor{black!5} Only Crop Mapping & LPIS-I & & & \checkmark & \checkmark & \checkmark & 97.5 & 53 & X & X & 39.2 & 87.2 \\
        \midrule
        Multi-task & LC-L  & \checkmark & \checkmark & \checkmark & \checkmark &  \checkmark & 286.6 & 81 & 64.7 & 77.9 & 34.7 & 88.1 \\
        \rowcolor{black!5} Multi-task & LPIS-I &  & & \checkmark & \checkmark & \checkmark & 102.6 & 87 & 47.8 & 66.9 & 36.1 & 87.6 \\        
        \bottomrule
    \end{tabular}}
\end{table*}

In \Cref{tab:Baselines_LPIS_KFOLD}, the performance of the LPIS-I configuration is reported across the five folds of the KFold cross-validation. OA remains consistently high across splits, ranging from 85.2\% to 88.7\%, while mean IoU (mIoU) varies more substantially, from 34.2\% to 42.8\%, reflecting sensitivity to specific domain characteristics, supervision availability and acquisition dates regarding phenologies. It is important to note that mIoU is computed only over the classes present in the test set for each split, excluding those marked with dashes. This introduces some complexity when comparing mIoU values across folds, as the class composition can differ. Notably, major crop classes such as wheat, maize, and rapeseed show relatively stable and high IoU scores, whereas rare or under-represented classes (\textit{e.g.}, nut orchards, tobacco, and other permanent crops) exhibit large variability or near-zero performance. These results highlight the strong impact of class imbalance and domain-specific variation on class-wise segmentation performance.\\

\subsection{Multitask training}

\Cref{tab:Baselines_Multitask} presents a quantitative evaluation of the UPerFuse architecture in a multitask setting, comparing its performance on land cover and crop type mapping when trained either separately or jointly. Both tasks has been assigned the same weight for the experiments. Overall, the results indicate that multitask learning does not yield performance improvements for either task. In fact, crop mapping performance slightly decreases in the multitask setting, with LPIS mIoU dropping from 39.2\% (single-task) to 36.1\%, while land cover results remain relatively stable (65.8\% to 64.7\% mIoU). This suggests that land cover segmentation has a more robust learning, likely due to more balanced class distributions, greater visual separability in the data and sufficient learning data. The degradation in LPIS performance aligns with earlier observations about the complexity of the crop mapping task: the strong class imbalance, scarcity of rare crop types, and reliance on subtle temporal dynamics make it more sensitive to architectural or training changes. Additionally, the moderate increase in model parameters in the multitask setting does not appear sufficient to offset this trade-off. These findings highlight the need for more tailored multitask architectures, as well as improved strategies to handle data imbalance in crop type segmentation.\\

\subsection{Qualitative results}

\begin{figure*}[!h]
    \centering
    \includegraphics[width=0.8\textwidth]{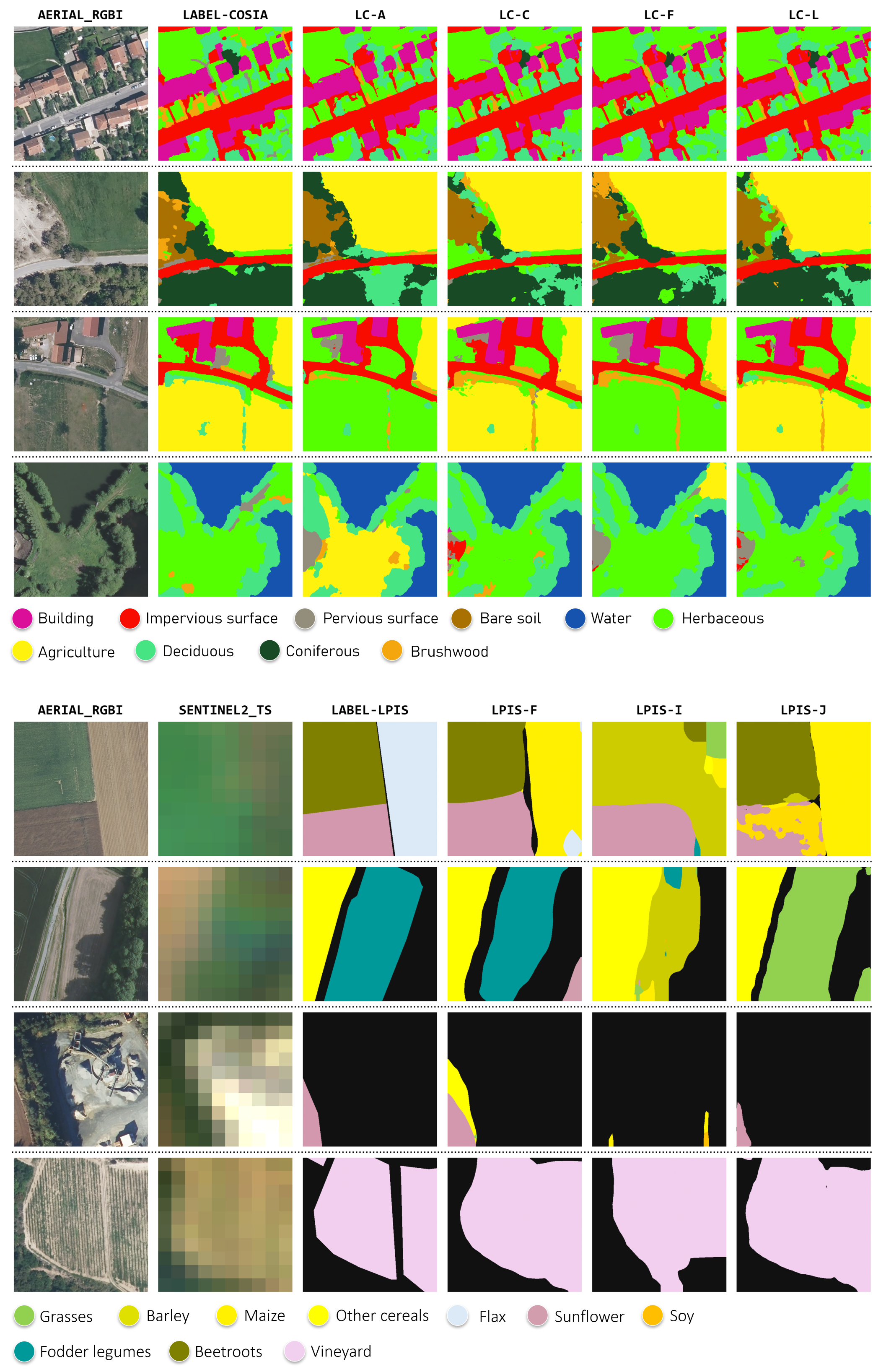}
    \caption{{\bf Comparison of patch-level inference of LC and LPIS models.} Top: LC results. Bottom: LPIS results. One can compare the ground truth (labels COSIA or LPIS) with the predictions of different unimodal or multimodal models. Details of the input data for each model can be found in the \Cref{tab:Baselines_LC,tab:Baselines_LPIS}.}
    \vspace{-2em}
    \label{fig:cosia-patches-inf}
\end{figure*}

\Cref{fig:cosia-patches-inf} presents patch-level inferences on the test split across various models for both the land-cover (top) and crop-type (bottom) classification tasks. For the land-cover task, the results indicate that using aerial imagery alone already yields highly accurate predictions. As such, models incorporating additional modalities that include this source exhibit only marginal improvements. Nonetheless, certain confusions—such as those between tree types or between agricultural and herbaceous covers—are slightly mitigated when temporal information from complementary modalities is introduced. In contrast, the LPIS task remains significantly more challenging, as previously discussed. Quantitative performance is notably lower, and several rare classes are often not retrieved, with predictions defaulting to one of the dominant categories in the supervision dataset. Despite this limitation, some parcels are accurately classified. The LPIS-J model, which integrates aerial, SPOT, Sentinel-1, and Sentinel-2 imagery, appears to yield the most visually coherent and accurate results.\\

\begin{figure*}[!b]
    \centering
    \includegraphics[width=1\textwidth]{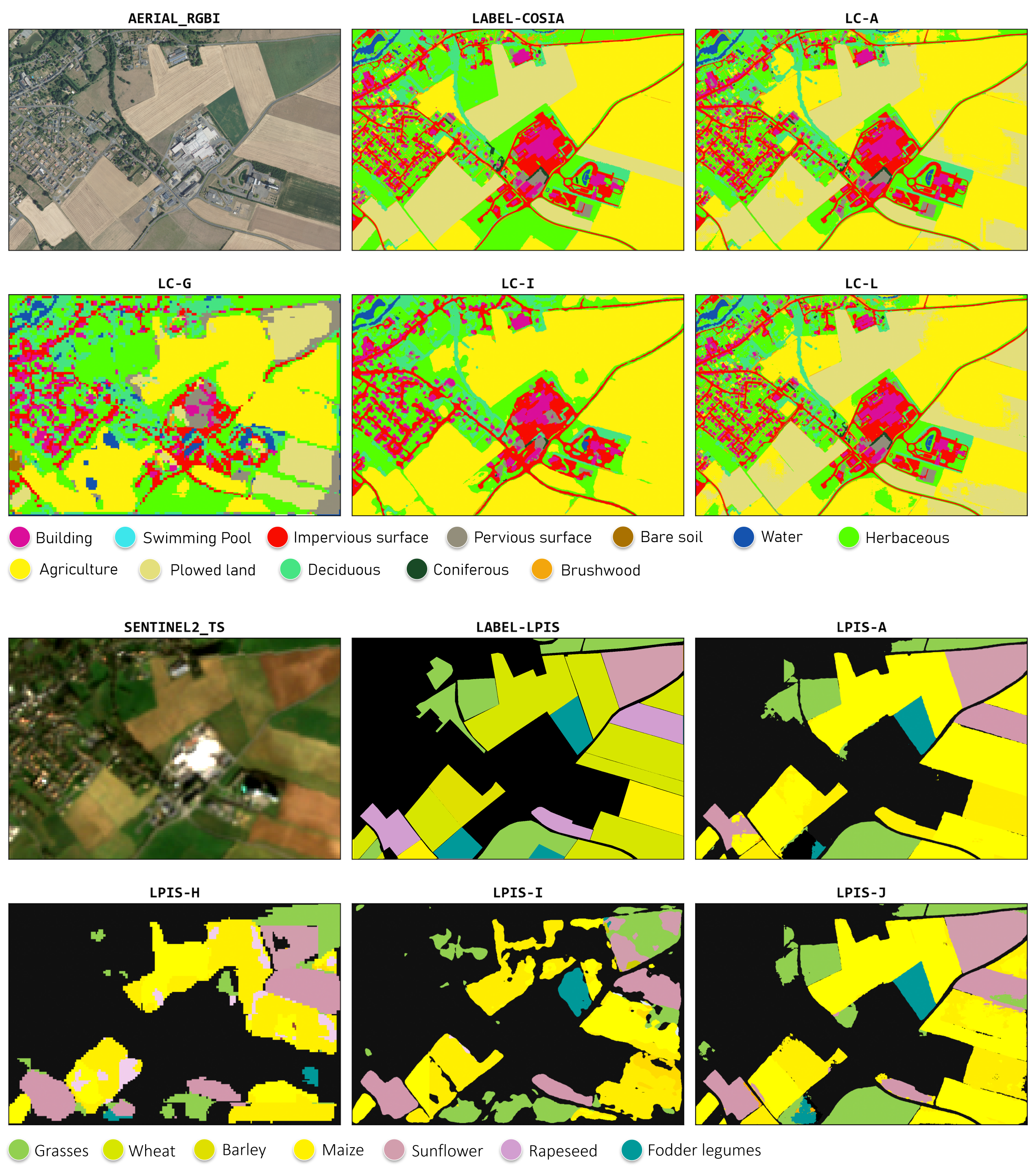}
    \caption{{\bf ROI detections.} We provide inference on large zones to illustrate the capacity of the different monomodal and multimodal models. Details of the input data for each model can be found in the \Cref{tab:Baselines_LC,tab:Baselines_LPIS}.}
        \label{fig:zonal_detections}
\end{figure*}

\begin{figure*}[!h]
    \centering
    \includegraphics[width=0.80\textwidth]{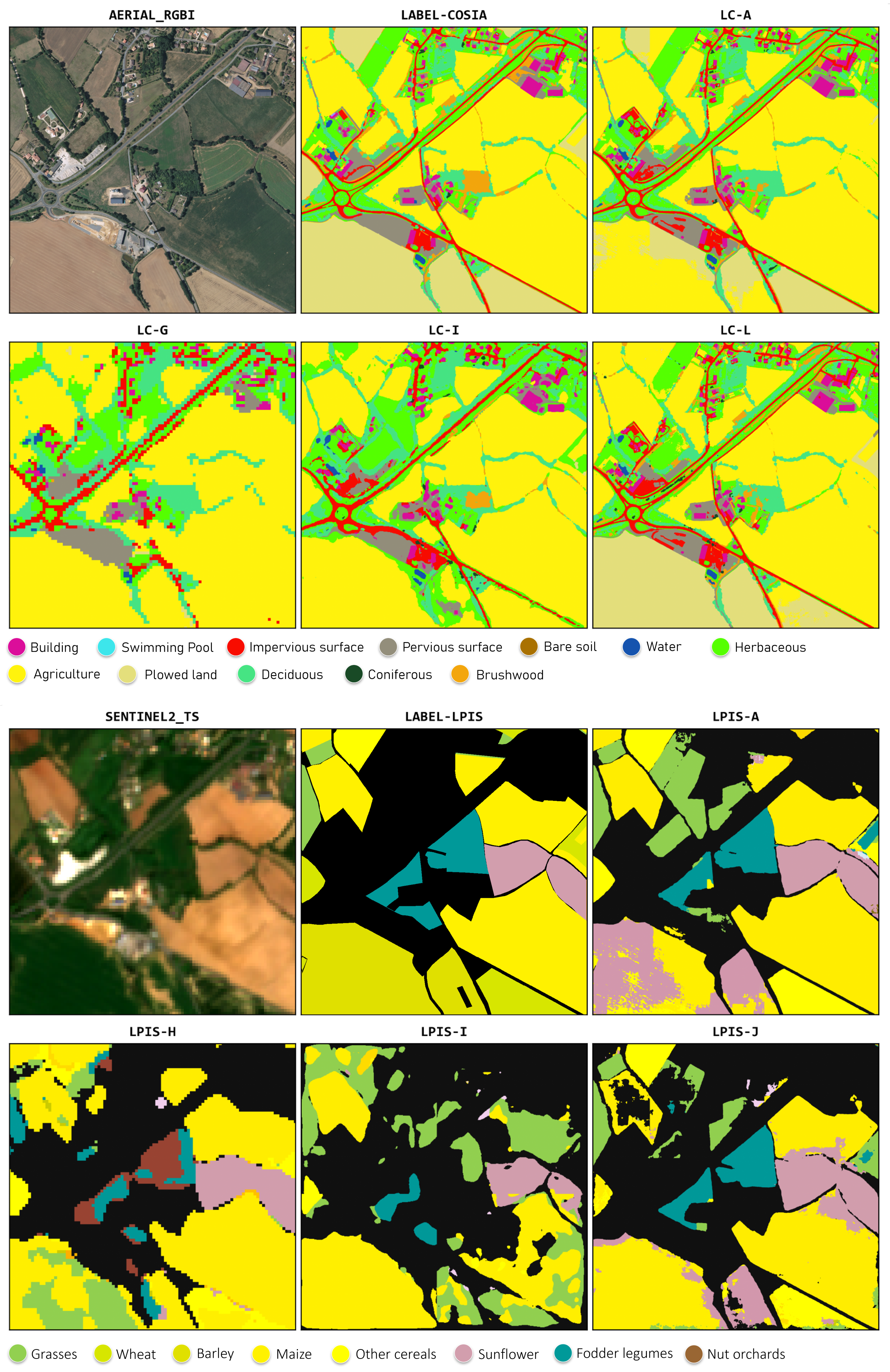}
    \caption{{\bf ROI detections.} We provide inference on large zones to illustrate the capacity of the different monomodal and multimodal models. Details of the input data for each model can be found in the \Cref{tab:Baselines_LC,tab:Baselines_LPIS}.}
        \label{fig:zonal_detections2}
        \vspace{-2em}
\end{figure*}

To further evaluate model performance, predictions were extended to larger geographic regions. This qualitative assessment is crucial due to the spatial continuity inherent in geospatial data. The input ROI is divided into patches, and overlapping inferences are integrated to reduce edge effects. Results are shown over two ROI of the test-set in \Cref{fig:zonal_detections} and \Cref{fig:zonal_detections2}.

For the land-cover task, four models are visualized: aerial-only (LC-A), Sentinel-2 (LC-G), SPOT-only (LC-I), and the best-performing configuration (LC-L). Region-level predictions are consistent with patch-level observations, affirming strong performance in the land-cover task. The Sentinel-2-only model (LC-G) delivers less precise results, as expected given its coarser spatial resolution of 10.24\:m. Moreover, this model employs only the UTAE architecture and appears to generalize less effectively over broad areas compared to attention-based alternatives. This limitation is particularly visible in agricultural zones, where class mixing is observed. In comparison, the SPOT-only model (LC-I), benefiting from a finer 1.6\:m resolution, yields accurate predictions, though with less defined parcel boundaries than those seen in the 0.2\:m aerial imagery (LC-A). It is worth noting that, due to differing acquisition dates relative to the aerial imagery used for ground-truth labelling, some classes—such as plowed land—are absent in SPOT-based predictions.

Results for the crop-type classification are markedly more variable. Models relying solely on aerial imagery (LPIS-A), Sentinel-1 and -2 (LPIS-H), SPOT combined with Sentinels (LPIS-I), and the best-performing configuration (LPIS-J) demonstrate differing behaviours. Notably, the aerial-only model performs well in delineating parcel boundaries and accurately classifies dominant crop types. In contrast, models using only Sentinel or SPOT data suffer from reduced spatial resolution, frequently omitting agricultural areas and exhibiting significant class confusions. However, adding very high-resolution aerial imagery to the SPOT and Sentinel modalities in the LPIS-J configuration substantially enhances visual coherence and parcel delineation, despite slightly lower overall quantitative scores compared to LPIS-I.

Overall, land-cover classification performs well due to a balanced dataset and clear annotations, while crop-type classification remains challenging. Temporal information is helpful but not fully exploited, and high spatial resolution appears key in both tasks.

\section{Perspectives}

\subsection{Improving the deep learning model performances}

The encoders used in the mono-temporal architectures of this work were pretrained on ImageNet. However, foundation models trained with self-supervised learning techniques are now widely recognized as more effective for pretraining and transfer \cite{astruc2025omnisat,astruc2024anysat,xiong2024neural,stewart2024ssl4eo,jakubik2023foundation,fuller2023croma,tseng2023lightweight}. Therefore, it would be relevant to assess the benefit of using off-the-shelf foundation models on the \FLAIRHUB dataset. Furthermore, given the large volume of \FLAIRHUB, training dedicated foundation models directly on this dataset appears feasible. This could pave the way for designing foundation models tailored to temporal or multi-modal tasks.

Several metadata provided with the \FLAIRHUB dataset were not used in the experiments presented. However, properly integrating this information into architectures could lead to performance gains \cite{jakubik2025terramind,marsocci2023geomultitasknet}. In particular, certain land cover classes, as well as all classes in the LPIS nomenclature, exhibit significant variations in appearance throughout the year, depending on the image sources. Encoding the acquisition date using the MTD\_DATES metadata could therefore improve the temporal robustness of the models. Additionally, it could be beneficial to encode the geographical location, either through the MTD\_GEOM metadata or by using information from the GeoTIFF file headers. Indeed, class appearances may vary greatly across different French regions, and incorporating spatial context could help better model this geographic variability.

Both supervision nomenclatures exhibit significant variability in class frequency. This aspect was not addressed in the current study, although it is illustrated in Tables~\ref{tab:LABEL_COSIA} and \ref{tab:LABEL_LPIS}. The classes were assigned binary weights of 1 or 0. Investigating weighted loss functions, exploring alternative loss functions \cite{azad2023loss}, or applying different dataset sampling strategies \cite{nogueira2024prototypical} could lead to significant improvements. Hierarchical loss functions \cite{li2022deephierarchical,landrieu2021leveraging} could also be considered for the LPIS task to take advantage of the nested structure of the three levels of the nomenclature.

In our experiments, we focus exclusively on a mid-level fusion strategy. However, other approaches, such as early fusion (input-level fusion) or late fusion (prediction-level fusion) could also be explored \cite{li2022deep}. In addition, late fusion strategies may be particularly well suited to self-supervised pretraining performed independently for each modality.

In the selection of baselines for mono-temporal (Swin UPerNet) and multi-temporal (UTAE) image sources, we observe a significant imbalance in the number of parameters, to the disadvantage of multi-temporal architectures. For instance, as shown in Table \ref{tab:Baselines_LC}, UTAE models have around 1 million parameters, while mono-temporal setups can reach several hundred million. It would be valuable to conduct studies aimed at finding a better trade-off in parameter allocation between mono-temporal and multi-temporal branches.

\subsection{Other Potential Uses of the FLAIR-HUB Dataset}

The \FLAIRHUB dataset could also be of interest to researchers working on transfer learning and unsupervised domain adaptation. 
First, the pretrained weights can be used to fine-tune models on new visual categories or other types of sensors (\textit{e.g.}, UAV, thermal imaging) or new tasks (\textit{e.g.}, panoptic segmentation).

A thematically relevant application of transfer learning would be the development of AI models capable of generating land cover maps of the past \cite{le2020cnn}. To support this, we have included the AERIAL\_195X modality in \FLAIRHUB. A promising direction would be to train models using annotations from recent aerial imagery and apply them to older aerial images. This type of experiment is also methodologically challenging due to the significant changes between the two modalities, including strong radiometric and domain shifts.

Furthermore, novel transfer methods could be evaluated based on their ability to train a model in a given spatial and/or temporal domain and transfer it to another in a supervised or unsupervised manner \cite{ibanez2025inter,lucas2023bayesian}. \FLAIRHUB includes 74 distinct spatio-temporal domains along with the necessary metadata to support such experiments.

In the remote sensing community, many studies focus on super-resolution methods using multiple image sources. Since the \FLAIRHUB dataset provides image patches with spatially aligned modalities, it is particularly well suited for super-resolution methods using single or multiple images \cite{nguyen2021self,kowaleczko2023real,satlassuperres}. Specifically, models that aim to enhance the spatial resolution of Sentinel-2 images using SPOT or aerial images could be effectively evaluated using \FLAIRHUB. The dataset is also tailored for the cloud removal task \cite{UnCRtainTS} on optical images thanks to SAR ones. 

The \FLAIRHUB dataset's Land Cover and LPIS annotations provide an interesting opportunity for remote sensing image synthesis \cite{khanna2024diffusionsat,dash2023review}. These annotations and metadata can serve as prompts for controlling generative models. These models could learn to capture the relationship link between land cover classes and the visual characteristics of the various sensors enabling the creation of augmented datasets. Such synthetic data could be used for tasks where annotations are scarce or expensive to obtain, like change detection \cite{benidir2025change}.

\subsection{Possible Future Extensions of the FLAIR-HUB Dataset}

First, the \FLAIRHUB dataset is limited to metropolitan France. Although France's territory is quite diverse, featuring oceanic, continental, Mediterranean, and mountainous bioclimatic regions, it does not contain tropical or desert areas. The aerial images were captured under favorable weather conditions between April and November, leading to a bias in the acquisition dates (see \Cref{fig:tempo-mono}). It could be interesting to expand the dataset to other countries or sensors (such as UAV) with a large variation of acquisition conditions (\textit{e.g.}, angle, weather) but with an interoperable nomenclature to learn more generic models.

In the coming years, we plan to enhance \FLAIRHUB by adding new modalities, metadata, or tasks. For instance, we are waiting for the completion of the national coverage of France with high-definition LIDAR \cite{lidarhd} (at a density of 10 to 20 points per square meter) to incorporate this point cloud modality as in \cite{xu2019advanced,weinstein2020benchmark}. 
Hyperspectral images, such as those in \cite{xu2019advanced,hu2023mdas}, could also be included. 
Additional auxiliary data such as weather, transportation, and socioeconomic indicators \cite{yan2024multimodal} could also be integrated into \FLAIRHUB for multimodal learning studies.

For approximately 200 ROIs in the \FLAIRHUB dataset, we are currently generating orthoimages from historical aerial imagery, spanning from 1960 to 2015. These images will be accompanied by temporally and spatially consistent land cover annotations for the semantic segmentation task. This new supervision dataset will be released upon completion. Historical aerial imagery presents significant variability in spatial resolution and radiometric characteristics. 

Beyond its thematic interest such as enabling models to produce temporal series of LABEL-COSIA, this dataset also offers a valuable opportunity to study the generalization capabilities of AI models when faced with highly heterogeneous spatial and spectral inputs.Finally, we also have other types of labels that could be made available in FLAIR-HUB, including object detection tasks (\textit{e.g.}, wind turbines, solar farms) or semantic segmentation tasks (\textit{e.g.}, roads, hedgerows, isolated trees).

One of the most likely annotation extensions involves expanding the LPIS-labelled areas. As mentioned previously, a current limitation of the \FLAIRHUB dataset lies in the significant imbalance of LPIS classes, which likely results in poor model training and high variability across training runs. Therefore, a priority for improving crop mapping performance lies more in enhancing the dataset than refining the model architecture, particularly by increasing the representation of low-frequency crop-type classes.

However, adding new ROIs or tasks raises several important questions. First, could the current best-performing land cover model be leveraged to assist in annotating new ROIs? Second, in the context of multitask learning, is it possible to integrate both land cover and crop-type annotations into a unified labelling scheme, and how would performance compare between this joint task and two separate tasks? We have initiated preliminary investigations to explore these directions, including strategies for selecting new ROIs to improve LPIS class balance (across all three annotation levels), and methods for merging land cover and crop-type labels into a comprehensive, multi-class label set. To reduce the need for manual annotation in new ROIs, we plan to evaluate semi-automated and soft-labelling approaches, using the current best land cover models. Once this improved dataset is available, it will facilitate the exploration of hierarchical classification strategies (particularly for the second and third levels of crop-type annotations) and support the creation of a unified benchmark dataset for evaluating multimodal fusion methods and architectures.

Beyond the crop mapping task, additional annotation types could further enrich the \FLAIRHUB dataset. For example, land cover predictions from \FLAIRHUB models are already employed by IGN to support the development of the OCSGE product \cite{ocsge}. This land use/land cover product features generalized geometries (with a minimum mapping unit of 200\:m$^{2}$ for buildings and 500\:m$^{2}$ for other classes) and results from conflation with existing databases such as LPIS. Initially, the decision was made not to include such generalized labels due to poor results in generalization. However, we now have access to OCSGE annotations for existing \FLAIRHUB ROIs, and they represent a valuable opportunity to assess model performance under varying degrees of label generalization. 

In parallel, it may be worthwhile to introduce instance-level annotations for specific detection tasks (sometimes called panoptic segmentation). Notably, parcel boundary detection could be derived from raw LPIS vector data, while building detection could leverage existing resources such as the INRIA aerial labelling dataset \cite{maggiori2017can} or the WHU building dataset \cite{ji2018fully}. Building detection is especially relevant, as current land cover labels prioritize the topmost visible cover (\textit{e.g.}, trees over buildings), and do not preserve the high-quality building geometries found in dedicated building databases.

Finally, similarly to other work \cite{lu2017exploring}, we are also considering the generation of textual descriptions for ROIs to enable patch-level text annotations. This modality could serve to train CLIP-like models \cite{silva2024multilingual} and enhance few- and zero-shot capabilities, as demonstrated in recent multimodal frameworks \cite{irvin2025teochat}.

\section{Conclusion}

We presented \FLAIRHUB, the largest high-resolution multimodal dataset to date for land cover and crop type mapping. It contains over 63 billion annotated pixels across 2,528 km² of metropolitan France, with six spatially aligned modalities: aerial imagery, SPOT, Sentinel-1 and -2 time series, digital elevation models, and historical aerial photographs. These diverse sources capture a wide range of spatial, spectral, and temporal characteristics.

Through extensive benchmarks using state-of-the-art deep learning models, we highlighted both the challenges and opportunities of multimodal fusion. Our experiments show that combining complementary data sources significantly improves land cover and crop classification. At the same time, they underscore the difficulty of fine-grained crop mapping, multimodal integration, and multitask training in remote sensing.

\FLAIRHUB supports various learning settings, including supervised and self-supervised training, transfer learning, and domain adaptation.

By releasing this large-scale, extensively labelled dataset along with standardized benchmarks, we aim to support reproducible research and foster progress in the remote sensing, geospatial, and machine learning communities. \FLAIRHUB offers value for both methodological development and real-world applications.

\resetSectionNumbering
\section*{Footprint of computations}
The experiments presented in this article required computational resources equivalent to 27\;311 hours on a single NVIDIA Tesla V100 GPU, producing 528.58 kg CO$_{2}$e. Based in France, this corresponds to a carbon footprint of 10.31\:MWh, which is equivalent to 48.05 tree-years (calculated using green-algorithms.org v3.0 \cite{lannelongue2020green}).

\resetSectionNumbering
\section*{Acknowledgment}
The experiments conducted in this study were performed using HPC/AI resources provided by GENCI-IDRIS (Grant 2024-A0161013803, 2024-AD011014286R2 and 2025-A0181013803).

\section*{Data access}
The dataset, pretrained models and codes are available at the following website: \url{https://ignf.github.io/FLAIR/flairhub}. 

\bibliographystyle{unsrt}
\bibliography{BIB.bib}

\clearpage

\end{document}